%% file: main.tex
\documentclass{article}

\usepackage[preprint]{neurips_2026}

\usepackage[utf8]{inputenc} 
\usepackage[T1]{fontenc}    
\usepackage{hyperref}       

\usepackage{url}            
\usepackage{booktabs}       
\usepackage{amsfonts}       
\usepackage{nicefrac}       
\usepackage{microtype}      
\usepackage{xcolor}         
\usepackage{adjustbox}
\usepackage[fixed]{fontawesome5} 
\usepackage{multirow}
\usepackage{amssymb}
\usepackage{amsmath}
\usepackage{cleveref}
\usepackage{subcaption}
\usepackage{placeins}

\Crefname{figure}{Fig.}{Figs.}
\Crefname{equation}{Eq.}{Eqs.}
\Crefformat{equation}{#2Eq.~#1#3}

\DeclareMathOperator*{\argmax}{arg\,max}

\newcommand{\Cow}[1][]{\includegraphics[width=10pt,trim={6cm 9cm 5cm 6cm},clip]{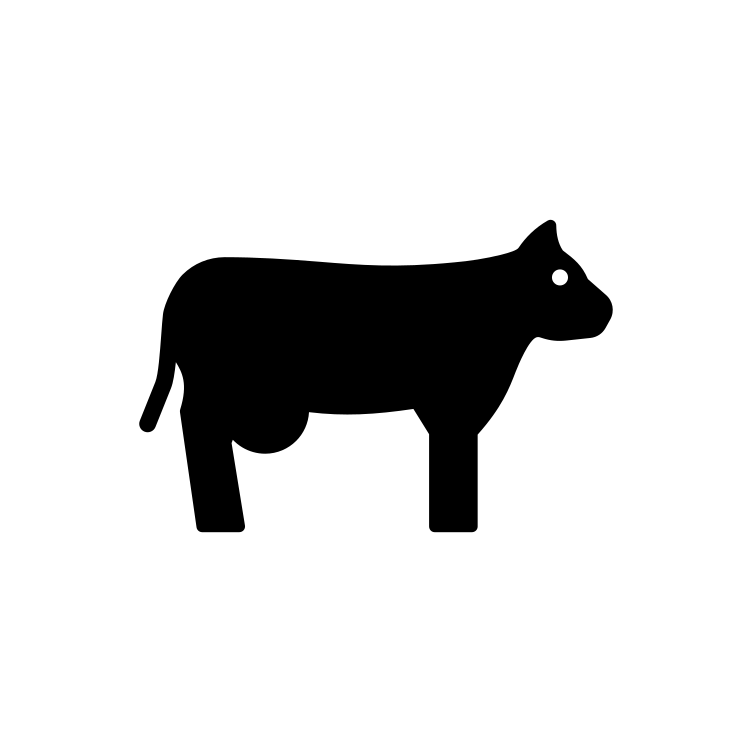}}
\newcommand{\Plant}[1][]{\includegraphics[width=10pt,trim={7cm 6cm 5cm 2cm},clip]{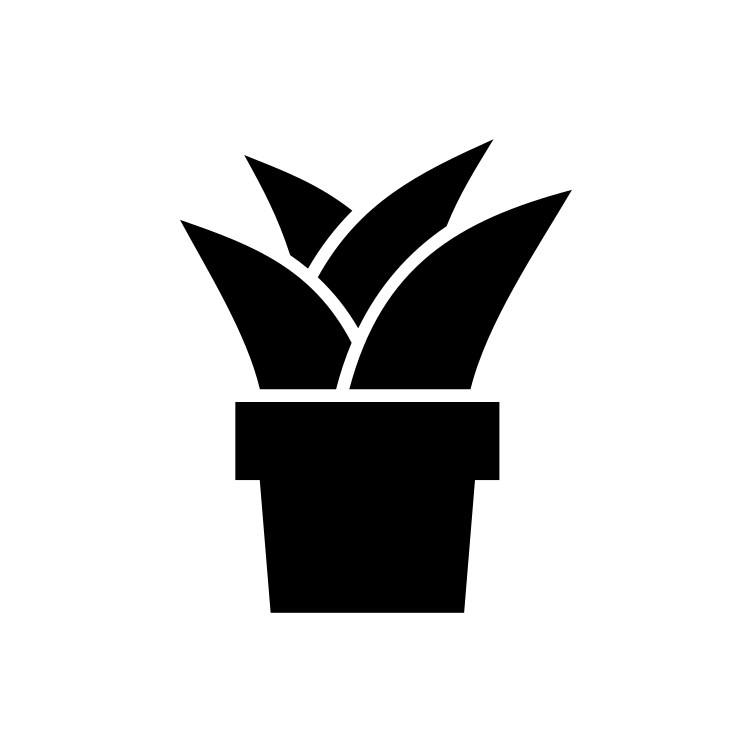}}
\newcommand{\Sheep}[1][]{\includegraphics[width=10pt,trim={6cm 7cm 5cm 6cm},clip]{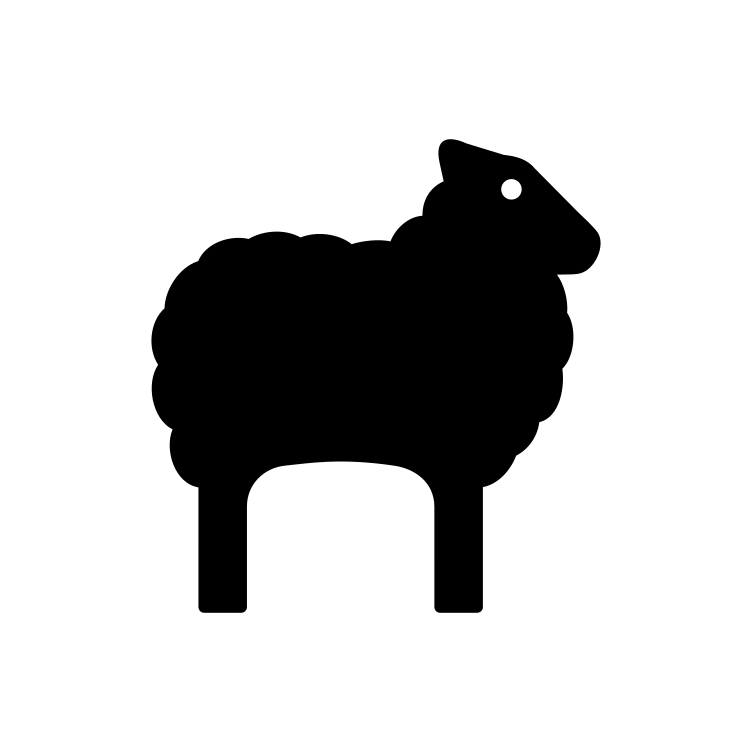}}

\newcommand{\DiningTable}[1][]{\includegraphics[width=10pt,trim={0cm 0.1cm 0cm 0cm},clip]{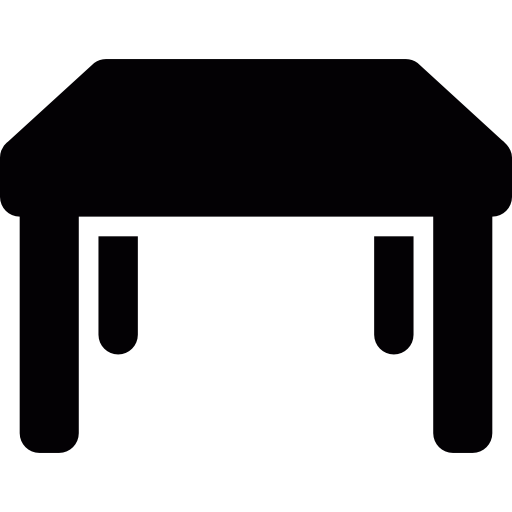}}

\usepackage{colortbl}
\definecolor{tablebg}{gray}{1.0}
\newcommand{\myrowcolor}{\rowcolor[gray]{0.925}\cellcolor{tablebg}}
\definecolor{ourgreen}{RGB}{46, 204, 113}

\title{\emph{Déjà Vu}: Cycle-Consistent Re-ranking for Semantic Correspondence}

\author{%
  Xiaoran Zhang\textsuperscript{1}\thanks{This work was primarily conducted during Xiaoran Zhang’s internship at Amazon.}\quad
  Huidong Liu\textsuperscript{2}\quad
  Yang Liu\textsuperscript{2}\quad
  Chien-Chih Wang\textsuperscript{2} \\
  \bfseries
  Jinfa Huang\textsuperscript{3}\quad
  Lingkai Kong\textsuperscript{4}\quad
  Bryan Wang\textsuperscript{2}\quad
  Yan Gao\textsuperscript{2} \\
  \bfseries
  Hongdong Li\textsuperscript{2,5}\quad
  Alex Wong\textsuperscript{1}\quad
  James S. Duncan\textsuperscript{1} \\[4pt]
  \textsuperscript{1}Yale University\quad
  \textsuperscript{2}Amazon.com, Inc.\quad
  \textsuperscript{3}University of Rochester \\
  \textsuperscript{4}Georgia Institute of Technology\quad
  \textsuperscript{5}Australian National University \\[2pt]
  \small\texttt{\{xiaoran.zhang, alex.wong, james.duncan\}@yale.edu} \\
  \small\texttt{\{liuhuido, yliuu, ccwang, brywan, yanngao\}@amazon.com} \\
  \small\texttt{jhuang90@ur.rochester.edu, lkong75@gatech.edu, hongdong.li@anu.edu.au}
}

\begin{document}

\maketitle

\input{sections/0_abstract}

\input{sections/1_intro}

\input{sections/2_related_works}

\input{sections/3_methods}

\input{sections/4_experiments}

\input{sections/5_results}

\input{sections/6_conclusion}

{\small
\bibliographystyle{plain}
\bibliography{references}
}
\appendix
\input{sections/X_suppl}



\end{document}

%% file: sections/0_abstract.tex
\begin{abstract}
We propose \emph{Déjà Vu}, a test-time decoding module for semantic correspondence. Matching semantically equivalent keypoints across images of the same object category is challenging due to variations in appearance, viewpoint, and scale. While features from large pretrained vision models carry strong zero-shot correspondence ability, existing methods reduce final matching to greedy top-$1$ nearest-neighbor selection. We show that the correct correspondence appears among the top-ranked candidates far more often than top-$1$ matching achieves: with $30$ candidates, zero-shot features exceed the top-$1$ accuracy of a supervised model. This reveals a \textit{decoding bottleneck} rather than a limit of the representation. Motivated by this observation, we formulate semantic correspondence as a \textit{recall-then-rerank} problem. Our method first nominates an adaptive candidate set, larger when the similarity distribution is ambiguous and smaller when it is confident, then re-ranks the candidates with a graded cycle-consistency score: each candidate is matched back to the source image and penalized by how far its reverse match lands from the original keypoint. The module operates entirely on the similarity matrix already computed by the base method. It consistently improves three representative baselines on SPair-71k, its geometric subset, and PF-Pascal under both unsupervised and weakly supervised settings, with gains of up to $+1.8$ PCK@0.10 on SPair-71k and $+3.4$ PCK@0.15 on PF-Pascal, and $+4.9$ under Keypoint Average Precision.%

\end{abstract}

%% file: sections/1_intro.tex
\section{Introduction}
\label{sec:intro}
Semantic correspondence aims to locate semantically equivalent points across images of the same object category and is a fundamental problem in computer vision with applications in pose estimation \cite{leonardis_gs-pose_2025,avidan_neural_2022} and image \cite{mou_dragondiffusion_2023,ofri-amar_neural_2023} or video editing \cite{wang_cove_2024} etc. Unlike geometric correspondence, which benefits from photometric consistency \cite{linguraru_heteroscedastic_2024,leonardis_adaptive_2025,melekhov_dgc-net_2019}, semantic correspondence requires bridging large intra-class variation in appearance, viewpoint, and scale while matching across different objects and therefore is more challenging. This difficulty is especially pronounced when locally similar textures, repeated parts, or large pose changes create multiple plausible matches for the same query.

Classical methods \cite{lowe_distinctive_2004,dalal_histograms_2005,long_convnets_2014,choy_universal_2016} rely on handcrafted features or annotated data, making them difficult to scale. Recent large pretrained vision models have changed this setting. Features from diffusion models~\cite{tang_emergent_2023,luo_diffusion_2023,stracke_cleandift_2025} and self-supervised transformers~\cite{zhang_tale_2023,oquab_dinov2_2023} provide strong semantic representations, enabling zero-shot correspondence without task-specific training. These representations often encode rich part-level semantics and category-level structure, making them surprisingly effective even when they are not explicitly trained for correspondence. However, the final matching step remains largely unchanged: most methods still use greedy top-$1$ nearest-neighbor selection in feature space, making the prediction sensitive to noisy or ambiguous feature responses.

\begin{figure*}[tb]
    \centering
    \includegraphics[width=\linewidth]{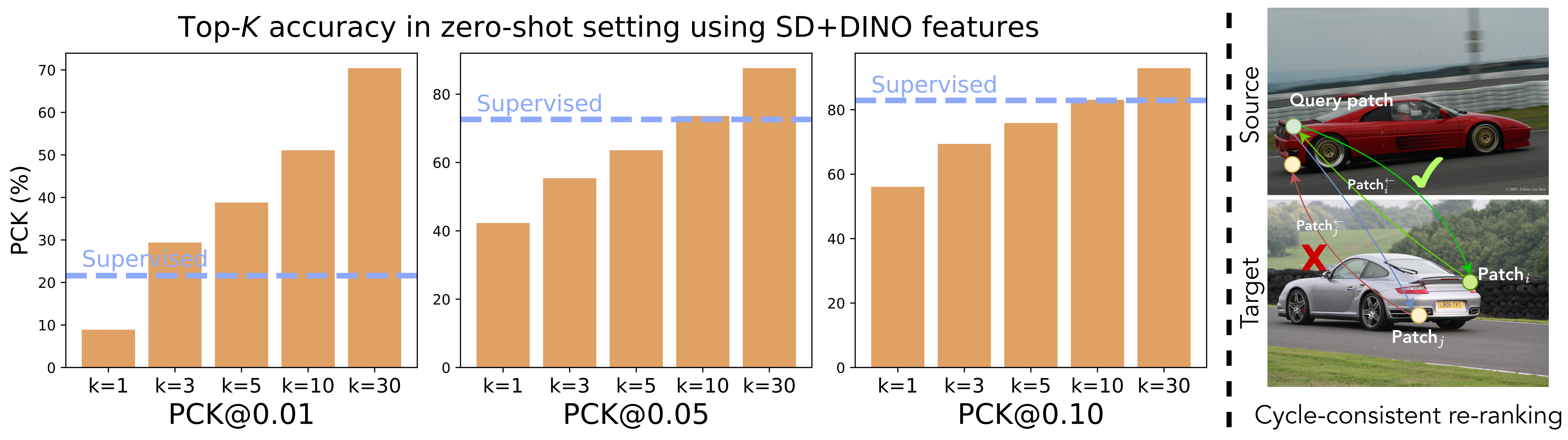}
    \caption{Left: Top-$k$ zero-shot recall on SPair-71k using SD+DINO features. Right: our cycle-consistent re-ranking. The gap between top-$1$ accuracy and top-$k$ recall shows that correct matches often appear among top candidates but are missed by similarity-only top-$1$ matching. As $k$ increases, performance rises sharply and even surpasses the fully supervised baseline, motivating a \textit{recall-then-rerank} strategy (\textit{Déjà Vu}). We therefore exploit the bijective structure of semantic correspondence through cycle-consistent re-ranking of candidates (e.g., Patch$_{i,j}$ in the right figure).}
    \label{fig:teaser}
\end{figure*}

We ask a simple but revealing question: how often is the correct match the single top-ranked candidate, and how often does it appear among the top-$k$ candidates? As shown in \Cref{fig:teaser}, top-$k$ recall in the unsupervised setting rises sharply with $k$, eventually surpassing fully supervised baselines, while top-$1$ matching remains substantially lower. This suggests that correct correspondences are often already present in the candidate set, but are not selected by greedy nearest-neighbor matching. In other words, modern feature spaces often contain the correct answer, but the standard top-$1$ decoding rule collapses the candidate distribution too early. We refer to this as a \textit{decoding bottleneck}: for modern pretrained features, the limiting factor lies not only in representation quality but also in the final decoding step.

This finding motivates a \textit{recall-then-rerank} strategy: first construct a high-recall candidate pool in the pretrained feature space, then select among candidates using a criterion beyond the raw similarity map. We draw inspiration from multi-view geometry, where cycle consistency has long been used to enforce reliable correspondences \cite{zach_disambiguating_2010}. In structure-from-motion and visual localization, the mutual nearest-neighbor (MNN) constraint requires that if a source patch $p$ maps forward to a target patch $q$, then $q$ should also map back to $p$, serving as a reliable inference-time filter for rejecting outliers \cite{wang_learning_2019}. Semantic correspondence can be viewed as an instance of this broader correspondence problem: the similarity is computed in semantic feature space rather than photometric pixel space, and matching occurs across object instances rather than views of the same scene. Nevertheless, the same bijective intuition remains applicable: a correct semantic match should be self-consistent under forward and reverse feature-space queries. Importantly, this criterion does not require learning a new representation or imposing a rigid geometric model; it only checks whether a candidate match remains stable under a reverse query.

Therefore, we propose a training-free, plug-and-play re-ranking module that improves base semantic correspondence methods without modifying their features or retraining. Our method has two stages. First, we generate a high-recall candidate set via top-$p$ sampling~\cite{holtzman_curious_2019}: rather than using a fixed top-$k$ cutoff, we adaptively select the smallest set of candidates whose cumulative similarity probability exceeds a threshold $p$. This yields larger pools for ambiguous queries and smaller pools for confident matches. Second, we re-rank these candidates using a combined score that preserves forward similarity while penalizing cycle-inconsistent matches. The entire procedure operates directly on the similarity matrix computed by the base method at test time. Because our module only changes the final decoding step, it can be seamlessly integrated into existing correspondence pipelines. We apply our method to three representative methods: (1) TLR~\cite{zhang_telling_2024}, (2) CleanDIFT~\cite{stracke_cleandift_2025}, and (3) DIY-SC~\cite{dunkel_it_2025}, covering both unsupervised and weakly supervised settings. We evaluate on SPair-71k~\cite{min_spair-71k_2019}, its geometric subset~\cite{zhang_telling_2024}, and PF-Pascal~\cite{ham_proposal_2016}, and we consistently improve all base methods. 

\textbf{Our contributions}: (1) To the best of our knowledge, we are the first to identify and quantify a \textit{decoding bottleneck} in semantic correspondence: the correct match is present in the top-ranked candidate pool but is missed by top-$1$ selection. (2) We propose a graded re-ranking objective that combines forward similarity with a continuous geometric penalty on the reverse-mapping distance, applied over an adaptively nominated candidate pool. (3) We consistently improve state-of-the-art methods across multiple datasets with minimal computational and memory overhead.

%% file: sections/2_related_works.tex
\section{Related Works}
\label{sec:related_works}

\textbf{Feature-based semantic correspondence.} Classical approaches relied on handcrafted descriptors such as SIFT~\cite{lowe_distinctive_2004} and HOG~\cite{dalal_histograms_2005}, followed by CNN-based methods~\cite{long_convnets_2014,choy_universal_2016} that learned features from labeled data. More recently, features from large pretrained models have shown strong zero-shot correspondence ability. Tang et al.~\cite{tang_emergent_2023} (DIFT) extract correspondence-relevant activations from diffusion models~\cite{rombach_high-resolution_2022,esser_scaling_2024}, while Luo et al.~\cite{luo_diffusion_2023} aggregate these across timesteps and layers. Hedlin et al.~\cite{hedlin_unsupervised_2023} optimize per-query text embeddings to localize keypoints via attention maps. Zhang et al.~\cite{zhang_tale_2023} combine diffusion and DINOv2~\cite{oquab_dinov2_2023} features, and Stracke et al.~\cite{stracke_cleandift_2025} remove noise-level dependency through unsupervised fine-tuning. Fundel et al.~\cite{fundel_distillation_2024} distill these representations into a more efficient model. Despite promising results, these approaches reduce correspondence to a greedy top-1 nearest-neighbor selection, which is not robust to imperfectly localized features.

\textbf{Geometry-aware methods} incorporate structural priors to address the limitations of purely local matching. Zhang et al.~\cite{zhang_telling_2024} align object orientations using a pose template before matching. Mariotti et al.~\cite{mariotti_improving_2024} project features onto a sphere to enforce shape-consistent mappings. Several methods lift the problem to 3D: Qian et al.~\cite{qian_bridging_2025} exploit a 3D reconstruction model for geometric matching in 3D space, Wandel et al.~\cite{wandel_semalign3d_2025} align per-category point clouds to 2D images, and Dunkel et al.~\cite{dunkel_it_2025} generate 3D-aware pseudo-labels to fine-tune off-the-shelf feature extractors. Im et al.~\cite{im_shape--you_2026} combine a Gromov-Wasserstein transport plan with a 3D foundation model. While promising, these methods require extensive 3D pretraining data and remain challenging on non-rigid categories. Critically, even geometry-aware approaches default to top-1 nearest-neighbor selection at inference time~\cite{zhang_tale_2023,zhang_telling_2024,dunkel_it_2025,stracke_cleandift_2025,mariotti_jamais_2025,jin_dual_2026,zhang_semantic_2026}, neglecting the bijective structure inherent to the correspondence.


\textbf{Cycle-consistent matching and adaptive candidate nomination.}
The principle that valid correspondences should be bidirectionally consistent has a long history in visual matching: if a source patch $p$ maps forward to a target patch $q$, then $q$ should map back to $p$. In optical flow and dense tracking, this forward-backward consistency has been widely used to identify occlusions, reject unreliable matches, and provide self-supervised training signals~\cite{hutchison_dense_2010,wang_learning_2019,dwibedi_temporal_2019}. In multi-view geometry and image matching, related loop- or cycle-consistency constraints are used to detect inconsistent relations across view graphs, while mutual nearest-neighbor or cross-check filtering serves as a simple inference-time mechanism for suppressing one-way ambiguous matches~\cite{zach_disambiguating_2010,lowe_distinctive_2004,sarlin_superglue_2020}. Similar ideas have also appeared in dense and semantic correspondence, where cycle consistency has been used for multi-image alignment, weakly supervised training, or representation learning~\cite{tinghui_zhou_flowweb_2015,zhou_learning_2016,chen_deep_2020}. However, our work differs in both where and how this principle is used. Rather than using cycle consistency as a training loss, we use it as a lightweight test-time re-ranking criterion over candidates produced by a base semantic correspondence method. This requires a candidate set with sufficiently high recall. To this end, we integrate the idea of adaptive probability-mass truncation from nucleus sampling~\cite{holtzman_curious_2019}: instead of committing to a fixed top-$k$ cutoff, we nominate the smallest set of candidates whose cumulative similarity probability exceeds a threshold $p$. The resulting procedure separates recall from precision: top-$p$ nomination preserves plausible semantic matches, while cycle-consistent re-ranking selects the candidate that is most stable under the reverse feature-space query.

%% file: sections/3_methods.tex
\section{Method}
\label{sec:method}

\begin{figure*}
    \centering
    \includegraphics[width=\linewidth]{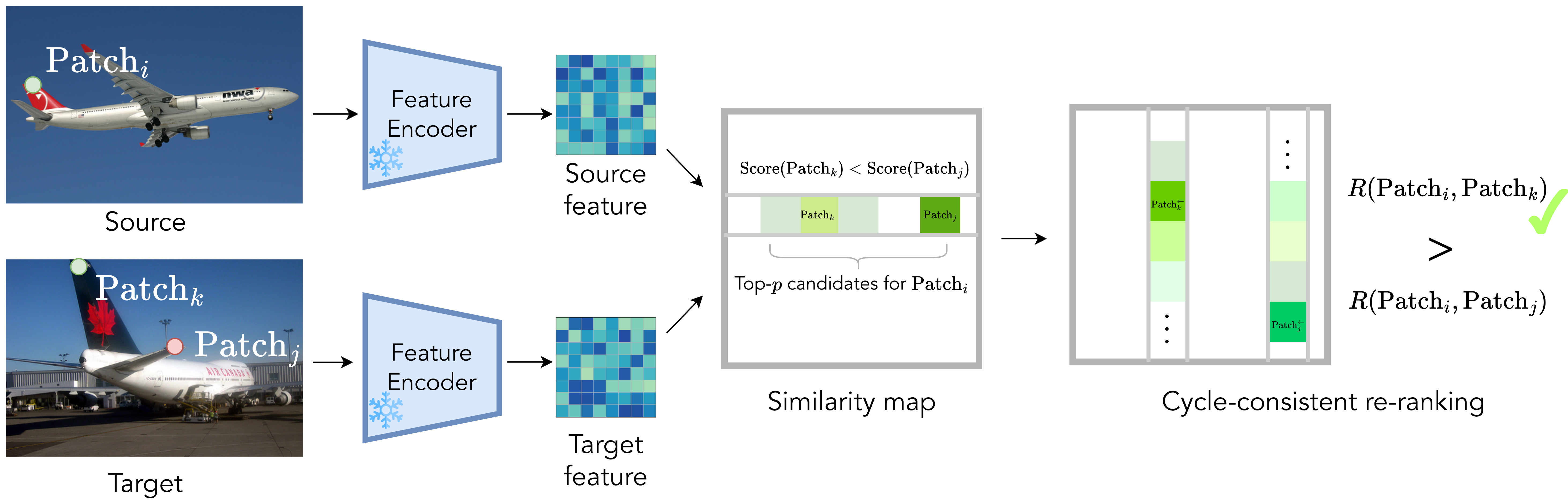}
    \caption{Method overview. Given a query source patch, we apply top-$p$ sampling on the row of the similarity matrix to form a candidate pool (forward pass). For each candidate, we perform a reverse lookup on its column to identify the nearest source patch (backward pass). Each candidate's final score is its forward similarity minus a penalty that grows with the distance between its reverse match and the original query.}
    \label{fig:main_framework}
\end{figure*}

\noindent\textbf{Problem Formulation.}
Let $I_s, I_t \in \mathbb{R}^{H \times W \times 3}$ denote a source image and a target image, respectively. We denote the image domain by $\Omega \subset \mathbb{R}^2$, and view each RGB image as a function $I:\Omega \to \mathbb{R}^3$. We assume access to a pretrained feature extractor $\Phi:\mathbb{R}^{H \times W \times 3}\to\mathbb{R}^{h \times w \times d}$, e.g., DINOv2~\cite{oquab_dinov2_2023}, which partitions each image into a regular $h \times w$ patch grid and produces a $d$-dimensional feature descriptor for each patch. We vectorize the resulting feature maps as $\mathbf{F}_s, \mathbf{F}_t \in \mathbb{R}^{L \times d}$, where $L=hw$ is the number of patches. Each query keypoint $q_i$ is mapped to a flat patch index $\ell \in \{1,\ldots,L\}$ according to the feature grid stride. Given a set of $N$ query keypoints $\mathcal{Q}=\{q_i\}_{i=1}^N$ annotated on the source image $I_s$, where $q_i \in \Omega$, our goal is to predict the corresponding location $\hat{q}_i \in \Omega$ in the target image $I_t$ for each query keypoint $q_i$.  
Our method operates entirely at inference time and requires no additional training beyond $\Phi(\cdot)$ and can be integrated seamlessly into existing semantic correspondence methods. It proceeds in two main stages: (1) top-$p$ candidate nomination, and (2) cycle-consistency re-ranking with combined scoring. \Cref{fig:main_framework} shows an overview of our method. 

\noindent\textbf{Top-$p$ Candidate Nomination.}
To estimate the corresponding location $\hat{q}_i$ in the target image, we first compute patch-wise feature similarities between the source and target. Let $\mathbf{F}_s,\mathbf{F}_t\in\mathbb{R}^{L\times d}$ denote the $\ell_2$-normalized source and target feature matrices. We compute the cosine similarity matrix as
\begin{align}
    \mathbf{S}=\mathbf{F}_s\mathbf{F}_t^\top\in \mathbb{R}^{L\times L}, \label{eq:similarity_matrix}
\end{align}
where $\mathbf{S}_{\ell m}$ denotes the similarity between source patch $\ell$ and target patch $m\in\{1,\ldots,L\}$. Given a query keypoint $q_i$, let $\ell_i\in\{1,\ldots,L\}$ denote its corresponding source patch index. A standard nearest-neighbor matching predicts by greedily selecting the most similar target patch:
\begin{align}
    \hat{m}_i^{\mathrm{NN}}=\argmax_{m\in\{1,\ldots,L\}}\mathbf{S}_{\ell_i m}.
    \label{eq:nn_prediction}
\end{align}
The predicted target keypoint can then be obtained by mapping $\hat{m}_i^{\mathrm{NN}}$ back to image coordinates. However, this one-shot strategy is brittle in practice because it relies only on the top-$1$ candidate. Foundation-model descriptors are often semantic but imperfectly localized; therefore, the true correspondence may not be the highest-scoring target patch, even when it appears among the most plausible candidates. This is especially common when multiple target patches share similar semantic appearance or when the top similarity scores are nearly flat. Our analysis in \Cref{fig:teaser} shows that the correct match is substantially more likely to appear within a set of top-ranked candidates than to be selected by the standard nearest neighbor, motivating a \textit{recall-then-rerank} strategy.

To get a high-recall candidate set, we propose an adaptive top-$p$ criterion inspired by nucleus sampling~\cite{holtzman_curious_2019}. Instead of nominating a fixed number of top-$k$ candidates for every query, we convert query-specific similarity vector into a temperature-scaled probability distribution over target patches:
\begin{align}
    \boldsymbol{\alpha}_i=\mathrm{softmax}\left(\frac{\mathbf{S}_{\ell_i,:}}{\tau}
    \right)\in \mathbb{R}^{L},
    \label{eq:softmax_distribution}
\end{align}
where $\tau>0$ is the temperature parameter and $\alpha_{im}$ denotes the probability assigned to target patch $m$ for query $q_i$. Let
$\alpha_{i,(1)}\geq\alpha_{i,(2)}\geq\cdots\geq\alpha_{i,(L)}$
denote the sorted probabilities in descending order, and let $m_{i,k}$ denote the target patch index corresponding to the $k$-th largest probability $\alpha_{i,(k)}$. Given a nucleus threshold $p\in(0,1]$, we define the adaptive candidate count as
\begin{align}
    K_i=\min\left\{K\in\{1,\ldots,L\}:\sum_{k=1}^{K}\alpha_{i,(k)}\geq p\right\}.
    \label{eq:topp_candidate_count}
\end{align}
To avoid degenerate candidate sets, we further clamp the candidate count as
\begin{align}
    \tilde{K}_i=\min\left(\max(K_i,K_{\min}),K_{\max}\right).
    \label{eq:clamped_candidate_count}
\end{align}
The resulting top-$p$ candidate set for query $q_i$ is
\begin{align}
    \mathcal{C}_i=\left\{m_{i,1},m_{i,2},\ldots,m_{i,\tilde{K}_i}\right\}.
    \label{eq:topp_candidate_set}
\end{align}
This adaptive nomination strategy assigns fewer candidates to confident queries with sharply peaked similarity distributions, while allowing more candidates for ambiguous cases. The nominated set $\mathcal{C}_i$ therefore improves recall before the subsequent re-ranking stage selects the final correspondence.

\noindent\textbf{Cycle-Consistent Re-ranking.}
The top-$p$ nomination stage produces a high-recall candidate set $\mathcal{C}_i$ for each query keypoint $q_i$. Since choosing the candidate with the highest forward source-to-target similarity can be unreliable, we introduce a training-free cycle-consistency criterion to re-rank the nominated candidates. This allows us to exploit the bijective structure of correspondence by favoring matches that are consistent under both forward and reverse feature-space queries.

For each nominated target candidate $m_{i,k}\in\mathcal{C}_i$, we compute its reverse nearest neighbor in source as
\begin{align}
    \ell_{i,k}^{\leftarrow}=\argmax_{\ell\in\{1,\ldots,L\}}\mathbf{S}_{\ell m_{i,k}},
    \label{eq:reverse_nn}
\end{align}
where $\ell_{i,k}^{\leftarrow}$ denotes the source patch that best matches the target candidate $m_{i,k}$. A cycle-consistent candidate should map back to a source patch close to the original source patch $\ell_i$ associated with query $q_i$. To measure this consistency, let
$\rho:\{1,\ldots,L\}\to\{1,\ldots,h\}\times\{1,\ldots,w\}$
map a flat patch index to its two-dimensional coordinate on the $h\times w$ feature grid. We define the cycle distance for the $k$-th candidate of query $q_i$ as
\begin{align}
    d_{\mathrm{cyc}}(i,k) = \left\|\rho\left(\ell_{i,k}^{\leftarrow}\right)-\rho(\ell_i)\right\|_2.
    \label{eq:cycle_distance}
\end{align}
Intuitively, a correct candidate should yield a small cycle distance, while an incorrect candidate may have high forward similarity but map back to a distant source patch, incurring a larger cycle-consistency penalty. We combine forward similarity and cycle-consistency into a unified score:
\begin{align}
    R_{i,k}=\mathbf{S}_{\ell_i m_{i,k}}-\lambda\, d_{\mathrm{cyc}}(i,k),
    \label{eq:reranking_score}
\end{align}
where $\lambda\geq 0$ controls the strength of the penalty, which we treat as a hyperparameter. The final candidate rank and the predicted patch index are obtained as
\begin{align}
    \hat{k}_i=\argmax_{k\in\{1,\ldots,\tilde{K}_i\}}R_{i,k},\qquad \hat{m}_i=m_{i,\hat{k}_i}.
    \label{eq:final_candidate_rank}
\end{align}
Finally, the predicted target patch index $\hat{m}_i$ is mapped back to image coordinates to obtain the target keypoint prediction $\hat{q}_i\in\Omega$. This re-ranking step implements our \textit{recall-then-rerank} design: top-$p$ nomination preserves plausible matches, while cycle consistency selects the candidate that remains both semantically similar and self-consistent under reverse target-to-source matching.


%% file: sections/4_experiments.tex
\section{Experiments}
\label{sec:exp}
\noindent\textbf{Datasets and Metrics.}
We evaluate on two public semantic correspondence benchmarks: PF-Pascal~\cite{ham_proposal_2016} and SPair-71k~\cite{min_spair-71k_2019}, using their default splits. For preprocessing, we follow prior protocols~\cite{zhang_telling_2024,stracke_cleandift_2025,dunkel_it_2025}: each image is resized to $840 \times 840 \times 3$ while preserving its aspect ratio, and deep features are extracted from the resized images. We also report results on the geometry-aware SPair-71k test split following~\cite{zhang_telling_2024}, which evaluates robustness under challenging appearance, scale and viewpoint variations. We use Percentage of Correct Keypoints (PCK) as the metric, measuring the fraction of predicted keypoints that fall within a normalized distance threshold from the ground-truth keypoints. Following prior work~\cite{zhang_telling_2024,stracke_cleandift_2025,dunkel_it_2025}, we normalize the threshold by the object bounding-box size. 
Formally, for a threshold $\alpha$, 
\begin{align}
\label{eq:pck}
    \text{PCK}(\alpha) = \frac{1}{N}\sum_{i=1}^{N} \mathbf{1}\Big[ \| \hat{p}_i - p_i \|_2 \leq \alpha \cdot L_i \Big],
\end{align}
where $N$ is the total number of annotated keypoints, $p_i$ is the ground-truth location of the $i$-th keypoint, $\hat{p}_i$ is the predicted correspondence, and $\| \cdot \|_2$ denotes Euclidean distance. The tolerance threshold is given by $\alpha \cdot L_i$, where $\alpha \in (0,1]$ is a normalized error tolerance and $L_i$ is a reference object scale (taken as the maximum of the object’s width and height). A prediction is considered correct if the error is below this threshold. 
Additional details are provided in the appendix.

\noindent\textbf{{Baselines}.}
Our proposed method can be seamlessly integrated into existing semantic correspondence pipelines. We compared its effectiveness against several representative baselines: (1) TLR~\cite{zhang_telling_2024}, (2) CleanDIFT~\cite{stracke_cleandift_2025}, (3) DIY-SC~\cite{dunkel_it_2025}. Throughout, TLR refers to its unsupervised variant; its supervised variant appears only as a reference point (\Cref{fig:teaser,tab:topk_backbones}) and in the appendix (\Cref{tab:supervised}). We follow the official implementations provided by these methods. Because our module changes only the decoding step, every baseline is run with its released refinement stack fully enabled: \textbf{adaptive pose alignment}~\cite{zhang_telling_2024,stracke_cleandift_2025} and \textbf{window soft-argmax} sub-patch refinement are active in every number we report, for the baseline and for our method alike. All reported gains are therefore measured \textit{on top of} these two components rather than against them. These are complementary refinements rather than competing ones: window soft-argmax refines the location of a match within a patch neighborhood, whereas our re-ranking changes which patch is selected. \Cref{tab:composition} measures each component in isolation and in composition, and shows that our module adds more when window soft-argmax is disabled (+1.83) than when it is enabled (+1.37).

\noindent\textbf{{Implementation}.}
All experiments are implemented in PyTorch with CUDA 11.7 and run on NVIDIA A5000 GPUs with 24 GB memory. For top-$p$ candidate generation, we set $p=0.9$, $\tau=0.1$, $K_{\min}=1$, and $K_{\max}=30$. We treat $\lambda$ as a hyperparameter and analyze its sensitivity for each baseline in the following sections; sensitivity to $\tau$, $K_{\min}$ and $K_{\max}$ is reported in \Cref{tab:hyperparam_sweep}. The same values are used for every dataset, base method and supervision regime; we do not tune per benchmark.

For the analyses of the re-ranking criterion itself (\Cref{tab:cycle_rules} and \Cref{appendix:analysis}) we evaluate on the \textit{full} SPair-71k test set, comprising 12{,}234 pairs and 88{,}328 visible keypoints, and all variants are scored against the same similarity matrix in a single pass so that only the scoring rule differs between rows. In this setting our own run of the base TLR pipeline scores 69.79 PCK@0.10 against the 69.4 reported in \cite{zhang_telling_2024}; the difference comes from the mask-constrained candidate nomination described in \Cref{subsec:implement}, and is isolated in \Cref{tab:mask_ablation}. Holding it fixed across all rows makes the reported $\Delta$ attributable to the cycle rule alone. More details can be found in the appendix.

%% file: sections/5_results.tex
\section{Results}
\label{sec:results}
\noindent\textbf{Quantifying the Decoding Bottleneck.}
\Cref{fig:teaser} reports top-$k$ recall for SD+DINO. \Cref{tab:topk_backbones} extends it to three descriptor families and three PCK thresholds, each measured against the top-$1$ accuracy of a supervised reference. Let $k^\star$ be the smallest $k$ at which top-$k$ recall exceeds that reference. All nine combinations cross: seven by $k^\star{=}10$ and the remaining two by $k^\star{=}30$, with the weakly supervised family (DIY-SC) crossing earliest. The bottleneck is therefore a property of the descriptor class, not of one extractor. \Cref{appendix:analysis} extends this measurement to supervised features, where the same gap holds.

\input{tables/topk_backbones}

\noindent\textbf{Main Results.}
We first present quantitative results in \Cref{tab:spair_percat_perkpts_pck010}, which reports per-category PCK@0.10 on SPair-71k using the per-keypoint protocol. Our method consistently improves all base methods when applied directly at test time. Among unsupervised (US) methods, it improves TLR by \textbf{+1.8} points and CleanDIFT by \textbf{+1.6} points on average. The gains span both rigid classes, such as bottle and train, and deformable classes, such as dog and person, suggesting that cycle consistency provides a broadly useful geometric signal. Even when applied to the stronger weakly supervised (WS) DIY-SC baselines, which are fine-tuned on 3D-aware pseudo-labels, our method still yields consistent improvements of +0.3 and +0.5 points in the DINO and SD+DINO settings, respectively. These results indicate that our method is not tied to a specific feature backbone or supervision regime, but instead improves the final decoding step shared by different correspondence pipelines.

\input{tables/spair_percat_perkpts_pck010}

\begin{figure*}
    \centering
    \begin{tabular}{@{}c@{\hspace{0mm}}c@{}}
        \includegraphics[width=0.49\textwidth]{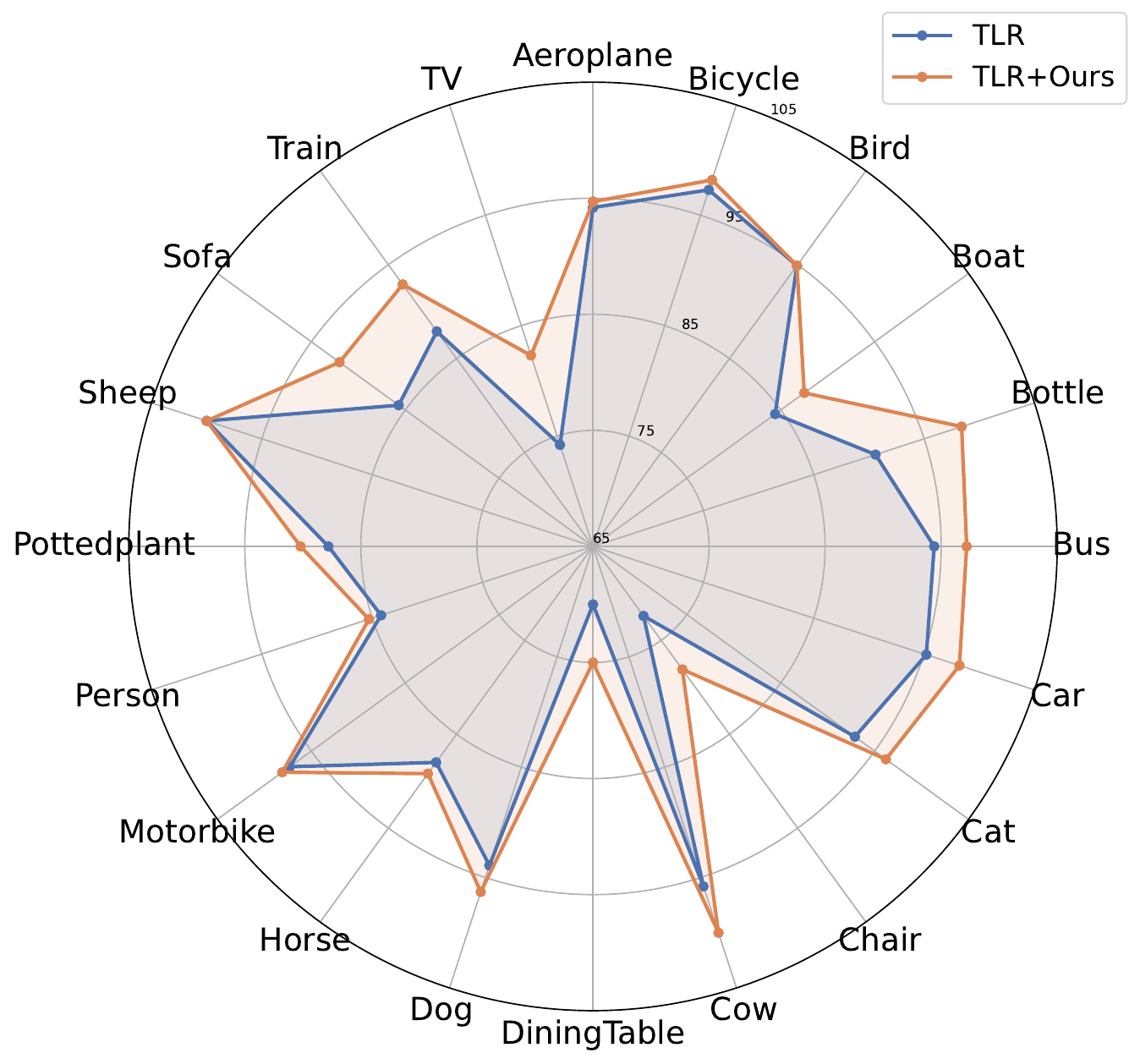} & \includegraphics[width=0.49\textwidth]{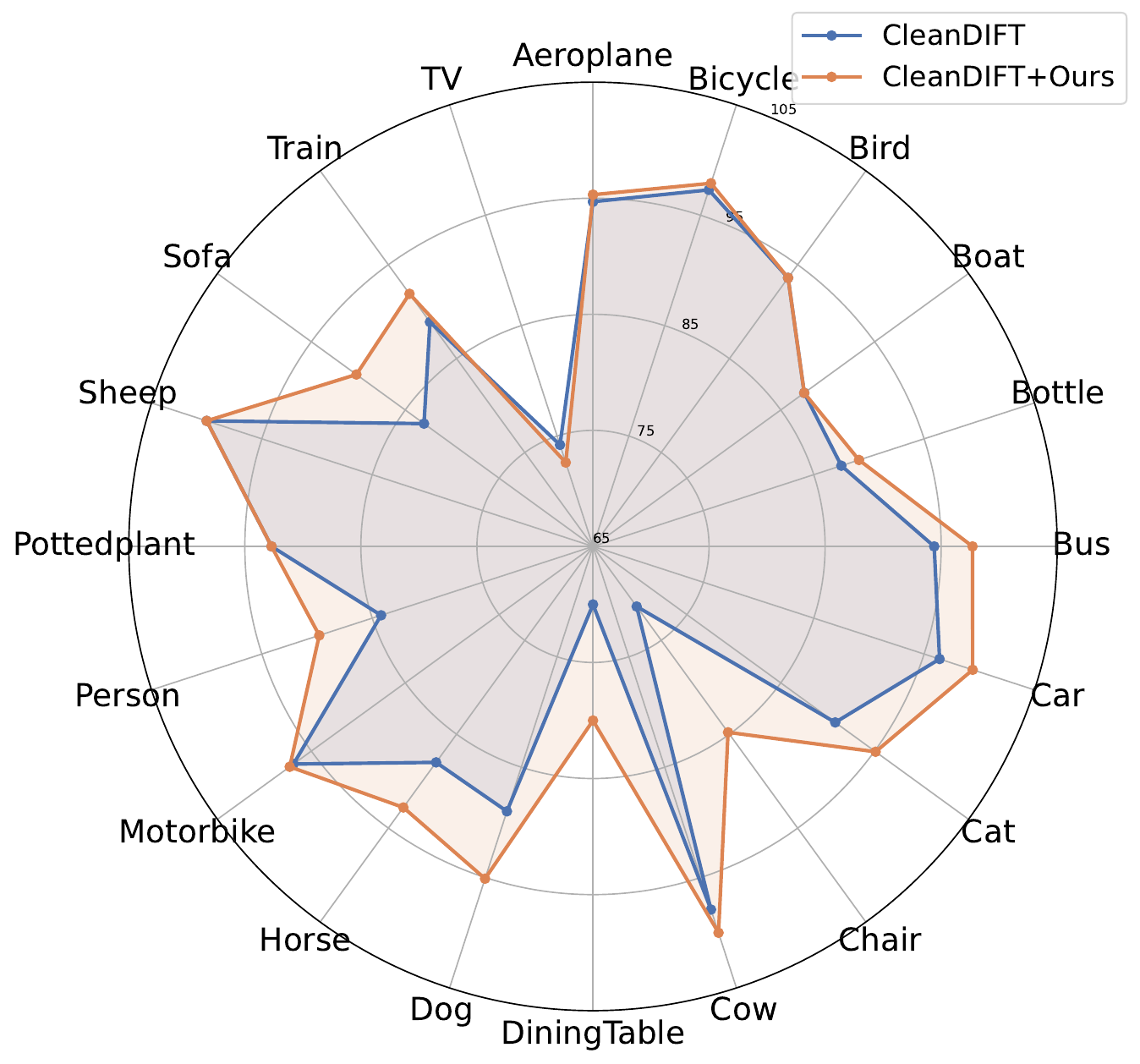} \\
    \end{tabular}
    \caption{\textbf{Per-category radar plots on PF-Pascal.} Each axis represents one object category, and values denote PCK@0.15 (\textit{per-keypoint}). The filled area shows our method applied to TLR (left) and CleanDIFT (right); the outline shows the respective baseline. Our method consistently expands the coverage across the majority of categories.}
    \label{fig:radar_pascal}
\end{figure*}

\begin{figure*}[tb]
    \centering
    \begin{tabular}{@{}c@{\hspace{0mm}}c@{\hspace{1mm}}|c@{\hspace{0mm}}c@{}}
         \includegraphics[width=0.24\linewidth]{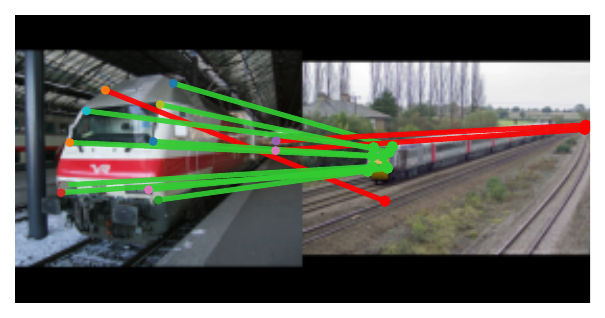}&  \includegraphics[width=0.24\linewidth]{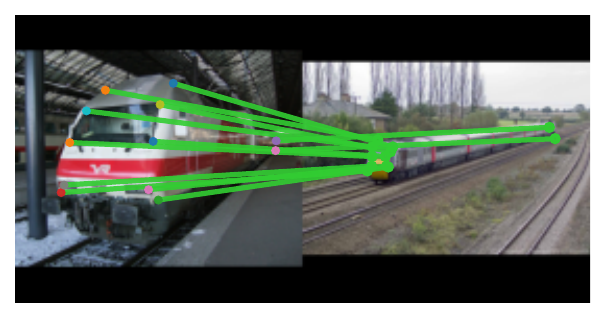} & \includegraphics[width=0.24\linewidth]{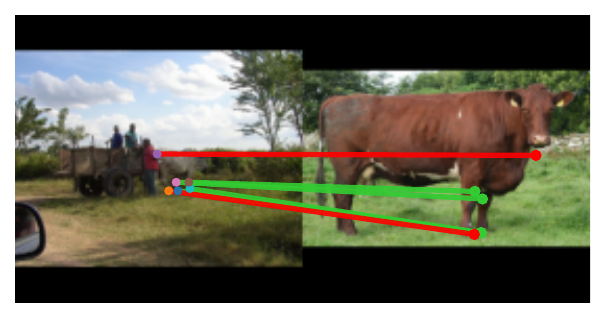}&  \includegraphics[width=0.24\linewidth]{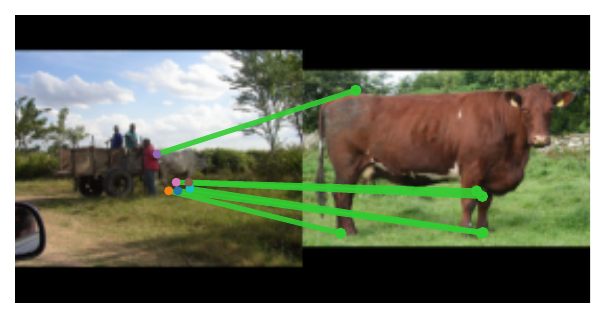} \\
         {\footnotesize TLR} & {\footnotesize TLR+Ours} & {\footnotesize TLR} & {\footnotesize TLR+Ours}\\

         \includegraphics[width=0.24\linewidth]{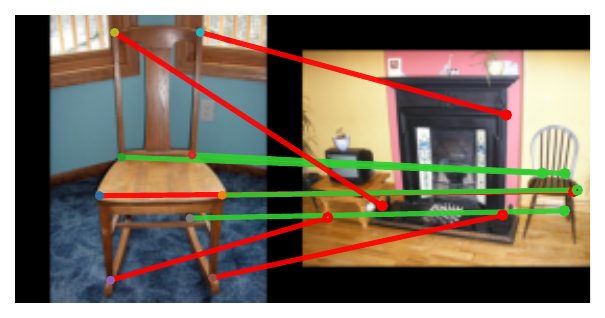}&  \includegraphics[width=0.24\linewidth]{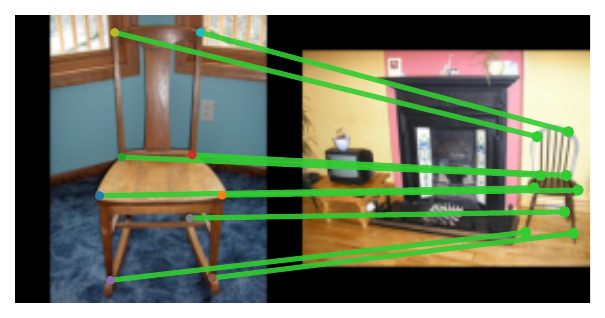} & \includegraphics[width=0.24\linewidth]{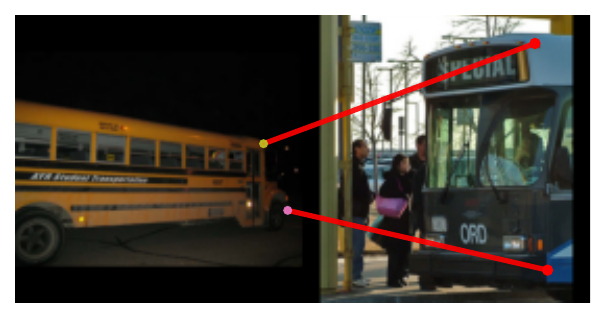}&  \includegraphics[width=0.24\linewidth]{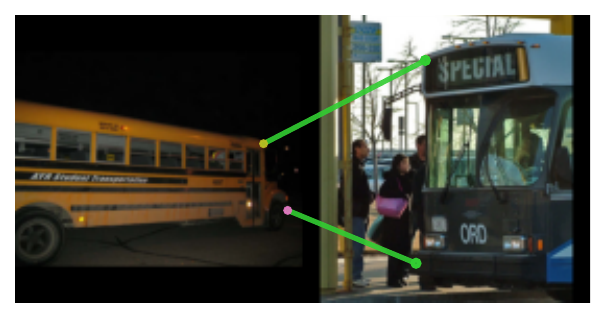} \\
         {\footnotesize CleanDIFT} & {\footnotesize CleanDIFT+Ours} & {\footnotesize CleanDIFT} & {\footnotesize CleanDIFT+Ours}\\

         \includegraphics[width=0.24\linewidth]{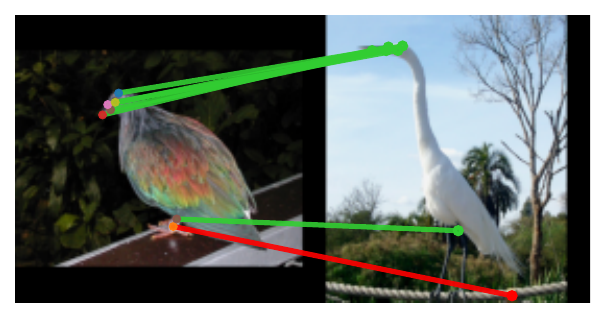}&  \includegraphics[width=0.24\linewidth]{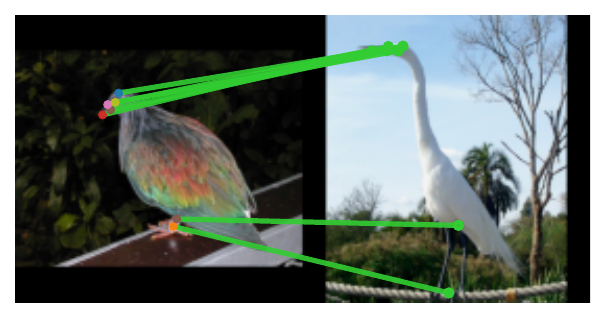} & \includegraphics[width=0.24\linewidth]{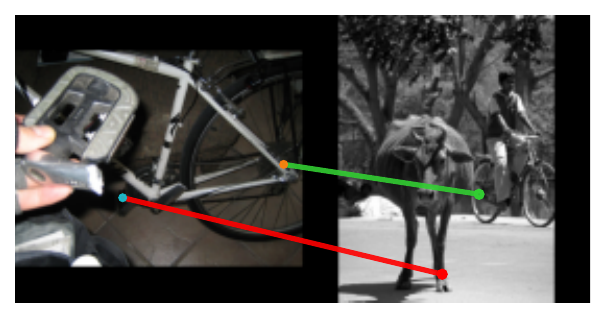}&  \includegraphics[width=0.24\linewidth]{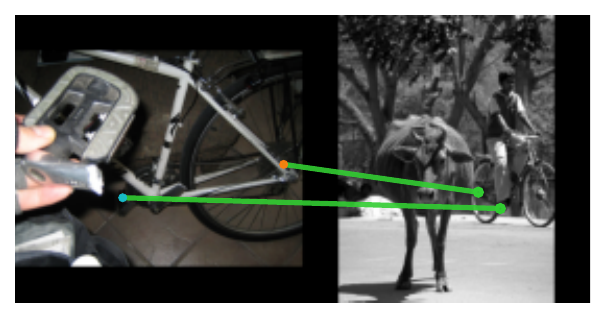} \\
         {\footnotesize DIY-SC} & {\footnotesize DIY-SC+Ours} & {\footnotesize DIY-SC} & {\footnotesize DIY-SC+Ours}\\
    \end{tabular}
    \caption{\textbf{Qualitative comparison on SPair-71k.} For each image pair, correspondences are shown as colored lines: green for correct matches and red for incorrect ones (PCK@0.1). We apply our cycle-consistent re-ranking to three base methods (TLR, CleanDIFT, DIY-SC). Across all three baselines and diverse object categories, our method consistently reduces incorrect matches.}
    \label{fig:main_results_spair}
\end{figure*}

\input{tables/perimg_allpck}
\Cref{tab:perimg_allpck} reports per-image results across multiple PCK thresholds on SPair-71k, its geometric subset, and PF-Pascal. Our method improves performance consistently across all benchmarks and thresholds. The gains are especially pronounced on the geometric subset of SPair-71k, which is designed to stress-test correspondence methods under large viewpoint and scale changes. At PCK@0.10, our module improves TLR by +1.8 points and CleanDIFT by +1.9 points, confirming the benefit of cycle consistency in geometrically challenging cases. On PF-Pascal, the improvements are even larger, reaching \textbf{+3.2} points for TLR and \textbf{+3.4} points for CleanDIFT at PCK@0.15. \Cref{fig:radar_pascal} further shows that the gains on PF-Pascal cover most object categories rather than being driven by a few outliers. \Cref{fig:main_results_spair} provides qualitative comparisons on SPair-71k across all three base methods. Our re-ranking step consistently refines incorrect matches (red) and produces more geometrically coherent correspondences (green). Additional per-keypoint results, per-category breakdowns, and qualitative comparisons for all base methods on both datasets are provided in the appendix.

\begin{figure*}
    \centering
    \begin{tabular}{@{}c@{\hspace{1mm}}c@{}}
         \includegraphics[width=0.493\linewidth]{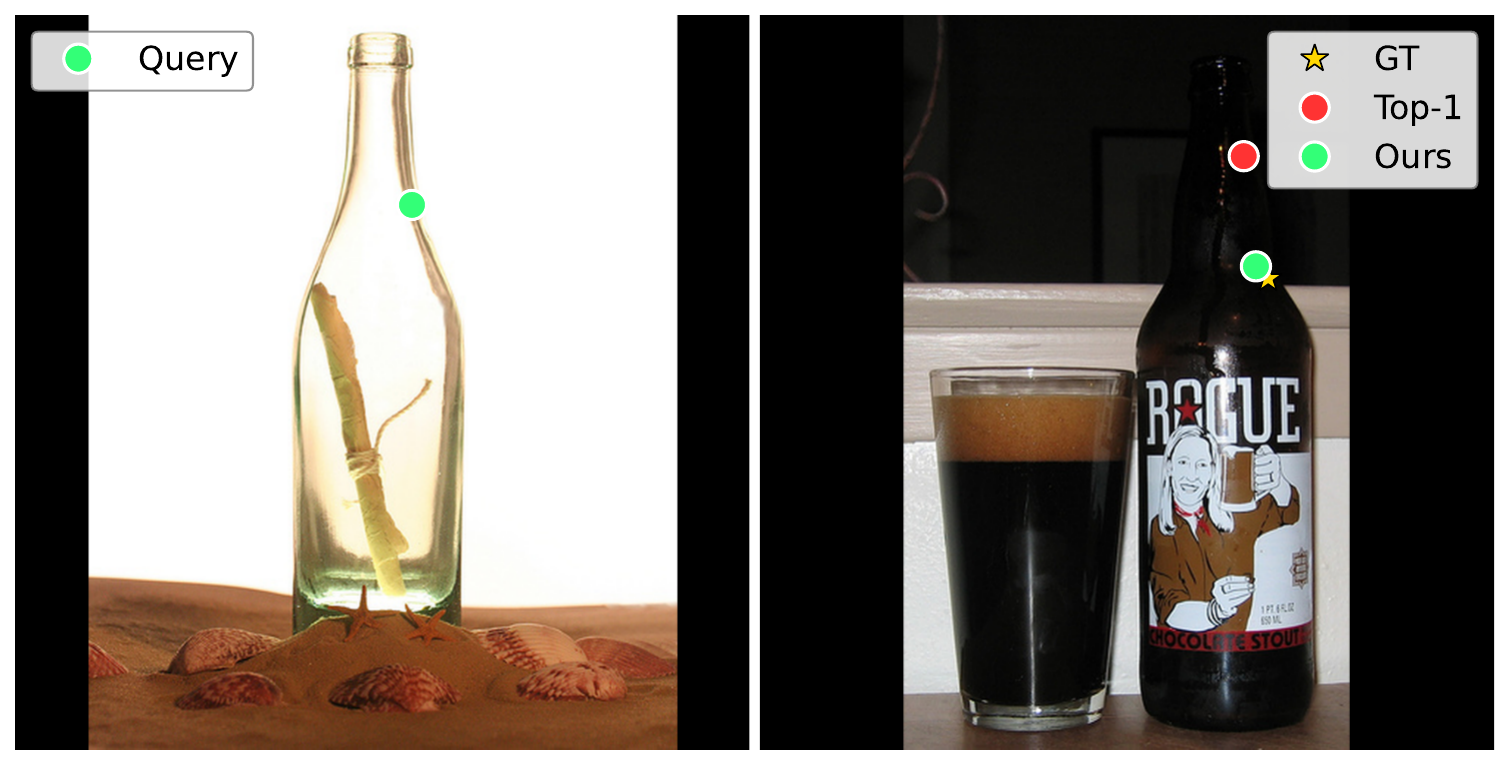} & \includegraphics[width=0.493\linewidth]{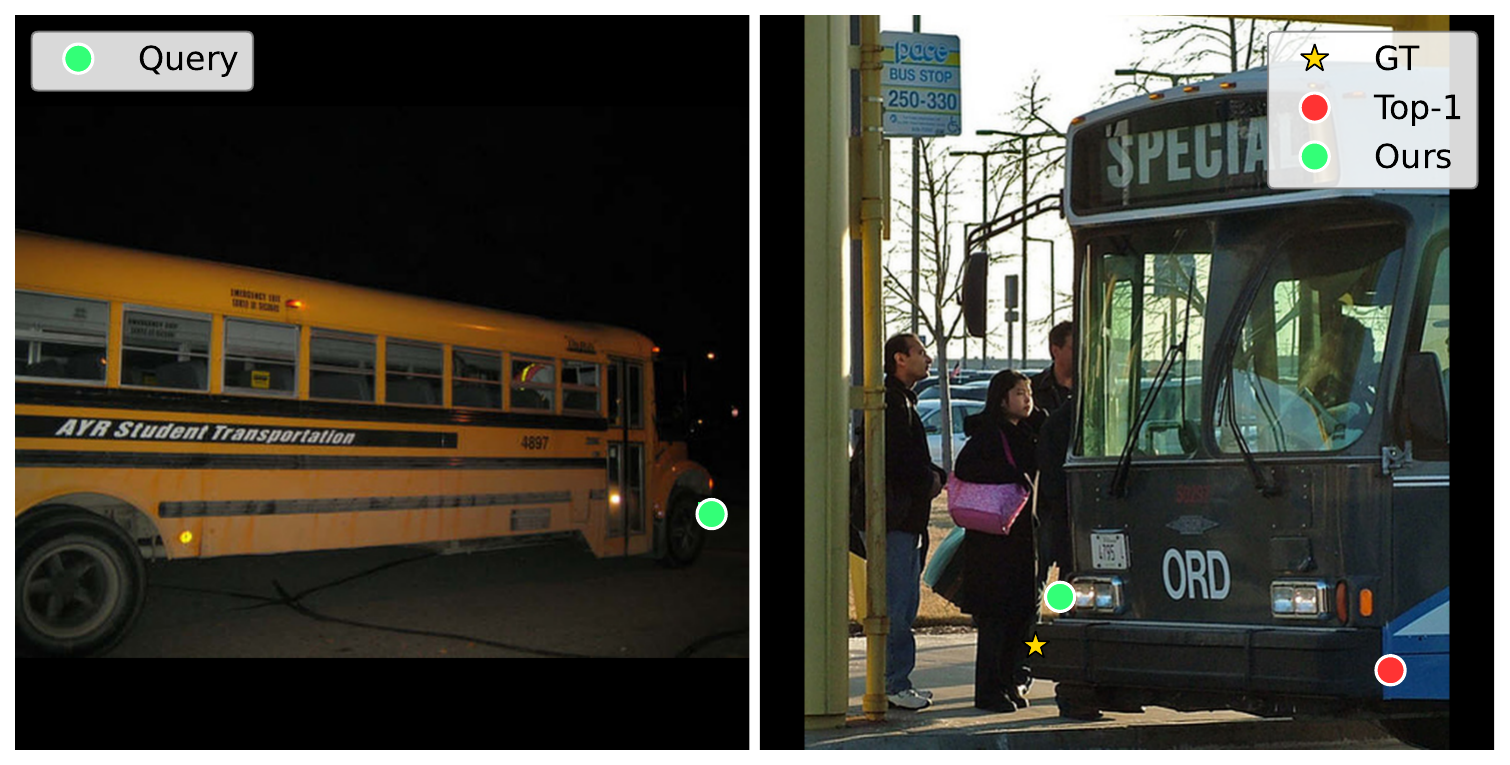}\\
         \includegraphics[width=0.493\linewidth]{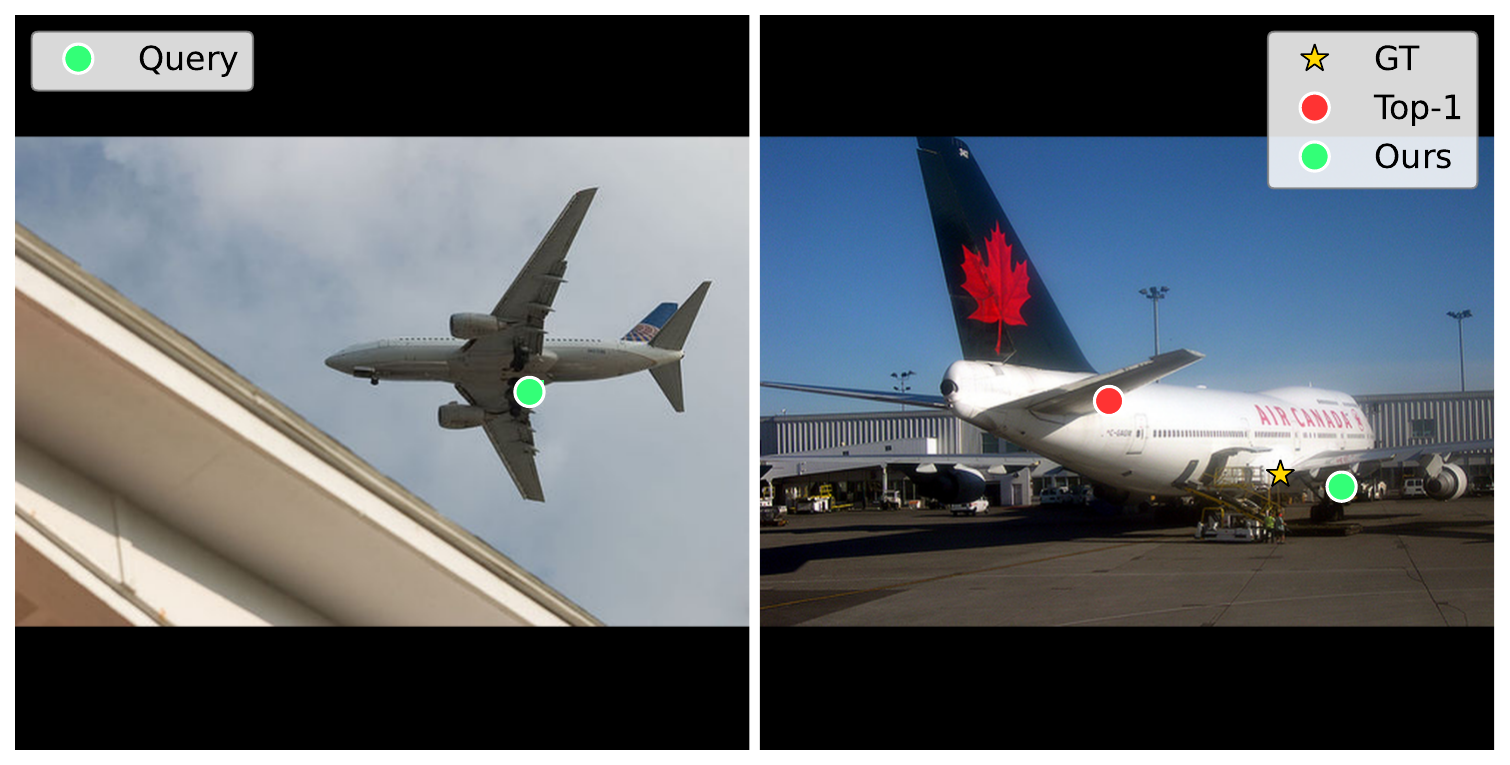} & \includegraphics[width=0.493\linewidth]{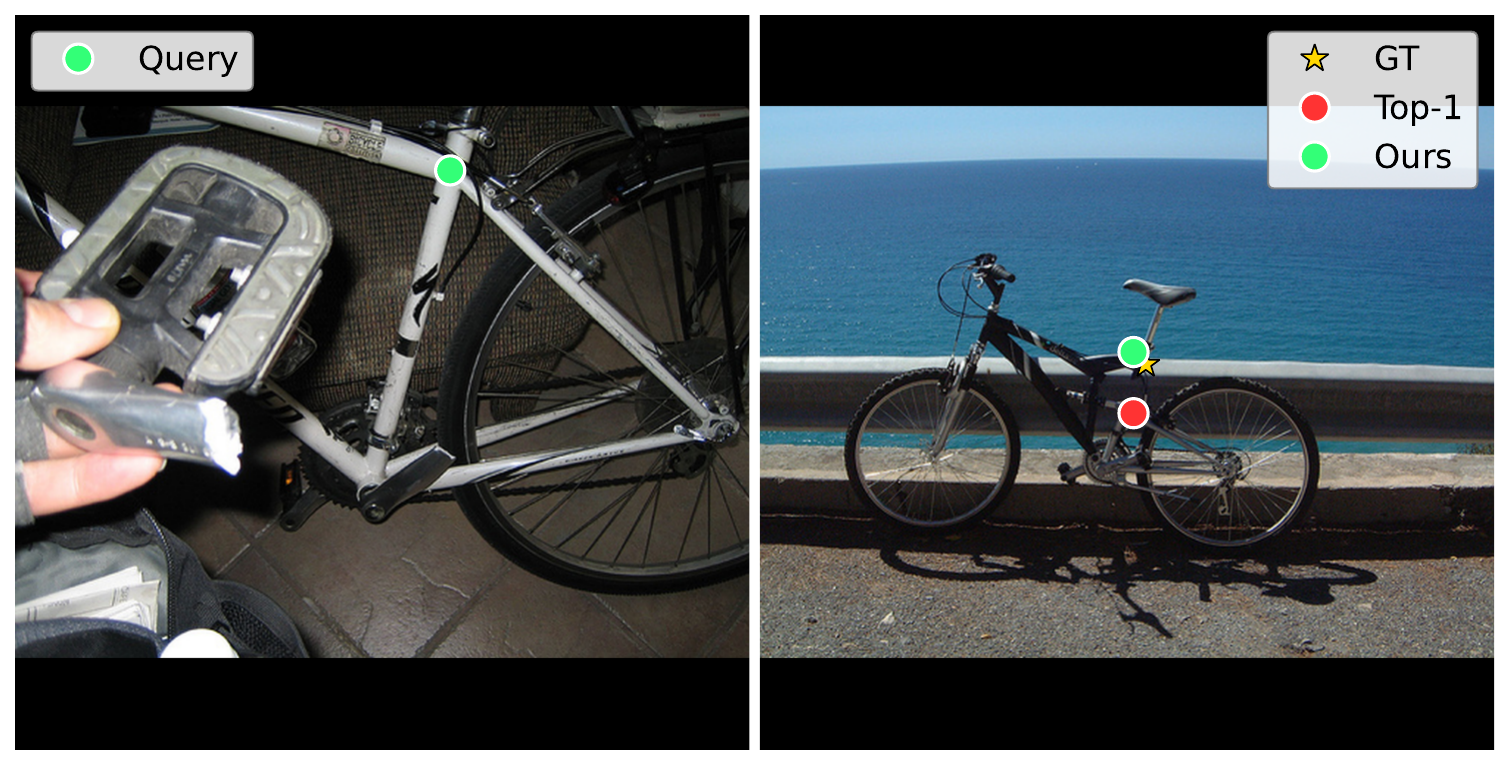}
    \end{tabular}
    \caption{\textbf{Qualitative examples showing the benefit of cycle-consistent re-ranking on SPair-71k.} For each pair, the left image marks the source keypoint and the right image shows three locations: ground truth (GT), the top-1 nearest-neighbor prediction, and our method's prediction. While top-1 matching selects an incorrect target point, our cycle-consistent re-ranking successfully recovers the correct correspondence from the top-$p$ candidate pool.}
    \vspace{10pt}
    \label{fig:top_k_main}
\end{figure*}

\noindent\textbf{Effectiveness of Cycle-Consistent Re-ranking.}
A key motivation for our method is that the top-$1$ nearest-neighbor prediction is often incorrect even when the correct match is reachable within a small candidate pool. \Cref{fig:top_k_main} directly illustrates this: across different object categories, the top-$1$ prediction (right image) falls far from the ground truth, yet our cycle-consistent re-ranking successfully identifies the correct correspondence from the same top-$p$ candidate pool. This demonstrates that the failure lies not in the recall of the feature descriptors, but in the decoding step. By penalizing candidates whose reverse match lands far from the source keypoint, our method suppresses false positives that happen to have high forward similarity but are geometrically inconsistent. Additional examples across more categories are provided in the appendix.

\noindent\textbf{Ablation Study.}
\Cref{tab:ablation} ablates our method across all three base methods. The full method (Top-$p$+Cyc) improves every base method at every threshold, raising PCK@0.10 from 67.9 to 70.1 on TLR, from 67.6 to 69.6 on CleanDIFT, and from 75.5 to 76.3 on DIY-SC. On TLR we also replace the adaptive top-$p$ pool with a fixed top-$k$ pool: both reach 70.1 PCK@0.10, and top-$p$ is slightly better at the stricter thresholds (10.9 vs.\ 10.8 at PCK@0.01 and 53.8 vs.\ 53.6 at PCK@0.05).

\input{tables/combined_ablation_alternative_formulation}

\noindent\textbf{Comparison to Standard Cycle-Consistency Rules.}
Cycle consistency is a long-established constraint, so the question is not whether it helps but whether the \textit{standard} form of it suffices once transplanted to test-time decoding. \Cref{tab:cycle_rules} answers this by holding the candidate pool fixed and varying only the rule applied to the cycle distance $d_{\mathrm{cyc}}$. The exact mutual-nearest-neighbor test, which rejects any candidate with $d_{\mathrm{cyc}}>0$, recovers $+0.28$; relaxing it to a tolerance of one patch triples this to $+0.85$; our graded penalty reaches $+1.37$. The gain grows monotonically with the spatial tolerance of the rule. The binary in-score variant, which subtracts $\lambda\,\mathbf{1}[d_{\mathrm{cyc}}>0]$ at exactly the point in the score where we subtract $\lambda\,d_{\mathrm{cyc}}$, reaches only $+0.42$, which rules out the alternative explanation that what matters is \textit{where} the term is applied rather than \textit{how graded} it is.

\input{tables/cycle_rules}

The graded penalty helps more mainly because it acts on more keypoints. Exact MNN changes the prediction for only 4.9\% of keypoints, while our penalty reorders the whole candidate pool and changes 13.7\%; when either rule does change a prediction, it is right a similar fraction of the time (win rates of 55.4\% and 60.7\%, \Cref{tab:winloss} in the appendix). Exact MNN also carries little information: a perfect round trip ($d_{\mathrm{cyc}}=0$) is almost equally common for correct and incorrect top-$1$ matches (24.5\% vs.\ 24.6\%), whereas the distance itself does separate them. We report the same criterion under Keypoint Average Precision in \Cref{tab:kap} of the appendix, a metric designed to penalize exactly the repeated-part and non-bijective matches PCK is blind to; there the improvement is $+4.94$, $3.6\times$ its PCK@0.10 counterpart on identical predictions.

\Cref{tab:alternative_formulation} compares our cycle-consistent re-ranking against two Optimal Transport~\cite{villani_optimal_nodate} (OT)-based re-ranking alternatives, Sinkhorn~\cite{cuturi_sinkhorn_2013} and Partial OT~\cite{figalli_optimal_2010}, using TLR as the base method. Sinkhorn re-ranking improves over the baseline at PCK@0.10 ($69.4$ vs.\ $67.9$), but still underperforms our method ($70.1$), as does Partial OT ($69.7$) and its entropic variant ($68.6$). This suggests that although OT provides a principled global matching formulation, its global one-to-one assignment constraint is not necessarily well aligned with semantic correspondence, where visually similar parts, repeated structures, occlusions, and background regions can make the matching distribution inherently ambiguous. In contrast, our criterion scores each keypoint's candidates independently, by how close their reverse match lands to the source keypoint, providing a simpler and more localized re-ranking signal without requiring a global one-to-one assignment, transport regularization or mass constraints.

\noindent\textbf{Complexity Analysis.}
Our method introduces negligible overhead over the base correspondence method. The similarity matrix $\mathbf{S}$ is already computed and stored by the base method; our module only reuses this matrix at test time. The additional cost comes from two steps. First, top-$p$ candidate nomination retrieves candidates from each query row of length $L$, where $L$ is the number of target patches. Since we cap the candidate pool by a fixed constant $K_{\max}$, this step can be implemented with a partial top-$K_{\max}$ selection rather than a full sort, giving $\mathcal{O}(L\log K_{\max})$ time per keypoint, or effectively linear in $L$ for fixed $K_{\max}$. Second, cycle-consistent re-ranking evaluates only the nominated candidates $\mathcal{C}_i$, with $|\mathcal{C}_i|\leq K_{\max}$. Thus, after nomination, both the number of candidates scored and the memory required per keypoint are bounded by a constant independent of patch count. Computing the reverse match requires scanning the corresponding column of $\mathbf{S}$. The re-ranking stage operates on a fixed-size candidate set and adds minimal overhead in practice.

\begin{figure*}
    \centering
    \begin{tabular}{@{}c@{\hspace{0mm}}c@{\hspace{0mm}}c@{}}
        \includegraphics[width=0.33\textwidth]{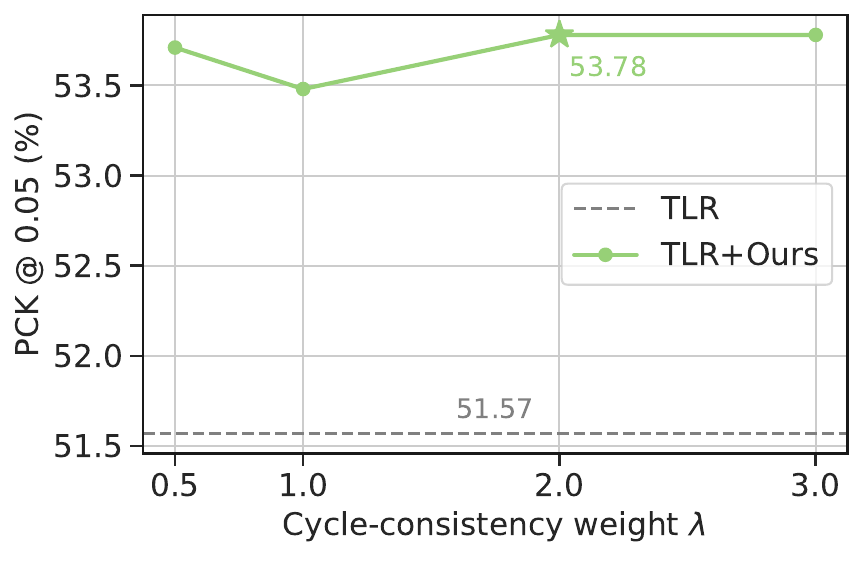} & \includegraphics[width=0.33\textwidth]{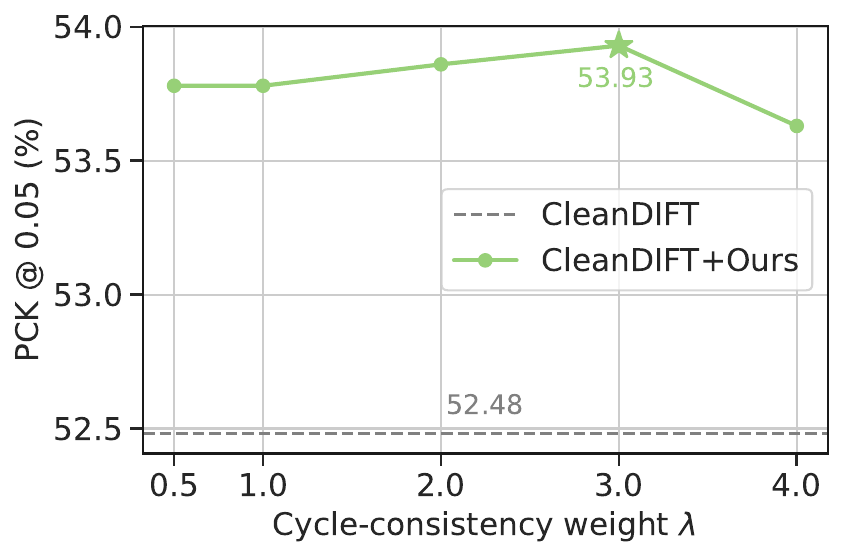} & \includegraphics[width=0.33\textwidth]{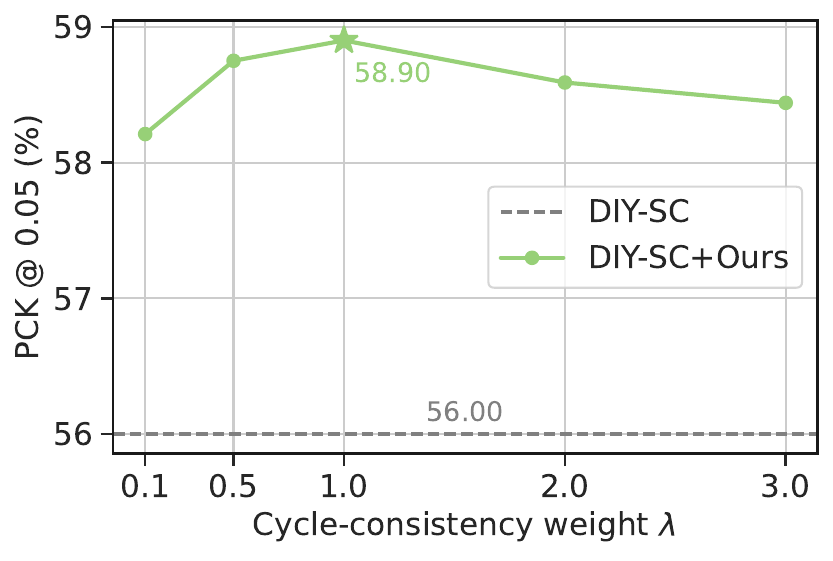}\\
    \end{tabular}
    \caption{\textbf{Sensitivity of the cycle-consistency weight $\lambda$ on SPair-71k (\textit{per-keypoint}, PCK@0.05)} for TLR (left), CleanDIFT (middle), and DIY-SC (right). Performance is stable across a broad range of $\lambda$ values for all three base methods.}
    \label{fig:hyperparam_spair}
\end{figure*}

\noindent\textbf{Hyperparameter Sensitivity.}
Our method has a hyperparameter: the cycle-consistency weight $\lambda$ (Eq.~\ref{eq:reranking_score}), which controls the trade-off between forward similarity and the cycle-distance penalty. \Cref{fig:hyperparam_spair} sweeps $\lambda$ for all three base methods on SPair-71k. Performance is stable across this range: all three base methods consistently improve over their respective baselines for a broad set of $\lambda$ values. This indicates that cycle consistency is a robust signal. The three nomination parameters $\tau$, $K_{\min}$ and $K_{\max}$ act only by setting the size of the candidate pool and do not enter \Cref{eq:reranking_score}, which is why $\lambda$ is the parameter swept here; \Cref{tab:hyperparam_sweep} sweeps the other three and finds a total range of 0.17 PCK@0.10 across every setting. Additional sensitivity analysis on PF-Pascal is provided in \Cref{fig:hyperparam_pascal} and per-category SPair-71k radar plots in \Cref{fig:radar_spair} in the appendix, confirming the same trend across datasets.

\noindent\textbf{Limitation.}
Two limitations bound what the criterion can do. First, a large gap remains between what our re-ranking achieves and the top-$k$ recall of \Cref{fig:teaser}. That curve is an upper bound on what \textit{any} re-ranking could recover, not a claim about what ours recovers: of the 48{,}651 keypoints where the ground-truth patch is nominated into the candidate pool but is not the top-$1$ candidate, our criterion promotes 3{,}457 of them, or 7.1\%. Cycle consistency is a local, training-free signal, and closing the remainder plausibly requires global or pose-level reasoning, or a learned re-ranking score, which we leave for future work. 

Second, cycle consistency distinguishes correct from spurious matches but cannot resolve genuinely mirror-equivalent parts, as both can be cycle-consistent. With TLR, only bird ($90.1 \rightarrow 89.9$) and horse ($75.0 \rightarrow 74.1$) regress among 18 SPair-71k categories, both being symmetric and deformable. Conversely, the largest gains occur where repeated parts are spatially distinguishable: tv ($+7.2$), plant ($+6.3$), and chair ($+4.9$). See \Cref{appendix:failure} for further analysis.

%% file: tables/topk_backbones.tex
\begin{table*}[tb]
    \centering
    \begin{adjustbox}{width=0.8\linewidth}
    \begin{tabular}{@{}llccccc|c@{}}
    \toprule
    Threshold & Features & $k=1$ & $k=3$ & $k=5$ & $k=10$ & $k=30$ & Sup.\ top-$1$ ref.\\
    \midrule
    \multirow{4}{*}{PCK@0.10}
    & SD+DINO \cite{zhang_tale_2023} & 64.5 & 75.3 & 80.7 & 87.3 & \textbf{94.9} & \multirow{4}{*}{87.55}\\
    & CleanDIFT \cite{stracke_cleandift_2025} & 65.0 & 75.7 & 80.8 & 87.5 & \textbf{94.8} & \\
    & DIY-SC \cite{dunkel_it_2025} & 73.7 & 82.3 & 86.3 & \textbf{90.7} & 95.8 & \\
    & TLR \cite{zhang_telling_2024} (supervised) & 87.1 & \textbf{92.3} & 94.2 & 96.4 & 98.4 & \\
    \midrule
    \multirow{4}{*}{PCK@0.05}
    & SD+DINO \cite{zhang_tale_2023} & 49.1 & 63.1 & 69.9 & \textbf{78.4} & 89.2 & \multirow{4}{*}{77.51}\\
    & CleanDIFT \cite{stracke_cleandift_2025} & 49.4 & 63.6 & 70.2 & \textbf{78.7} & 89.3 & \\
    & DIY-SC \cite{dunkel_it_2025} & 57.7 & 70.9 & 77.1 & \textbf{84.1} & 92.1 & \\
    & TLR \cite{zhang_telling_2024} (supervised) & 75.6 & \textbf{85.6} & 89.2 & 93.3 & 97.1 & \\
    \midrule
    \multirow{4}{*}{PCK@0.01}
    & SD+DINO \cite{zhang_tale_2023} & 9.1 & 17.1 & 21.3 & \textbf{27.1} & 36.1 & \multirow{4}{*}{21.96}\\
    & CleanDIFT \cite{stracke_cleandift_2025} & 9.1 & 17.1 & 21.4 & \textbf{27.4} & 36.1 & \\
    & DIY-SC \cite{dunkel_it_2025} & 10.6 & 20.0 & \textbf{24.8} & 31.1 & 40.1 & \\
    & TLR \cite{zhang_telling_2024} (supervised) & 17.0 & \textbf{30.8} & 35.9 & 41.7 & 47.0 & \\
    \bottomrule
    \end{tabular}
    \end{adjustbox}
    \caption{\textbf{Top-$k$ candidate recall across feature extractors on SPair-71k}. For each source keypoint we take the top-$k$ target patches by feature similarity and count the keypoint as recovered if any of them falls within the PCK radius. The reference column is the top-$1$ accuracy of the supervised TLR model on the full test set; the table itself uses 100 pairs per category without sub-patch refinement, so its $k{=}1$ column differs slightly from the reference; \textbf{bold} marks, within each row, $k^\star$: the smallest $k$ at which top-$k$ recall exceeds that reference.}
    \label{tab:topk_backbones}
\end{table*}

%% file: tables/spair_percat_perkpts_pck010.tex
\begin{table*}[tb]
    \centering
    \begin{adjustbox}{width=\textwidth}
    \begin{tabular}{@{}clcccccccccccccccccc|ll@{}}
    \toprule
    & Method & \faIcon{plane} & \faIcon{bicycle} & \faIcon{crow} & \faIcon{ship} & \faIcon{wine-bottle} & \faIcon{bus} & \faIcon{car} & \faIcon{cat} & \faIcon{chair} & \Cow & \faIcon{dog} & \faIcon{horse} & \faIcon{motorcycle} & \faIcon{walking} & \Plant & \Sheep & \faIcon{train} & \faIcon{tv} & Avg \\
    \midrule
    \cmidrule{1-21}
    \parbox[t]{0mm}{\multirow{1}{*}{\rotatebox[origin=c]{90}{\scriptsize \textbf{US}}}}
    & ASIC \cite{gupta_asic_2023} & 57.9 & 25.2 & 68.1 & 24.7 & 35.4 & 28.4 & 30.9 & 54.8 & 21.6 & 45.0 & 47.2 & 39.9 & 26.2 & 48.8 & 14.5 & 24.5 & 49.0 & 24.6 & 36.9 \\
    & DINOv2 \cite{oquab_dinov2_2023} & {{72.7}} & {{62.0}} & {{85.2}} & {{41.3}} & {40.4} & {{52.3}} & {{51.5}} & {71.1} & {36.2} & {67.1} & {{64.6}} & {{67.6}} & {{61.0}} & {{68.2}} & {30.7} & {{62.0}} & {54.3} & {24.2} & {55.6}\\
    & DIFT \cite{tang_emergent_2023} & {63.5} & {54.5} & {80.8} & {34.5} & {46.2} & {52.7} & {48.3} & {77.7} & {39.0} & {76.0} & {54.9} & {61.3} & {53.3} & {46.0} & {57.8} & {57.1} & {71.1} & {63.4} & {57.7}\\
    & SD+DINO \cite{zhang_tale_2023} & {{73.0}} & {{64.1}} & {{86.4}} & {{40.7}} & {{52.9}} & {{55.0}} & {{53.8}} & {{78.6}} & {{45.5}} & {{77.3}} & {{64.7}} & {{69.7}} & {{63.3}} & {{69.2}} & {{58.4}} & {{67.6}} & {{66.2}} & {{53.5}} & {{64.0}}\\
    & SphMap \cite{mariotti_improving_2024} &75.3 & 63.8 & 87.7 & \textbf{48.2} & 50.9 & \textbf{74.9} & \textbf{71.1} & 81.7 & 47.3 & 81.6 & 66.9 & 73.1 & 65.4 & 61.8 & 55.5 & 70.2 & 75.0 & 58.5 & 67.8\\
    & DistillDIFT \cite{fundel_distillation_2024} & 74.6 & 60.4 & 88.7 & 42.5 & 53.5 & 55.0 & 54.6 & 80.8 & 42.7 & 78.6 & \textbf{72.0} & 71.4 & 62.2 & 70.7 & 53.1 & 68.6 & 65.2 & 61.6 & 65.1 \\
    & SoY-FGW \cite{im_shape--you_2026} & 73.5 & 66.7 & \underline{89.9} & 40.8 & \textbf{58.0} & 55.4 & 51.1 & \textbf{84.8} & 52.1 & 81.0 & \underline{71.0} & \underline{75.3} & 64.1 & 71.8 & 62.1 & 70.2 & 70.2 & \textbf{64.4} & 67.9 \\
    & TLR \cite{zhang_telling_2024} & 78.0 & 66.6 & \textbf{90.1} & 44.7 & 54.2 & 66.2 & 60.8 & 82.7 & 53.4 & 82.3 & 69.5 & 75.0 & 66.1 & 72.1 & 59.3 & 71.6 & 83.8 & 54.7 & 69.4\\
    \myrowcolor
    & +Ours & \underline{78.1} & \underline{67.6} & \underline{89.9} & 44.8 & 56.4 & 66.7 & \underline{60.9} & \underline{83.2} & \underline{58.3} & \underline{83.1} & 70.8 & 74.1 & \underline{66.6} & \textbf{73.8} & \underline{65.6} & \underline{71.7} & \textbf{85.9} & 61.9 & \underline{71.2}\textsubscript{\color{ourgreen}$\blacktriangle$1.8}\\
    & CleanDIFT \cite{stracke_cleandift_2025} & 77.9 & 67.1 & 89.7 & 44.0 & 54.5 & \underline{67.2} & \underline{60.9} & 82.7 & 55.1 & 82.9 & 68.7 & \textbf{75.5} & 66.5 & 71.7 & 60.4 & \textbf{72.2} & \underline{83.9} & 57.7 & 69.9\\
    \myrowcolor
    & +Ours & \textbf{78.4} & \textbf{67.7} & 89.3 & \underline{45.9} & \underline{56.6} & 66.8 & 60.7 & 82.9 & \textbf{60.1} & \textbf{83.3} & 69.8 & 74.3 & \textbf{67.0} & \underline{73.5} & \textbf{66.9} & \underline{71.7} & \textbf{85.9} & \underline{64.1} & \textbf{71.5}\textsubscript{\color{ourgreen}$\blacktriangle$1.6}\\
    \midrule\midrule
    \parbox[t]{0mm}{\multirow{1}{*}{\rotatebox[origin=c]{90}{\scriptsize \textbf{WS}}}} & DistillDIFT$^*$ \cite{fundel_distillation_2024} & \textbf{78.2} & 63.8 & \underline{90.1} & 45.0 & 54.6 & 68.0 & 63.7 & 83.2 & 49.3 & 82.6 & \textbf{74.5} & 73.8 & 63.5 & 72.0 & 56.2 & 71.0 & \textbf{86.2} & 66.5 & 70.6\\
    & DIY-SC$^\dagger$ \cite{dunkel_it_2025} & 77.0 & 68.9 & \underline{90.1} & 52.0 & \underline{59.1} & 84.0 & \underline{79.6} & \textbf{88.6} & 52.1 & \textbf{87.5} & 74.1 & \textbf{79.8} & 69.4 & 73.3 & 63.6 & \underline{76.8} & 78.5 & \underline{70.5} & 74.3\\
    \myrowcolor
    & +Ours & 76.7 & 68.9 & 89.4 & 52.4 & \textbf{59.7} & \underline{84.3} & \textbf{80.0} & \underline{88.2} & \underline{54.6} & \underline{87.3} & \underline{74.2} & \underline{79.2} & 69.0 & \underline{74.5} & 65.4 & \textbf{78.1} & \underline{79.0} & \underline{70.5} & \underline{74.6}\textsubscript{\color{ourgreen}$\blacktriangle$0.3}\\
    & DIY-SC$^\ddagger$ \cite{dunkel_it_2025} & \underline{77.2} & \textbf{69.1} & \textbf{90.8} & \underline{54.2} & 57.9 & 83.7 & 77.5 & 86.5 & 53.2 & 86.8 & 73.2 & 78.4 & \textbf{72.5} & 74.2 & \underline{73.4} & 76.1 & 77.2 & 69.5 & 74.4\\
    \myrowcolor
    & +Ours & 76.6 & \underline{69.0} & 89.7 & \textbf{55.4} & 59.0 & \textbf{84.9} & 77.7 & 85.5 & \textbf{56.0} & 87.1 & 72.6 & 77.8 & \underline{72.3} & \textbf{74.6} & \textbf{75.0} & 76.3 & 77.8 & \textbf{71.4} & \textbf{74.9}\textsubscript{\color{ourgreen}$\blacktriangle$0.5}\\
    \bottomrule
    \end{tabular}
    \end{adjustbox}
    \caption{\textbf{Per-category PCK@0.1 scores (\textit{per-keypoint}) on SPair-71k}. DistillDIFT was evaluated in both unsupervised (US) and weakly supervised (WS, noted with $*$) \cite{dunkel_it_2025}. DIY-SC was evaluated in both DINO ($\dagger$) and SD+DINO ($\ddagger$) settings. The best results are bolded and the second best underlined.}
    \label{tab:spair_percat_perkpts_pck010}
\end{table*}


%% file: tables/perimg_allpck.tex
\begin{table*}[tb]
    \centering
    \begin{adjustbox}{width=\textwidth}
    \begin{tabular}{@{}cllll|lll|lll@{}}
    \toprule
    & \multirow{2}{*}{Method} & \multicolumn{3}{c}{SPair-71k} & \multicolumn{3}{c}{SPair-71k (Geo.)} & \multicolumn{3}{c}{PF-Pascal}\\
    \cmidrule(l){3-5} \cmidrule(l){6-8} \cmidrule(l){9-11} 
    & & 0.01 & 0.05 & 0.10 & 0.01 & 0.05 & 0.10 & 0.05 & 0.1 & 0.15\\\midrule
    \parbox[t]{0mm}{\multirow{1}{*}{\rotatebox[origin=c]{90}{\scriptsize \textbf{US}}}}
    & DINOv2+NN \cite{oquab_dinov2_2023} & 6.3 & 38.4 & 53.9 & 3.4 & 28.2 & 42.0 & 63.0 & 79.2 & 85.1\\
    & DIFT \cite{tang_emergent_2023} & 7.2 & 39.7 & 52.9 & 4.6 & 30.0 & 42.5 & 66.0 & 81.1 & 87.2\\
    & SD+DINO \cite{zhang_tale_2023} & 7.9 & 44.7 & 59.9 & 5.3 & 34.5 & 49.3 & 71.5 & 85.8 & 90.6 \\
    & DistillDIFT \cite{fundel_distillation_2024} & 8.0 & 45.4 & 60.8 & - & - & - & - & - & -\\
    & TLR \cite{zhang_telling_2024} & \underline{9.7} & 48.8 & 65.0 & 6.6  & 39.2 & 56.4 & \underline{72.9}  & 85.5 & 90.0\\
    \myrowcolor
    & +Ours & \textbf{9.9}\textsubscript{\color{ourgreen}$\blacktriangle$0.2} & \underline{50.0}\textsubscript{\color{ourgreen}$\blacktriangle$1.2} & \underline{66.3}\textsubscript{\color{ourgreen}$\blacktriangle$1.3} & \textbf{7.1}\textsubscript{\color{ourgreen}$\blacktriangle$0.5} & \underline{41.2}\textsubscript{\color{ourgreen}$\blacktriangle$2.0} & \underline{58.2}\textsubscript{\color{ourgreen}$\blacktriangle$1.8} & \textbf{75.3}\textsubscript{\color{ourgreen}$\blacktriangle$2.4} & \textbf{88.7}\textsubscript{\color{ourgreen}$\blacktriangle$3.2} & \textbf{93.2}\textsubscript{\color{ourgreen}$\blacktriangle$3.2}\\
    & CleanDIFT \cite{stracke_cleandift_2025} & \underline{9.7} & 49.1 & 65.4 & 6.5 & 39.6 & 56.8 & 71.9 & 84.6 & 89.4\\
    \myrowcolor
    & +Ours & \textbf{9.9}\textsubscript{\color{ourgreen}$\blacktriangle$0.2} & \textbf{50.2}\textsubscript{\color{ourgreen}$\blacktriangle$1.1} & \textbf{66.5}\textsubscript{\color{ourgreen}$\blacktriangle$1.1} & \underline{7.0}\textsubscript{\color{ourgreen}$\blacktriangle$0.5} & \textbf{41.6}\textsubscript{\color{ourgreen}$\blacktriangle$2.0} & \textbf{58.7}\textsubscript{\color{ourgreen}$\blacktriangle$1.9} & \textbf{75.3}\textsubscript{\color{ourgreen}$\blacktriangle$3.4} & \underline{88.3}\textsubscript{\color{ourgreen}$\blacktriangle$3.7} & \underline{92.8}\textsubscript{\color{ourgreen}$\blacktriangle$3.4}\\
    \midrule\midrule
    \parbox[t]{0mm}{\multirow{1}{*}{\rotatebox[origin=c]{90}{\scriptsize \textbf{WS}}}}  
    & DistillDIFT$^*$ \cite{fundel_distillation_2024} & 8.9 & 49.8 & 65.3 & - & - & - & - & - & -\\   
    & DIY-SC$^\dagger$ & 9.1 & 51.4 & 71.2 & 5.8 & 41.3 & \underline{63.3} & - & - & -\\
    \myrowcolor 
    & +Ours & 9.9\textsubscript{\color{ourgreen}$\blacktriangle$0.8} & 53.4\textsubscript{\color{ourgreen}$\blacktriangle$2.0} & 71.5\textsubscript{\color{ourgreen}$\blacktriangle$0.3} & \underline{7.0}\textsubscript{\color{ourgreen}$\blacktriangle$1.2} & \underline{44.2}\textsubscript{\color{ourgreen}$\blacktriangle$2.9} & \textbf{63.6}\textsubscript{\color{ourgreen}$\blacktriangle$0.3} & - & - & -\\
    & DIY-SC$\ddagger$ & \underline{10.1} & \underline{53.8} & \underline{71.6} & 6.7 & 42.7 & 63.0 & - & - & -\\
    \myrowcolor
    & +Ours & \textbf{11.0}\textsubscript{\color{ourgreen}$\blacktriangle$0.9} & \textbf{55.5}\textsubscript{\color{ourgreen}$\blacktriangle$1.7} & \textbf{71.7}\textsubscript{\color{ourgreen}$\blacktriangle$0.1} & \textbf{7.9}\textsubscript{\color{ourgreen}$\blacktriangle$1.2} & \textbf{45.3}\textsubscript{\color{ourgreen}$\blacktriangle$2.6} & 63.0 & - & - & -\\
    \bottomrule
    \end{tabular}
    \end{adjustbox}
    \caption{\textbf{PCK scores (\textit{per-image}) across multiple thresholds on SPair-71k, SPair-71k (Geo.), and PF-Pascal}. DistillDIFT was evaluated in both unsupervised (US) and weakly supervised (WS, noted with $*$) settings. DIY-SC was evaluated in both DINO ($\dagger$) and SD+DINO ($\ddagger$) settings. The best results are bolded and the second best underlined.}
    \label{tab:perimg_allpck}
\end{table*}

%% file: tables/combined_ablation_alternative_formulation.tex
\begin{table*}[tb]
    \centering
    \raisebox{-8pt}{%
    \begin{minipage}[t]{0.53\textwidth}
        \centering
        \begin{adjustbox}{width=0.76\linewidth}
        \begin{tabular}{@{}clccc@{}}
        \toprule
        & Method & PCK@0.01 & PCK@0.05 & PCK@0.10 \\
        \midrule
        & Baseline & 10.7 & 51.6 & 67.9\\
        & Top-$k$+Cyc & 10.8 & 53.6 & \textbf{70.1}\\
        & Top-$p$+Cyc & \textbf{10.9} & \textbf{53.8} & \textbf{70.1}\\
        \midrule
        & Baseline & 11.1 & 52.5 & 67.6\\
        & Top-$p$+Cyc & \textbf{11.4} & \textbf{53.9} & \textbf{69.6}\\
        \midrule
        & Baseline & 11.8 & 56.0 & 75.5\\
        & Top-$p$+Cyc & \textbf{12.8} & \textbf{58.9} & \textbf{76.3}\\
        \bottomrule
        \end{tabular}
        \end{adjustbox}
        \captionof{table}{\textbf{Ablation study on SPair-71k (\textit{per-keypoint})} across baselines (top to bottom: TLR, CleanDIFT, DIY-SC). The best results are highlighted in \textbf{bold}.}
        \label{tab:ablation}   
    \end{minipage}
    }
    \hfill
    \begin{minipage}[t]{0.45\textwidth}
        \centering
        \begin{adjustbox}{width=1\linewidth}
        \begin{tabular}{@{}clccc@{}}
        \toprule
        & Method & PCK@0.01 & PCK@0.05 & PCK@0.10 \\
        \midrule
        & Baseline & 10.7 & 51.6 & 67.9\\
        & Sinkhorn & 10.7 & 53.1 & 69.4\\
        & Entropic Partial OT  & 10.2 & 52.3 & 68.6\\
        & Partial OT  & 10.7 & 52.7 & 69.7\\
        & Ours & \textbf{10.9} & \textbf{53.8} & \textbf{70.1}\\
        \bottomrule
        \end{tabular}
        \end{adjustbox}
        \captionof{table}{\textbf{Alternative formulation comparison on SPair-71k (\textit{per-keypoint})} using TLR as the base. We compare OT-based re-ranking (Sinkhorn, Partial OT) against ours. The best results are highlighted in \textbf{bold}.}
        \label{tab:alternative_formulation}
    \end{minipage}
\end{table*}

%% file: tables/cycle_rules.tex
\begin{table*}[tb]
    \centering
    \begin{adjustbox}{width=0.9\linewidth}
    \begin{tabular}{@{}llcccc@{}}
    \toprule
    Method & Cycle rule & PCK@0.01 & PCK@0.05 & PCK@0.10 & $\Delta$@0.10\\
    \midrule
    Baseline & none & 11.04 & 54.10 & 69.79 & --\\
    MNN & reject if $d_{\mathrm{cyc}}>0$ & 11.07 & 54.27 & 70.07 & +0.28\\
    Binary penalty & $\mathbf{S}_{\ell_i m}-\lambda\,\mathbf{1}[d_{\mathrm{cyc}}>0]$ & 11.10 & 54.40 & 70.21 & +0.42\\
    Relaxed MNN & reject if $d_{\mathrm{cyc}}>1$ & \textbf{11.14} & 54.64 & 70.64 & +0.85\\
    \myrowcolor
    Ours & $\mathbf{S}_{\ell_i m}-\lambda\, d_{\mathrm{cyc}}$ & \textbf{11.14} & \textbf{54.85} & \textbf{71.16} & \textbf{+1.37}\\
    \bottomrule
    \end{tabular}
    \end{adjustbox}
    \caption{\textbf{Comparison against standard cycle-consistency rules on SPair-71k (\textit{per-keypoint}, full test set)} with TLR as the base method. All variants use the \textit{same} top-$p$ candidate pool and differ only in how the cycle distance $d_{\mathrm{cyc}}$ (\Cref{eq:cycle_distance}) is used.}
    \label{tab:cycle_rules}
\end{table*}

%% file: sections/6_conclusion.tex
\section{Conclusion}
\label{sec:conclusion}
We presented a training-free, plug-and-play re-ranking module for semantic correspondence. Our work is motivated by a simple observation: with modern pretrained features, correct matches often already appear in the top-ranked candidate pool, but are missed by greedy top-$1$ nearest-neighbor selection. This is a decoding bottleneck: the limiting factor is candidate selection, beyond representation quality alone. To address this, we combine adaptive top-$p$ candidate nomination with a soft cycle-consistency criterion inspired by the bijective mutual nearest-neighbor principle in multi-view geometry. The resulting method rewards matches that are not only semantically similar in the forward direction but also self-consistent under reverse feature-space queries. Across three representative base methods and both unsupervised and weakly supervised settings, our method consistently improves performance on SPair-71k, its geometric subset, and PF-Pascal. More broadly, our results suggest that revisiting the final selection step, rather than focusing solely on stronger feature representations, offers a simple and underexplored path toward better semantic correspondence.

%% file: sections/X_suppl.tex
\section{Appendix}
\subsection{Dataset Details}
\label{subsec:dataset}
\subsubsection{PF-Pascal}
We follow prior work \cite{zhang_tale_2023,zhang_telling_2024,dunkel_it_2025} and preprocess each image using aspect-ratio–preserving resizing, padding to a fixed resolution of $840\times840\times3$. Evaluation is conducted on the official test split, which contains 299 image pairs from 20 object categories.
\subsubsection{SPair-71k}
We use the same preprocessing pipeline, resizing all images to $840\times840\times3$ with aspect-ratio preservation. Experiments are performed on the official test split, comprising 12{,}234 image pairs across 18 categories.


\subsection{Implementation Details}
\label{subsec:implement}
We follow prior work \cite{zhang_tale_2023,zhang_telling_2024,dunkel_it_2025,stracke_cleandift_2025} and report both per-image-pair PCK (averaged over keypoints within each image pair) and per-keypoint PCK (averaged over all keypoints in the dataset). For efficiency on SPair-71k, we randomly subsample 10 examples per category for the ablation study in \Cref{tab:ablation,tab:alternative_formulation} and hyperparameter sensitivity analysis \Cref{fig:hyperparam_pascal,fig:hyperparam_spair}. Every analysis of the re-ranking criterion itself (\Cref{tab:cycle_rules} in the main paper and \Cref{appendix:analysis,appendix:kap,appendix:failure} below) is instead computed on the \textit{full} SPair-71k test set: 12{,}234 pairs and 88{,}328 visible keypoints. All variants within such a table are scored on the same similarity matrix in a single pass with the same code, so differences in the $\Delta$ columns come from the scoring rule alone.

We use the object masks from \cite{zhang_telling_2024} to restrict candidate generation to the object. This adds no new dependency: the released TLR pipeline already computes these masks for adaptive pose alignment, and we simply reuse them. It does, however, mean that our own run of the base pipeline scores 69.79 PCK@0.10 where \cite{zhang_telling_2024} reports 69.4, and we use 69.79 as the baseline in every full-set analysis so that only the cycle rule varies. \Cref{tab:mask_ablation} separates the two contributions.




\subsection{Additional Quantitative Results}
\label{appendix:additional_quantitative}
\Cref{tab:perkpts_allpck} supplements the per-image results in the main paper (\Cref{tab:perimg_allpck}) with the corresponding per-keypoint evaluation across all thresholds and datasets. The gains are consistent across both evaluation protocols and all thresholds, with the per-keypoint metric showing slightly larger absolute improvements—reflecting that our method is especially effective at recovering keypoints that are individually ambiguous but structurally consistent.

\input{tables/perkpts_allpck}

\Cref{tab:spair_geo_percat_perkpts_pck010} reports per-category per-keypoint PCK@0.10 on the geometric subset of SPair-71k, which filters image pairs for large viewpoint and scale changes. The category-level pattern is consistent with the full SPair-71k results: categories with repeated parts (e.g., chair, plant, tvmonitor) see the largest improvements, confirming that cycle consistency is most effective precisely when the forward nearest-neighbor is most susceptible to false positives from visually similar but geometrically inconsistent patches.

\input{tables/spair_geo_percat_perkpts_pck010}

\Cref{tab:pascal_percat_pck015} reports per-category PCK@0.15 on PF-Pascal under both per-keypoint and per-image evaluation. The gains on PF-Pascal are larger than on SPair-71k, with consistent improvements across the majority of the 20 object categories in both evaluation modes.

\input{tables/pascal_percat_pck015}

\Cref{tab:spair_percat_perimg_pck010,tab:spair_geo_percat_perimg_pck010} provide the per-image per-category breakdown on SPair-71k and its geometric subset, complementing the per-keypoint tables above. The gains are consistent across evaluation protocols, with the same categories benefiting most from cycle-consistent re-ranking in both settings.

\input{tables/spair_percat_perimg_pck010}

\input{tables/spair_geo_percat_perimg_pck010}

We additionally assess the statistical significance of the observed improvements using McNemar's paired test on all 12,234 test pairs in SPair-71k. For each image pair, a predicted keypoint is marked correct if its location falls within the PCK threshold $\alpha \cdot \mathrm{bbox\_size}$ ($\alpha = 0.1$) of the ground-truth, and only keypoints visible in both source and target images are evaluated, yielding 88,328 binary correctness observations for TLR. We use the two-sided test with continuity correction, which counts $b$ (baseline correct, ours wrong) and $c$ (baseline wrong, ours correct) and tests whether the two discordant counts are symmetric under $H_0$. For TLR, the test yields $p < 0.001$ (\Cref{tab:winloss}), confirming that the improvement is statistically significant.

\subsection{Analysis of the Re-ranking Criterion}
\label{appendix:analysis}

This section collects the analyses that isolate what the cycle criterion contributes, all on the full SPair-71k test set with TLR as the base method.

\noindent\textbf{Composition with existing refinement components.}
Adaptive pose alignment and window soft-argmax are enabled in the numbers reported in the main paper, and our gains are measured on top of them. \Cref{tab:composition} measures each component separately and in composition. Our re-ranking improves the full stack from 69.79 to 71.16 ($+1.37$), improves the pipeline with window soft-argmax disabled by more ($68.53 \rightarrow 70.36$, $+1.83$), and still helps with adaptive pose disabled ($65.36 \rightarrow 66.24$, $+0.88$). The two refinements act on different error sources, sub-patch \textit{localization} versus patch \textit{selection}, which is why they compose and why our contribution is larger when soft-argmax is absent.

\input{tables/composition}

\noindent\textbf{When re-ranking helps and when it hurts.}
\Cref{tab:winloss} breaks down all 88{,}328 keypoints. Our criterion changes 13.7\% of decisions: 3{,}423 from wrong to right against 2{,}213 from right to wrong, so 60.7\% of the keypoints whose correctness flips are flipped in our favor. Among the keypoints whose match actually changes, accuracy rises by $+10.0$ points, against $+1.37$ overall. The overall number is smaller only because the remaining 86.3\% of matches are left untouched, not because the criterion is unreliable. We report the 2{,}213 degradations rather than only the net, and characterize where they land in \Cref{appendix:failure}. Every rule in the table is statistically significant against the baseline under McNemar's test.

\input{tables/winloss}

\noindent\textbf{Sensitivity to the nomination parameters.}
The main paper sweeps $\lambda$, the one parameter that enters the re-ranking score. \Cref{tab:hyperparam_sweep} sweeps the remaining three. All three only control how many candidates are nominated, and none of them appears in the re-ranking score of \Cref{eq:reranking_score}, so they are not tuning knobs in the same sense as $\lambda$. The maximum deviation across every setting is 0.17 PCK@0.10. In particular, $K_{\max}=30$ is 0.8\% of the $L=3{,}600$-patch feature grid and removing the cutoff altogether costs only 0.13 points; it is a safeguard for the rare case where a similarity map is nearly flat, not a tuned parameter.

\input{tables/hyperparam_sweep}

\noindent\textbf{Supervised setting.}
\Cref{tab:supervised} applies our approach to the released supervised TLR checkpoint. Our method results in $+0.01$ with $p=0.81$, below the $+1.37$ obtained on unsupervised features. The headroom is still there: of the keypoints this supervised model gets wrong at top-$1$, $87.6\%$ have a correct match within its top $30$, against $85.6\%$ for SD+DINO (\Cref{tab:topk_backbones}).

\input{tables/supervised}

\noindent\textbf{Mask-constrained nomination.}
\Cref{tab:mask_ablation} separates the mask from the cycle criterion. With the mask stripped entirely, the criterion is still worth $+1.06$, against $+1.37$ with it.

\input{tables/mask_ablation}

\subsection{Beyond PCK: Keypoint Average Precision}
\label{appendix:kap}
PCK only asks whether a predicted point lands close enough to the ground truth. It does not check whether the match is one-to-one, which is exactly what our criterion enforces. We therefore additionally report Keypoint Average Precision (KAP) \cite{mariotti_improving_2024}, which penalizes both missed matches and wrong matches to repeated or symmetric parts. We follow the reference implementation of \cite{mariotti_improving_2024}: each keypoint contributes one positive, the maximum score inside the radius $\alpha \cdot \mathrm{bbox\_size}$, and one negative, the maximum score outside it, and keypoints whose target annotation is not visible are skipped. We report the per-category mean AP, the form used in prior work.

\Cref{tab:kap} shows that our improvement under KAP@0.1 is $+4.94$, which is $3.6\times$ our improvement under PCK@0.10 on identical predictions. KAP was introduced to expose exactly the symmetry and repetition errors that our penalty suppresses, so it shows more directly that our method fixes part-level confusions that PCK barely penalizes.

\input{tables/kap}

\subsection{Failure Mode Analysis}
\label{appendix:failure}

\Cref{tab:winloss} shows that our criterion turns 2{,}213 correct predictions into incorrect ones. This section characterizes them.

The regressions are concentrated rather than scattered. Of the 2{,}213 degraded keypoints, \textbf{64.5\%} land on another annotated keypoint rather than on background, meaning the prediction moves to a different semantic part rather than off the object.

The reason follows from what the criterion can and cannot tell apart. Cycle consistency distinguishes a correct part from a \textit{spurious} one, because the spurious candidate routes back to a distant source patch. It cannot distinguish two \textit{genuinely equivalent} ones: when two mirror-symmetric parts are near-equidistant in feature space, both are approximately cycle-consistent and the reverse query is uninformative between them. Resolving that requires a global or pose-level prior, which our criterion does not have. The reverse pattern confirms this: the largest gains occur where repeated parts are spatially distinguishable rather than mirror-equivalent, such as the two near-identical vertical edges of a tv ($+7.2$), the many similar leaves of a potted plant ($+6.3$) and the four similar legs of a chair ($+4.9$), where the reverse query does disambiguate.

The net effect on part confusion is nonetheless strongly positive. KAP penalizes exactly this error class, spurious matches to repeated or symmetric parts, and it is where our largest improvement occurs ($+4.94$, \Cref{tab:kap}). The residual failures are concentrated in symmetry, but the criterion resolves substantially more part ambiguity than it creates.

Finally, we quantify how much of the headroom in \Cref{fig:teaser} the criterion actually recovers, since top-$k$ recall is an upper bound rather than an achievable target. The ground-truth patch is nominated into the candidate pool but is not the top-$1$ candidate for 48{,}651 keypoints; of these our re-ranking promotes 3{,}457 to the top, a recovery rate of \textbf{7.1\%}. We report this plainly: a large, well-defined headroom exists at the decoding step, a training-free criterion captures a measurable share of it on top of every base method we tested, and the remainder is an open problem.

\subsection{Category-Level Analysis}
\label{sec:category_analysis}

\Cref{tab:spair_percat_perkpts_pck010} reveals a consistent pattern in which categories benefit most from cycle-consistent re-ranking. Across both TLR and CleanDIFT base methods, the largest per-category gains on SPair-71k occur in \textbf{tv} (+7.2 / +6.4), \textbf{plant} (+6.3 / +6.5), \textbf{chair} (+4.9 / +5.0), and \textbf{train} (+2.1 / +2.0). These categories share a common property: they contain repeated parts that are visually similar to one another, such as a tv screen has two nearly identical vertical edges, a potted plant has multiple visually similar leaves, and a chair has four legs with near-identical appearance. In such cases, the greedy forward nearest-neighbor can latch onto any of these equivalent parts, producing a match that is locally plausible but globally inconsistent. Our criterion mitigates this by penalizing candidates whose reverse match lands far from the source patch: when multiple patches in the target have similar forward similarity, only the one that is truly paired with the source maps back close to it.

Conversely, \textbf{horse} (-0.9 / -1.2) and \textbf{bird} (-0.2 / -0.4) regress under both base methods. Both are deformable categories with highly symmetric bilateral body structure: the left and right limbs of a horse, or the two wings of a bird, are often so similar in feature space that the cycle-consistency signal is ambiguous—a match to the wrong leg may still approximately pass the cycle check, since both legs are mutual nearest neighbors of similar source patches. This failure mode is consistent with the limitation discussed in \Cref{sec:results}: cycle consistency can be misled when symmetric parts are equidistant in feature space, making the reverse nearest-neighbor uninformative about which instance of a repeated part is the correct match.

On PF-Pascal (\Cref{tab:pascal_percat_pck015}) with TLR, \textbf{tv} (+8.1) and \textbf{bottle} (+7.8) show the largest per-keypoint gains, followed by \textbf{sofa} (+6.3) and \textbf{chair} (+5.7).

\subsection{Additional Visualizations of Cycle-Consistent Re-ranking}

\begin{figure*}
    \centering
    \begin{tabular}{@{}c@{\hspace{1mm}}c@{}}
         \includegraphics[width=0.49\linewidth]{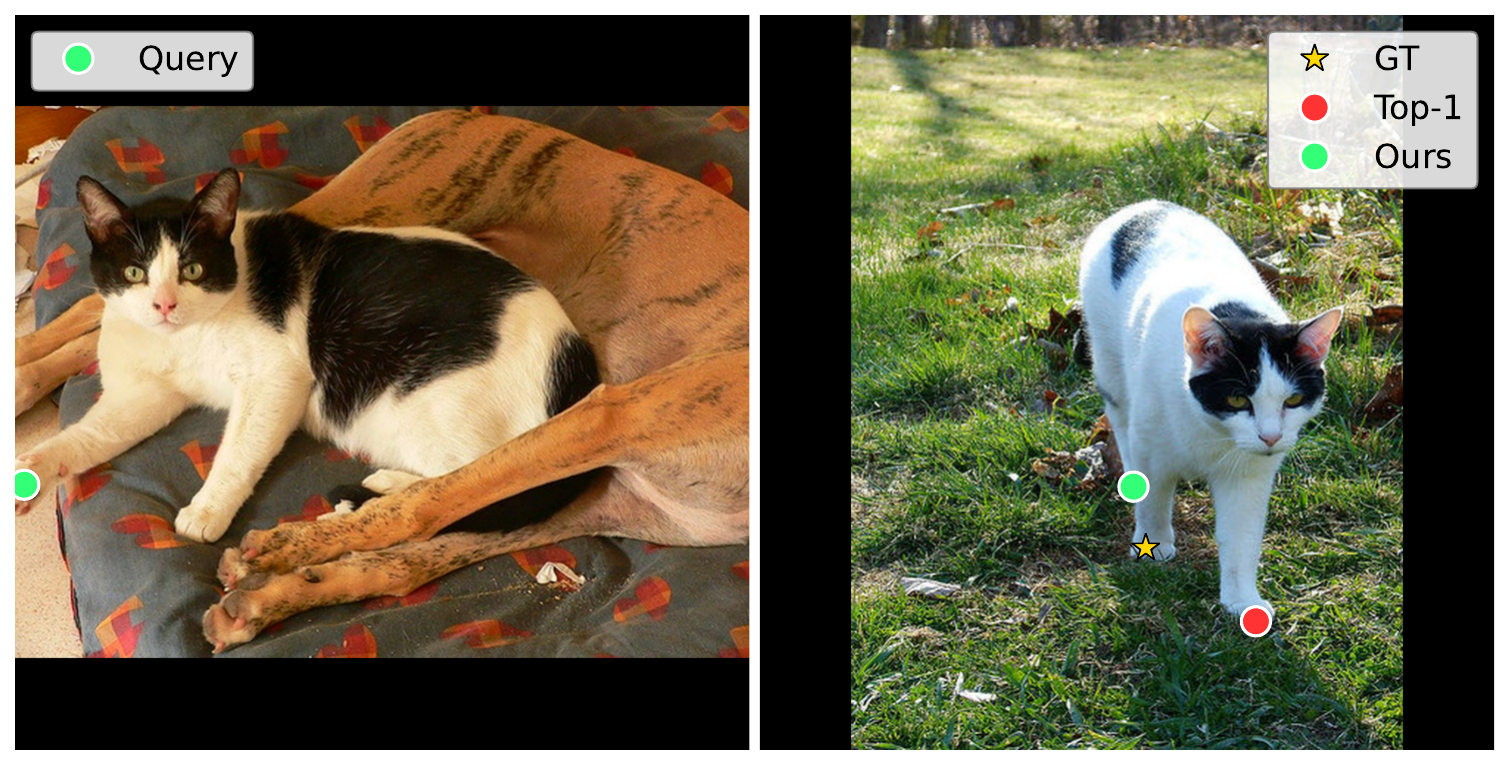} & \includegraphics[width=0.49\linewidth]{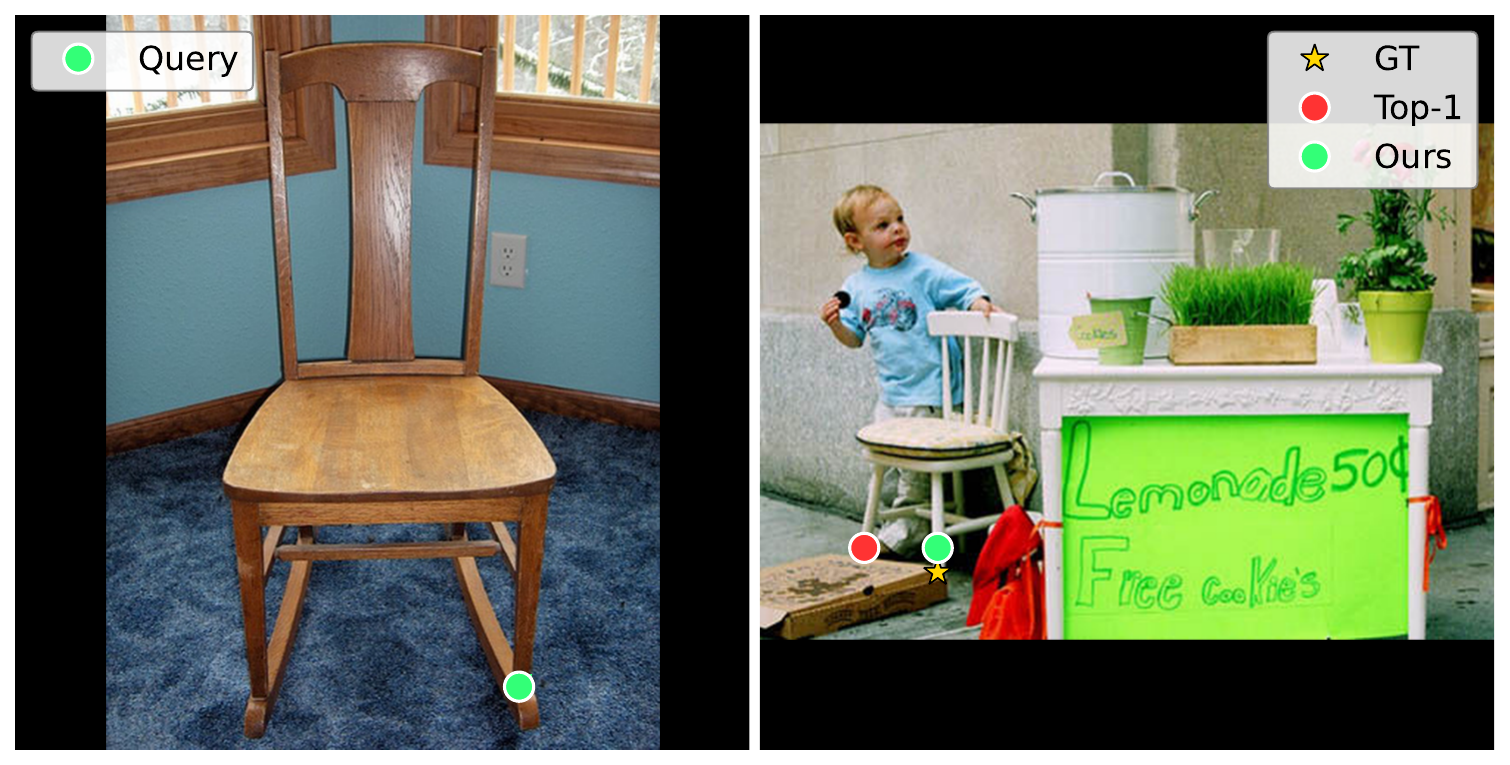}\\
         \includegraphics[width=0.49\linewidth]{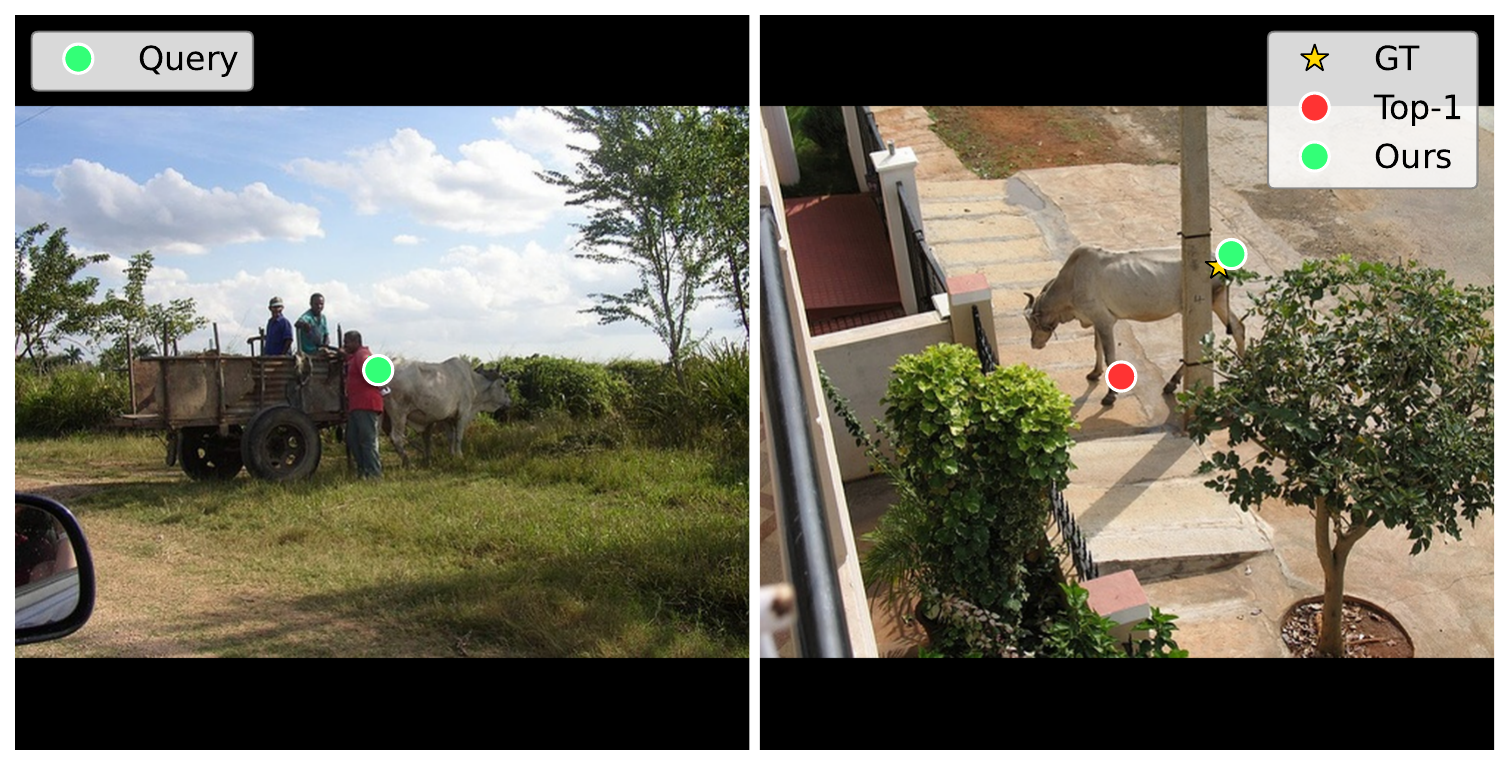} & \includegraphics[width=0.49\linewidth]{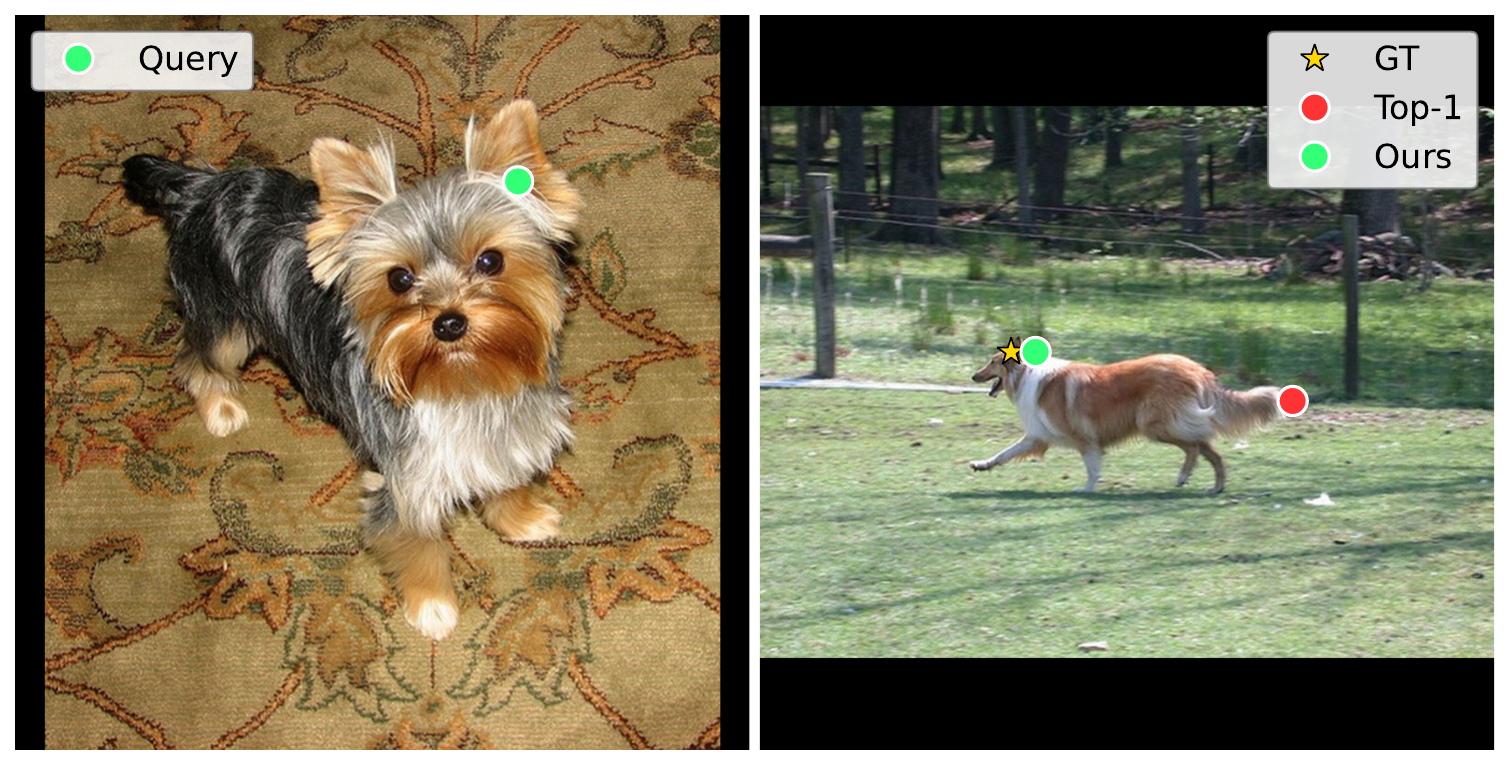} \\
         \includegraphics[width=0.49\linewidth]{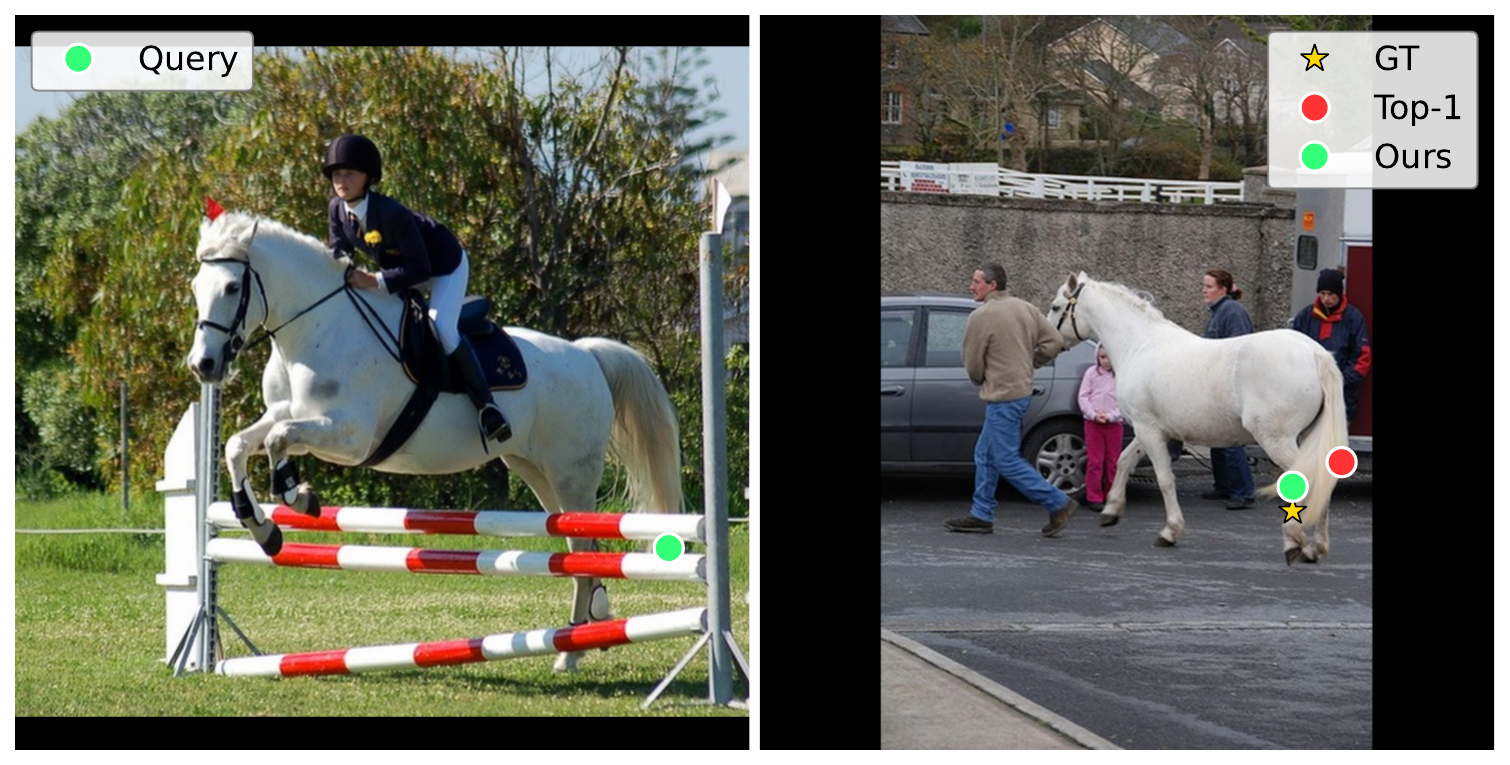} & \includegraphics[width=0.49\linewidth]{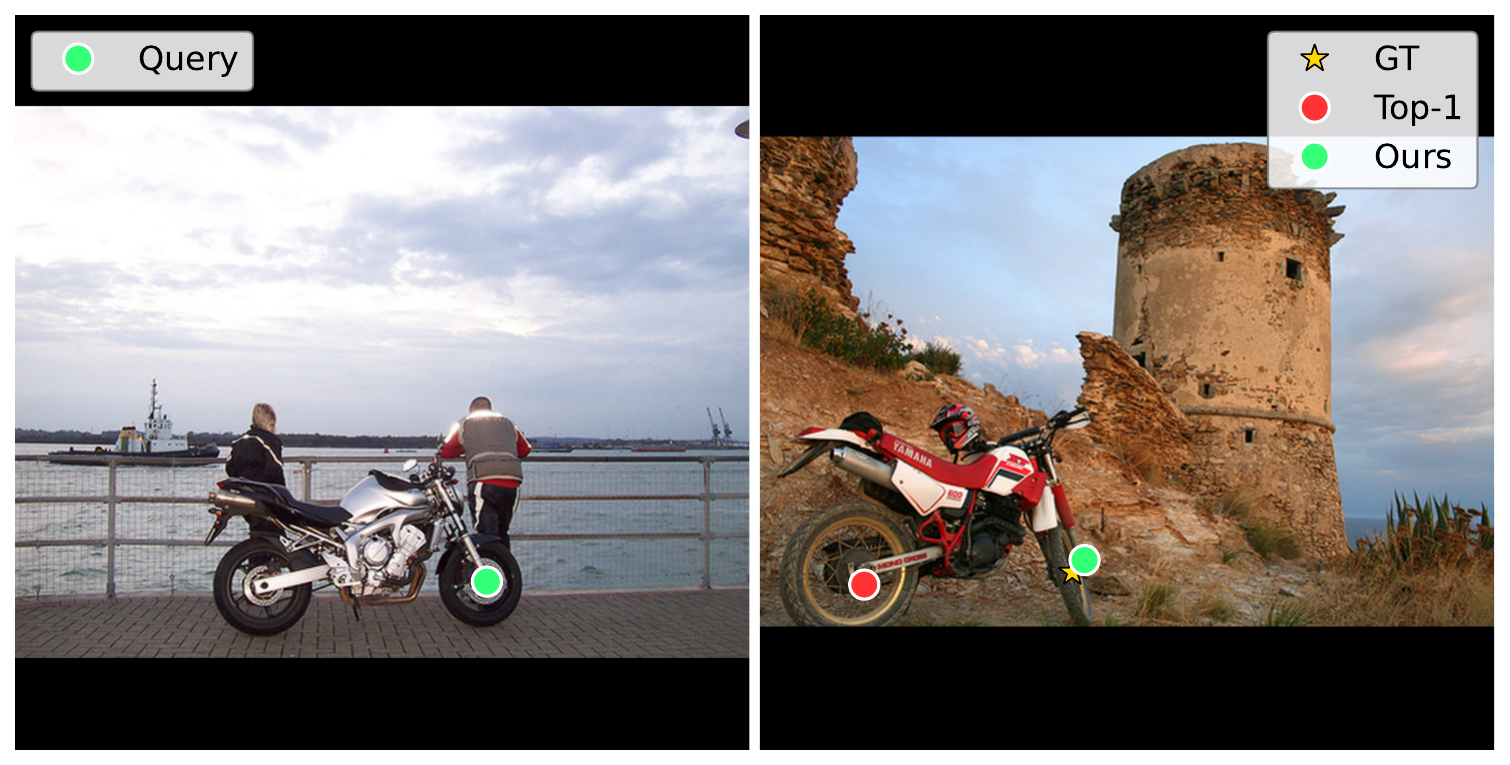} \\
         \includegraphics[width=0.49\linewidth]{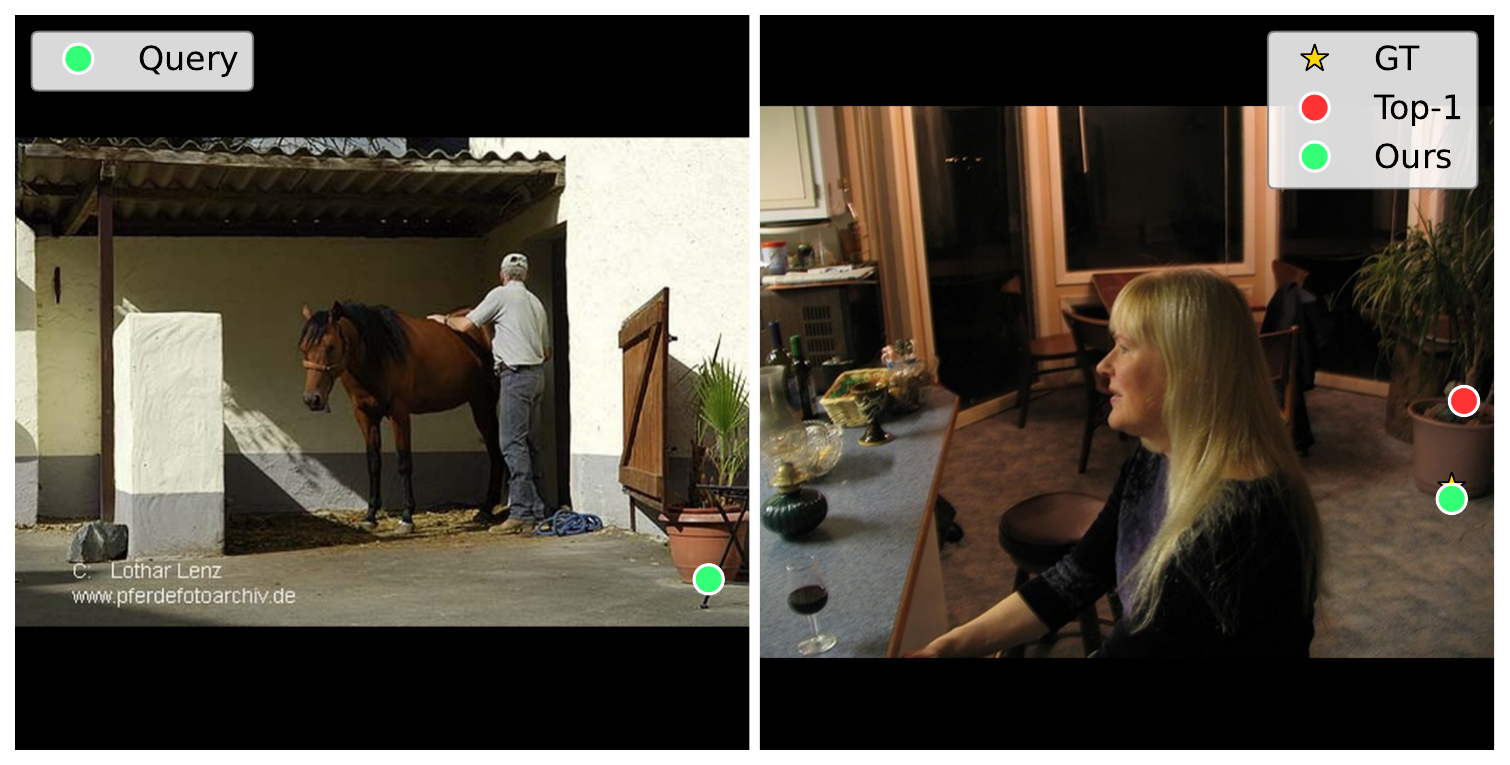} & \includegraphics[width=0.49\linewidth]{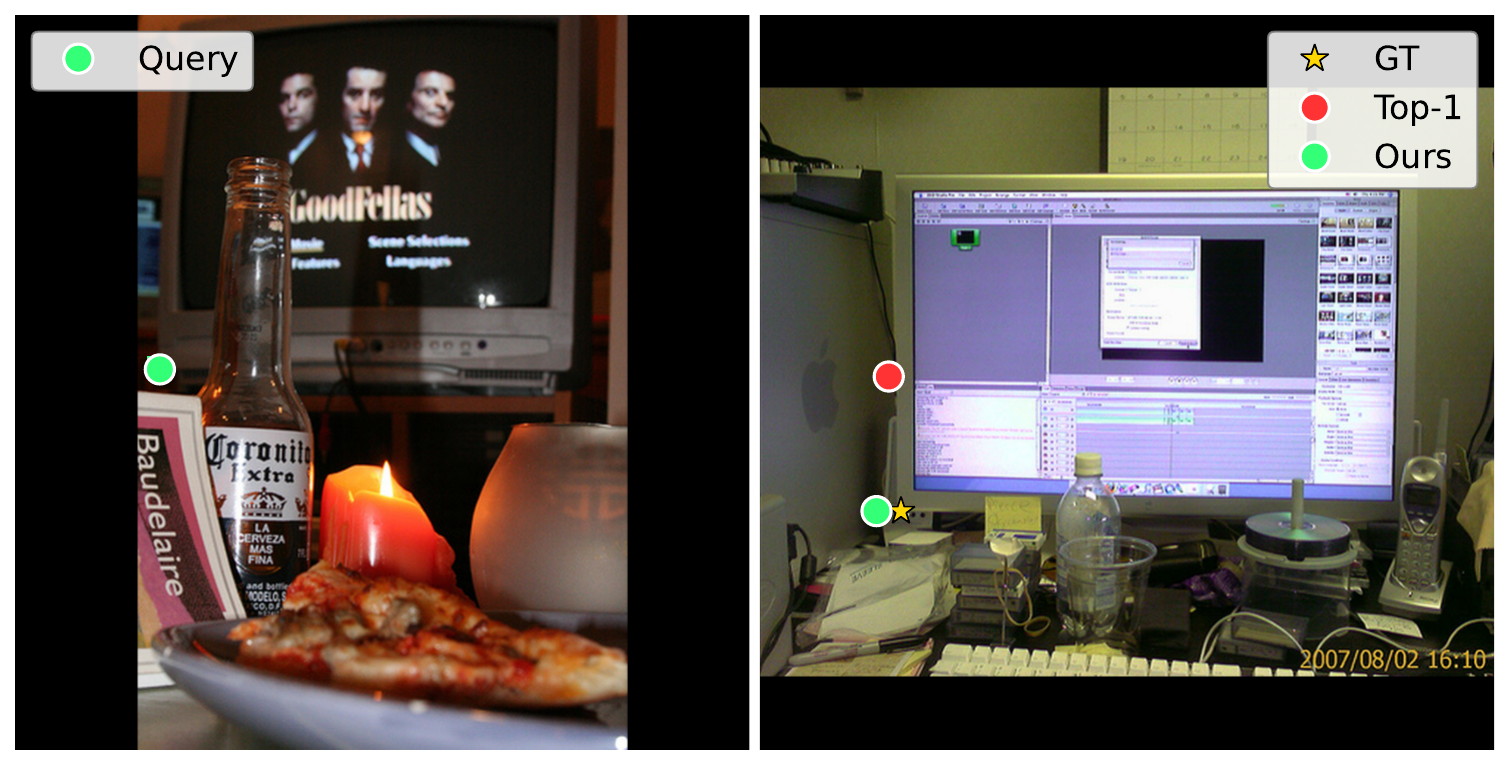}
    \end{tabular}
    \caption{\textbf{Additional qualitative examples showing the benefit of cycle-consistent re-ranking on SPair-71k.} Each panel shows the source keypoint (left) and three predicted locations on the target (right): ground truth (GT), the top-1 nearest-neighbor prediction, and our method's prediction. Our cycle-consistent re-ranking consistently recovers the correct correspondence that top-1 matching fails to identify.}
    \label{fig:top_k_supp}
\end{figure*}

\Cref{fig:top_k_supp} extends the qualitative analysis of \Cref{fig:top_k_main} in the main paper to eight additional object categories. In each case, the top-$1$ nearest-neighbor selects an incorrect target location while our cycle-consistent re-ranking recovers the correct correspondence from the same top-$p$ candidate pool. The examples span both rigid categories (e.g., tv monitor, chair) and deformable ones (e.g., horse, dog), demonstrating that the selection benefit of cycle consistency generalizes broadly.

\subsection{Additional Hyperparameter and Coverage Analysis}

\begin{figure*}
    \centering
    \begin{tabular}{@{}c@{\hspace{0mm}}c@{\hspace{0mm}}c@{}}
        \includegraphics[width=0.33\textwidth]{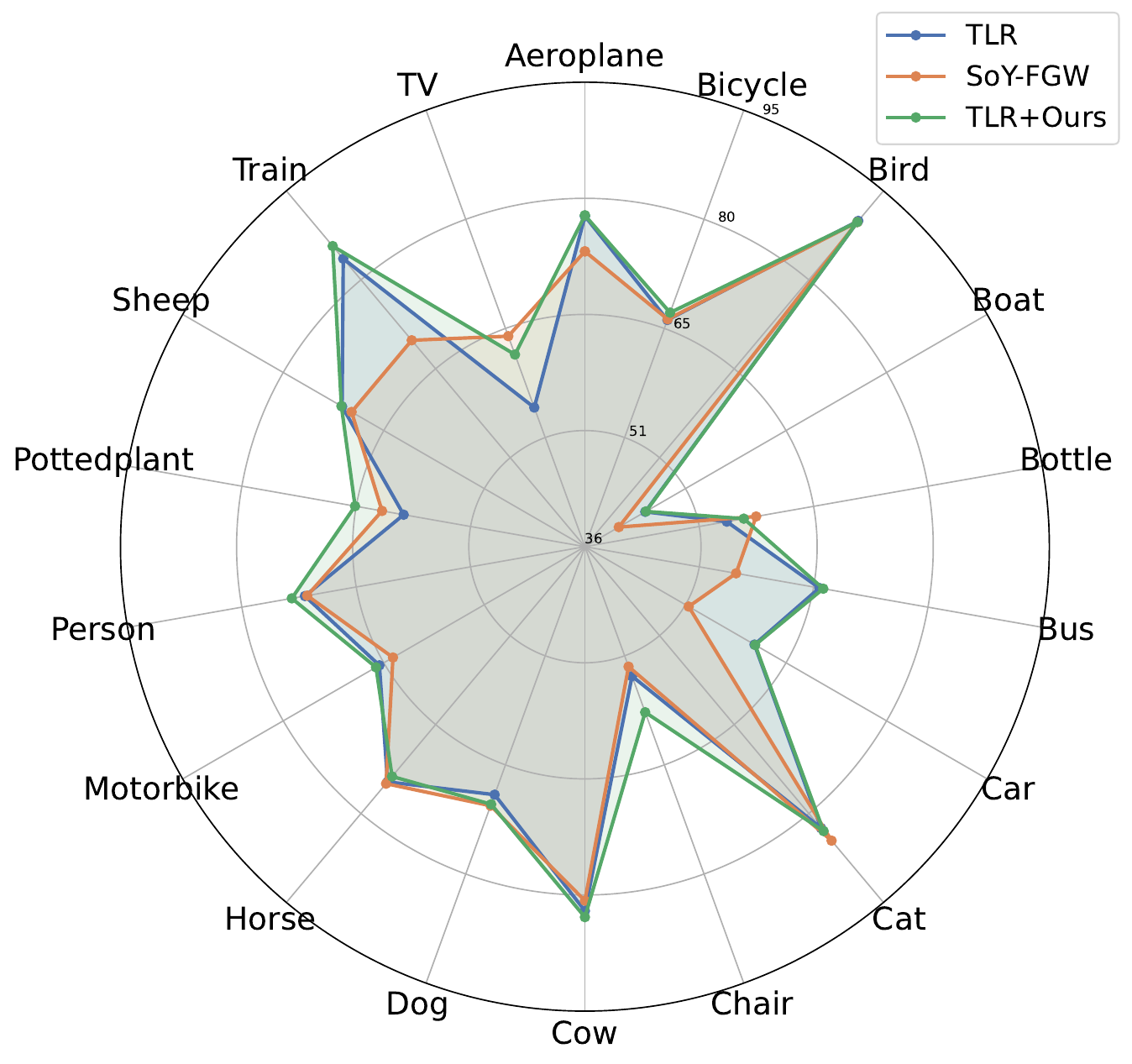} & \includegraphics[width=0.33\textwidth]{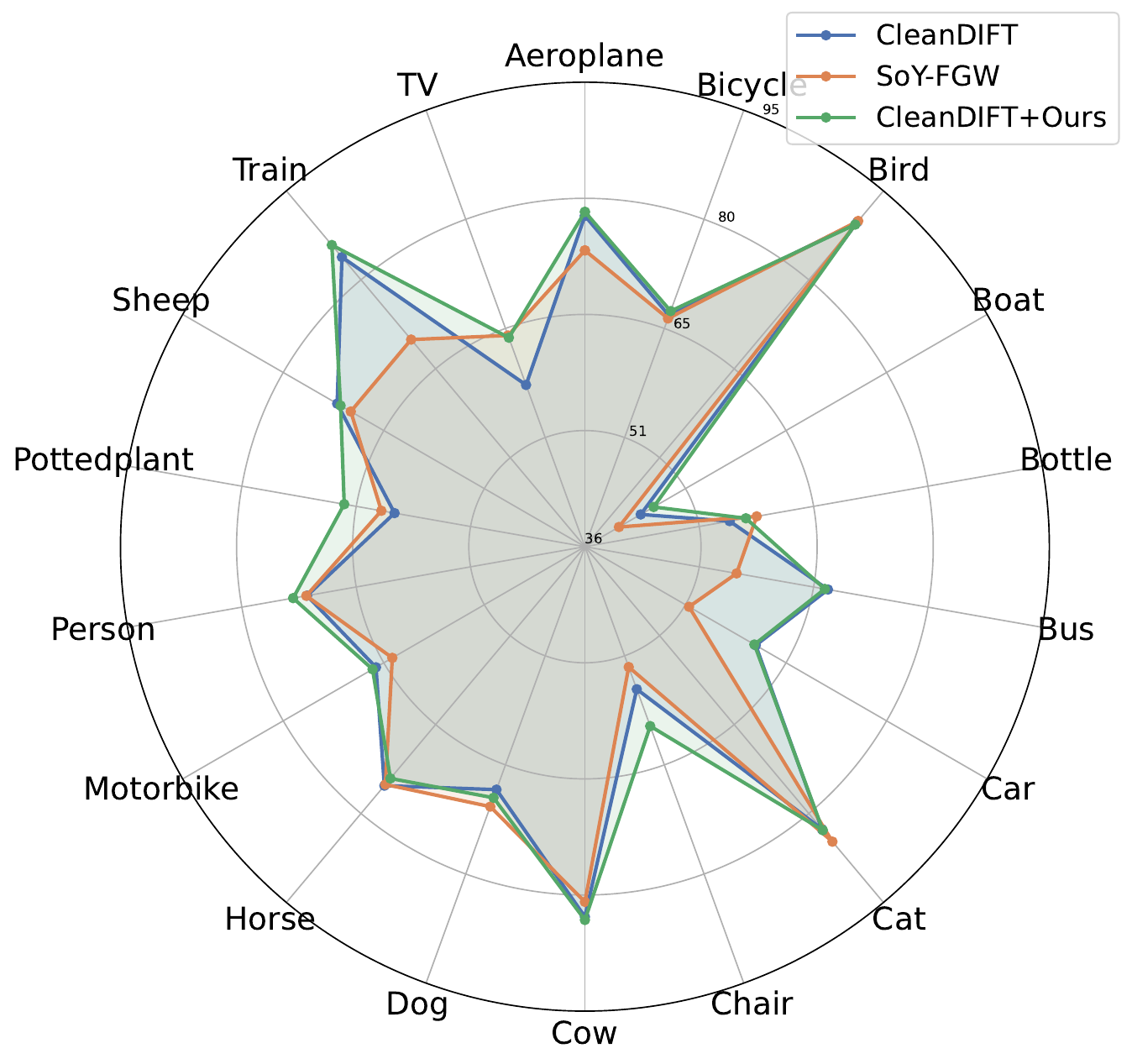} & \includegraphics[width=0.33\textwidth]{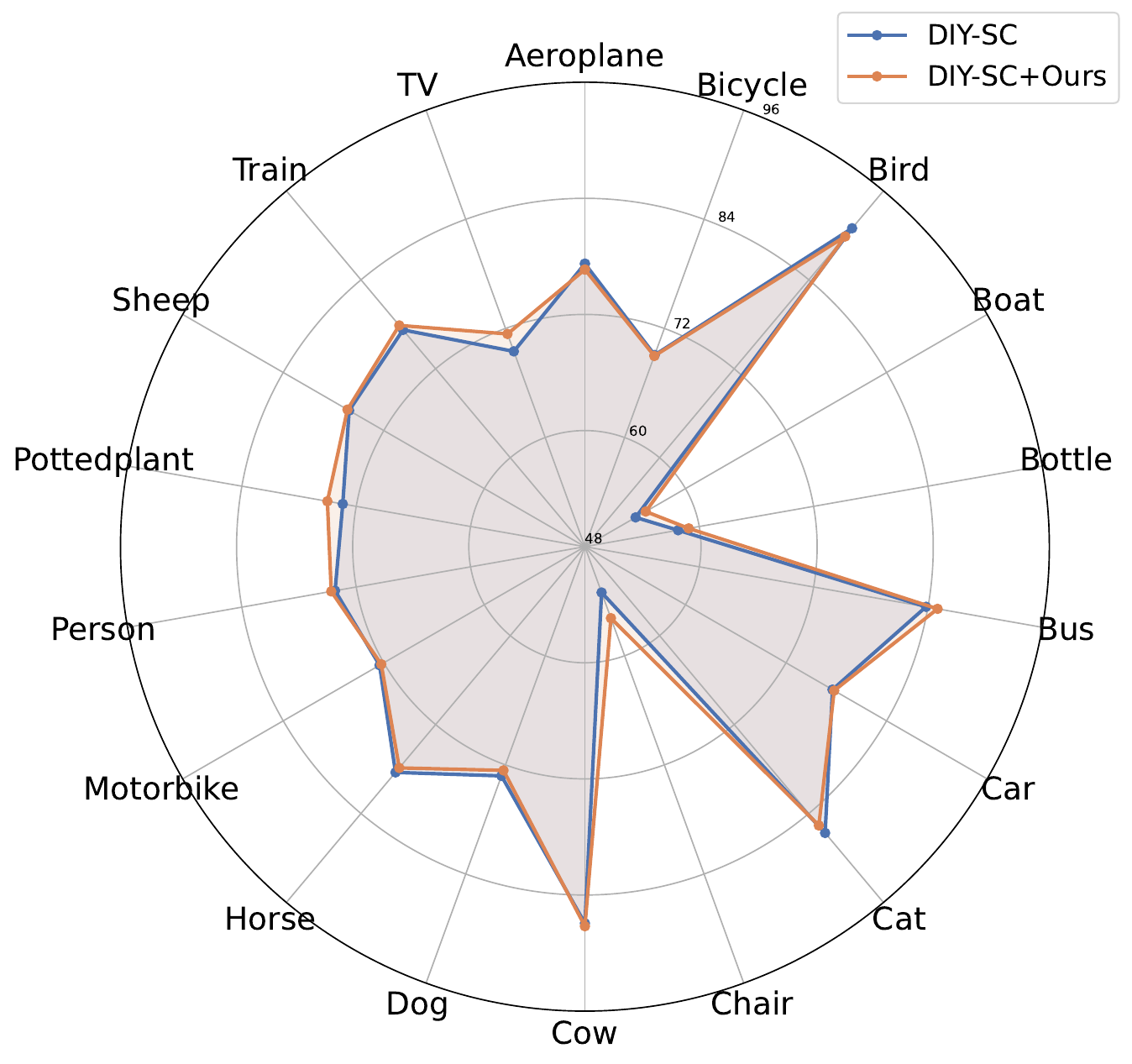} \\
    \end{tabular}
    \caption{\textbf{Per-category radar plots on SPair-71k (\textit{per-keypoint}, PCK@0.10).} Each axis represents one object category. The filled area shows our method applied to TLR (left), CleanDIFT (middle), and DIY-SC (right); the outline shows the respective baseline. Our method broadly expands coverage across categories for all three base methods.}
    \label{fig:radar_spair}
\end{figure*}

\Cref{fig:radar_spair} provides per-category radar plots on SPair-71k for all three base methods, extending the PF-Pascal radar plots in \Cref{fig:radar_pascal}. Across TLR, CleanDIFT, and DIY-SC, our method expands coverage in the majority of categories, with the largest expansions in tv, plant, and chair, which is consistent with the category-level analysis in \Cref{sec:category_analysis}. The radar plots also make visible the few categories (e.g., horse) where coverage does not expand or slightly contracts, confirming that the limitation is category-specific and traceable to part symmetry rather than being a general property of the method.

\begin{figure*}
    \centering
    \begin{tabular}{@{}c@{\hspace{0mm}}c@{}}
        \includegraphics[width=0.49\textwidth]{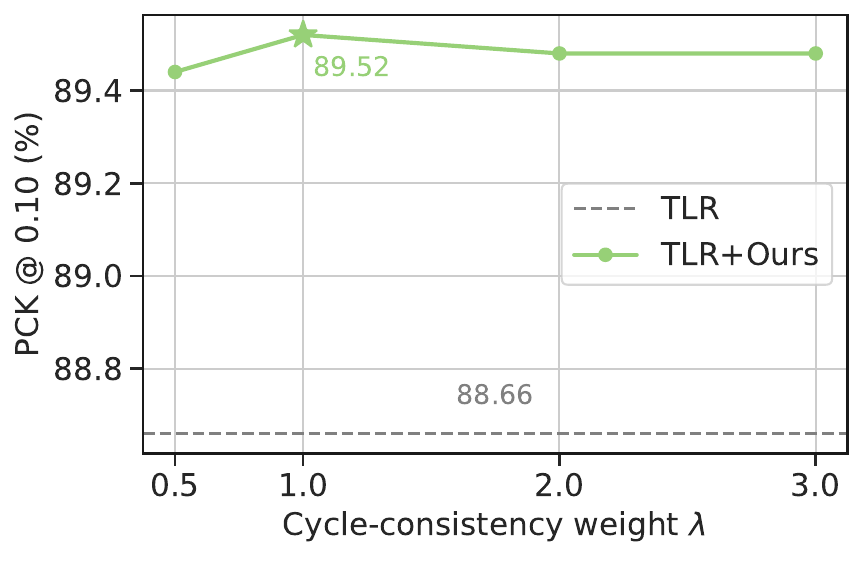} & \includegraphics[width=0.49\textwidth]{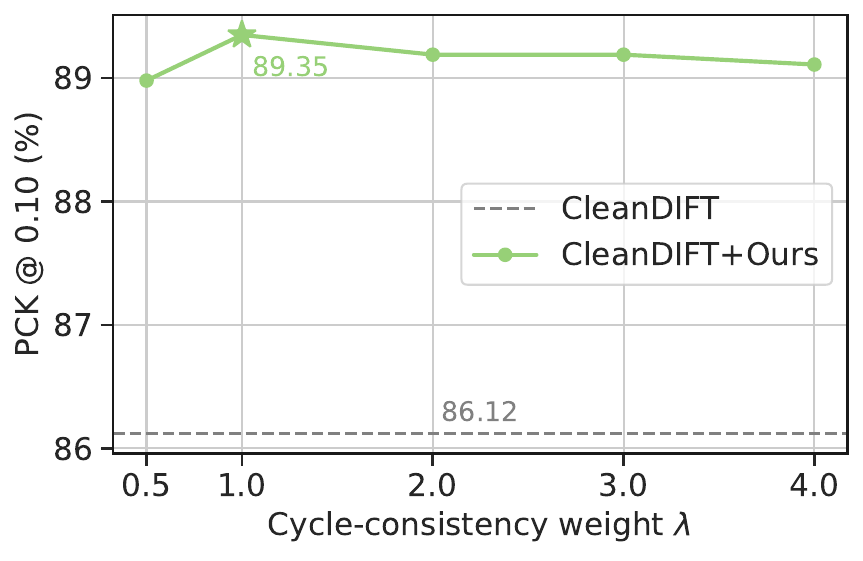} \\
    \end{tabular}
    \caption{\textbf{Sensitivity of the cycle-consistency weight $\lambda$ on PF-Pascal (\textit{per-keypoint}, PCK@0.1)} for TLR (left) and CleanDIFT (right). Performance remains stable across a broad range of $\lambda$ values for both base methods.}
    \label{fig:hyperparam_pascal}
\end{figure*}

\Cref{fig:hyperparam_pascal} reports the sensitivity of the cycle-consistency weight $\lambda$ on PF-Pascal, complementing the SPair-71k sweep in \Cref{fig:hyperparam_spair}. The same stability pattern holds: both TLR and CleanDIFT consistently improve over their baselines across a broad range of $\lambda$ values. The robustness of the hyperparameter across two datasets with different visual statistics (PF-Pascal has cleaner backgrounds and more canonical poses than SPair-71k) further supports that cycle consistency provides a reliable and dataset-agnostic re-ranking signal.

\subsection{Additional Qualitative Comparisons on SPair-71k}

\begin{figure*}[tb]
    \centering
    \begin{tabular}{@{}c@{\hspace{0mm}}c@{\hspace{1mm}}|c@{\hspace{0mm}}c@{}}
         \includegraphics[width=0.24\linewidth]{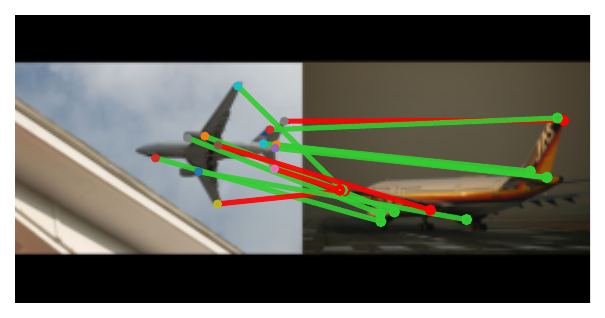}&
         \includegraphics[width=0.24\linewidth]{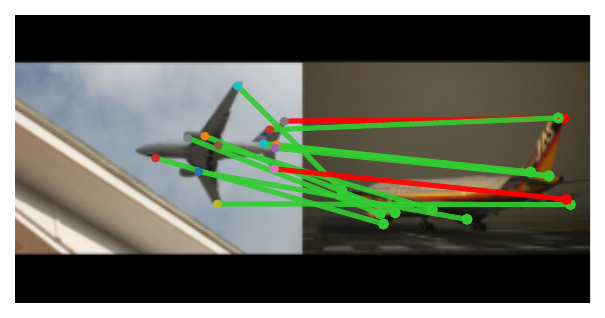} &
         \includegraphics[width=0.24\linewidth]{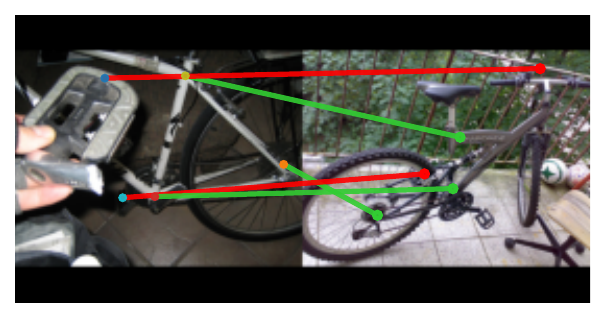}&
         \includegraphics[width=0.24\linewidth]{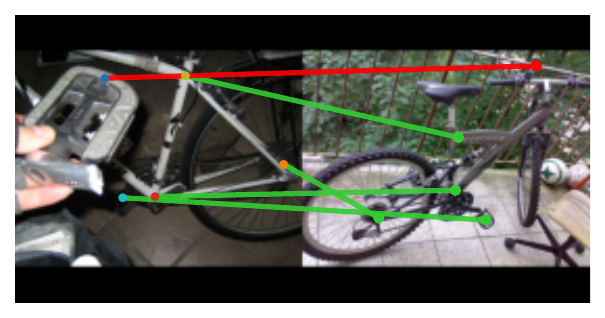} \\

         \includegraphics[width=0.24\linewidth]{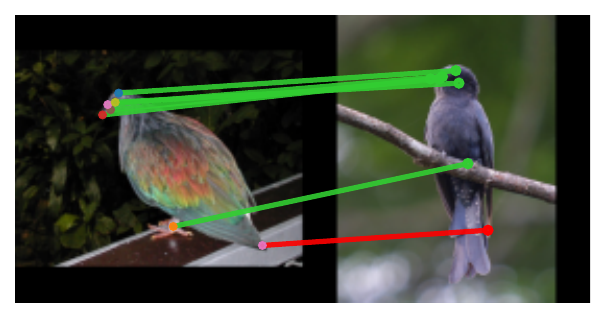}&
         \includegraphics[width=0.24\linewidth]{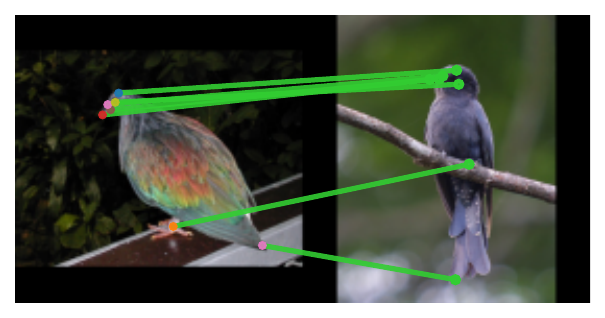} &
         \includegraphics[width=0.24\linewidth]{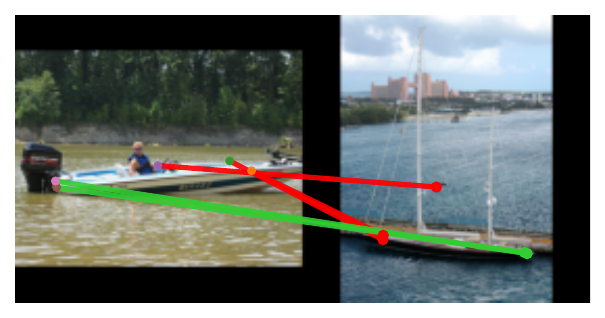}&
         \includegraphics[width=0.24\linewidth]{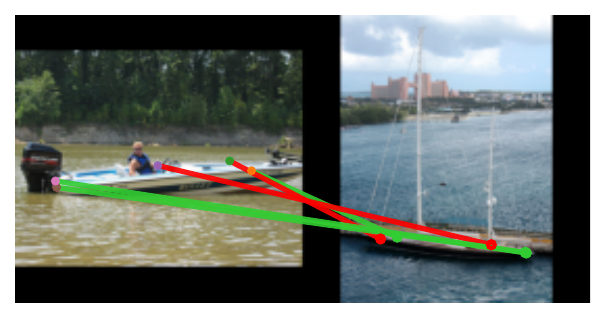} \\

         \includegraphics[width=0.24\linewidth]{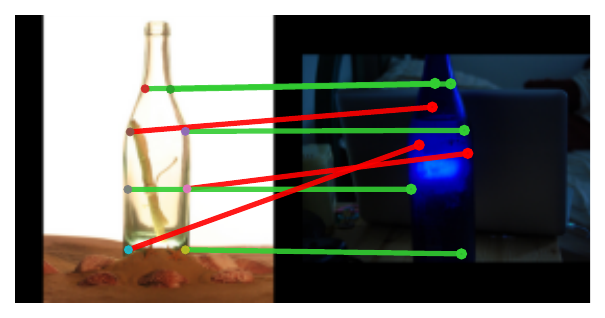}&
         \includegraphics[width=0.24\linewidth]{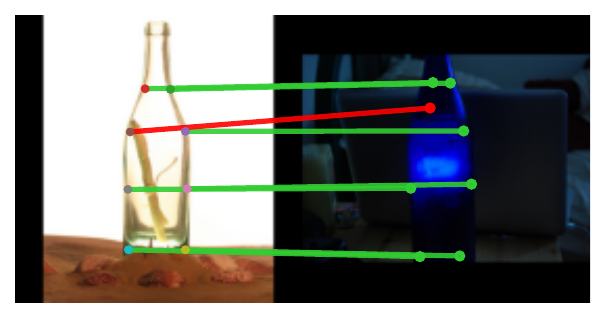} &
         \includegraphics[width=0.24\linewidth]{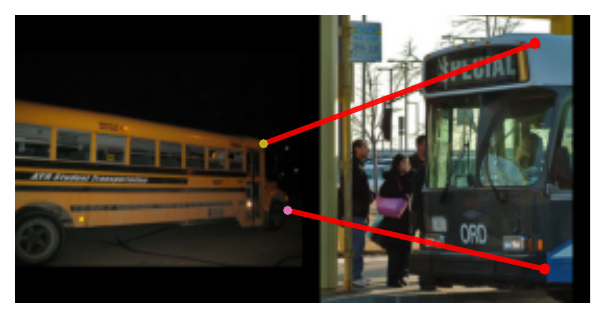}&
         \includegraphics[width=0.24\linewidth]{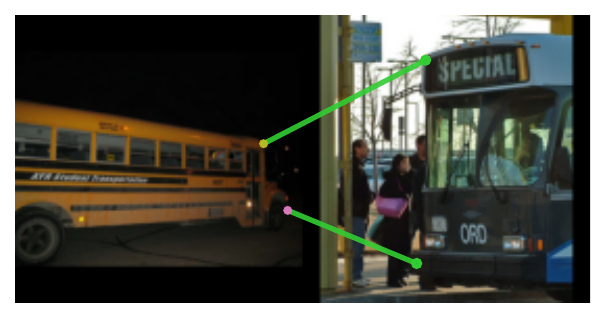} \\

         \includegraphics[width=0.24\linewidth]{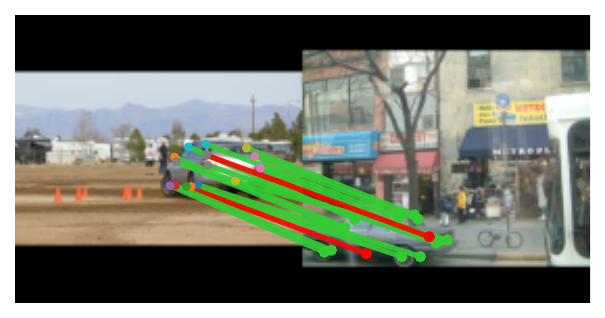}&
         \includegraphics[width=0.24\linewidth]{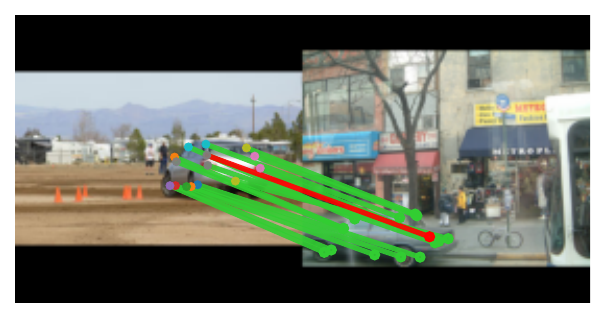} &
         \includegraphics[width=0.24\linewidth]{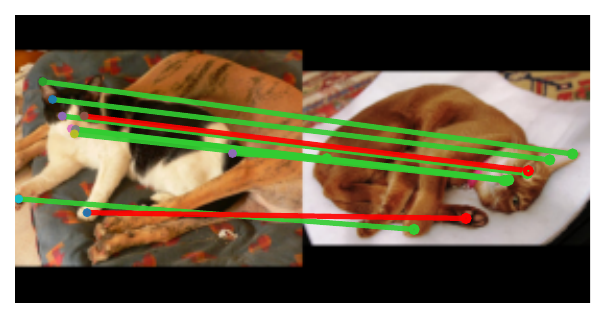}&
         \includegraphics[width=0.24\linewidth]{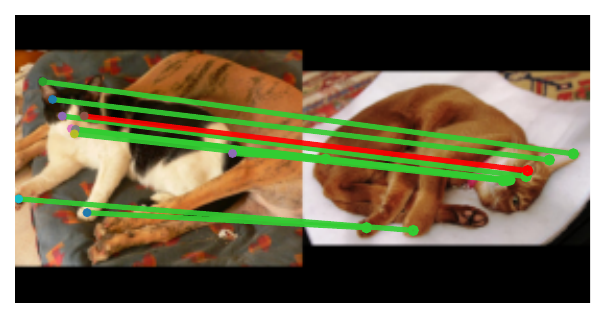} \\

         \includegraphics[width=0.24\linewidth]{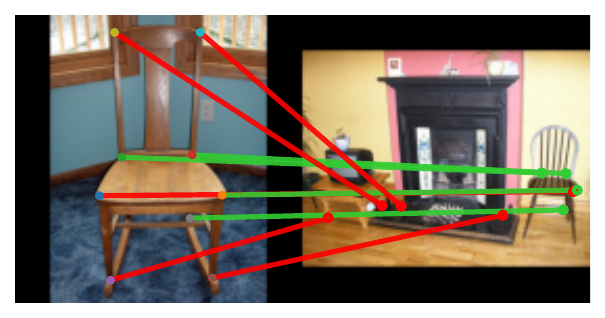}&
         \includegraphics[width=0.24\linewidth]{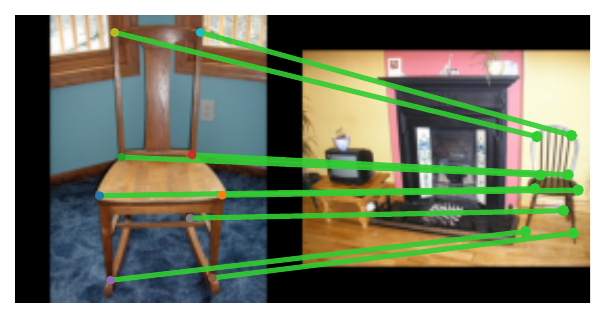} &
         \includegraphics[width=0.24\linewidth]{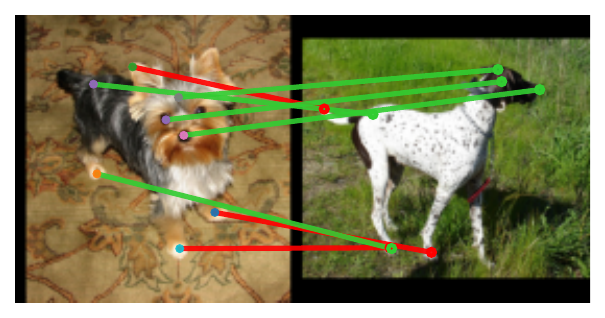}&
         \includegraphics[width=0.24\linewidth]{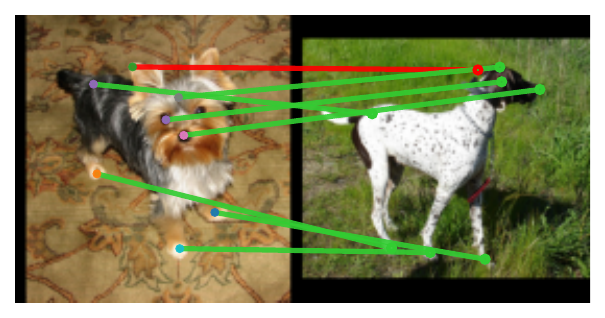} \\

         \includegraphics[width=0.24\linewidth]{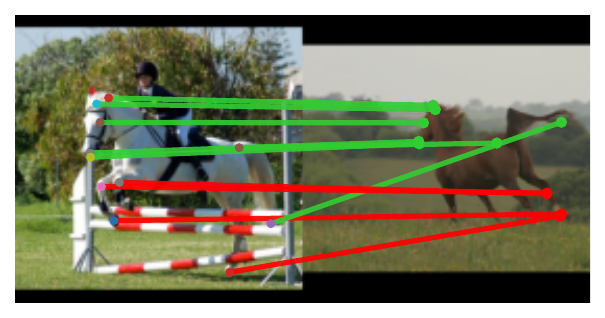}&
         \includegraphics[width=0.24\linewidth]{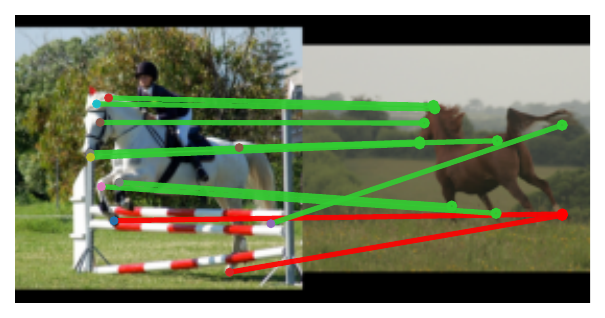} &
         \includegraphics[width=0.24\linewidth]{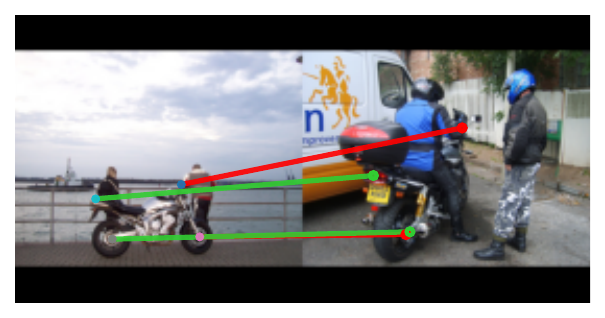}&
         \includegraphics[width=0.24\linewidth]{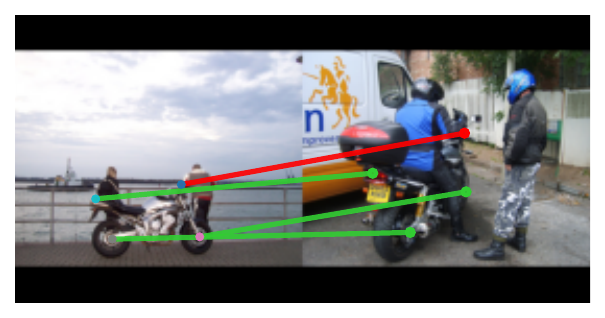} \\

         \includegraphics[width=0.24\linewidth]{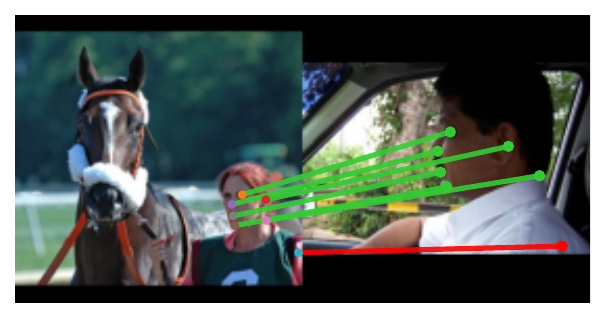}&
         \includegraphics[width=0.24\linewidth]{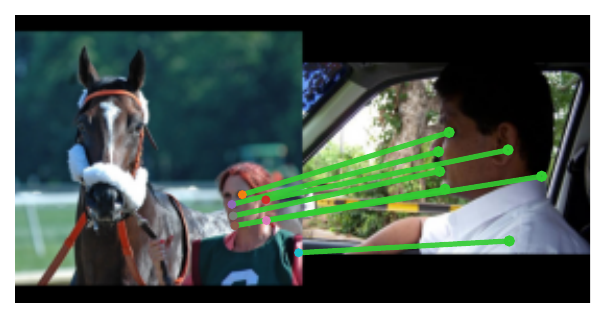} &
         \includegraphics[width=0.24\linewidth]{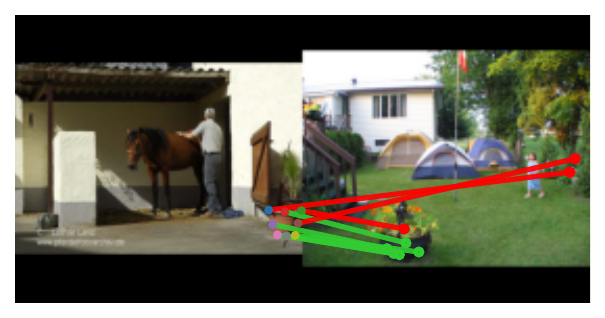}&
         \includegraphics[width=0.24\linewidth]{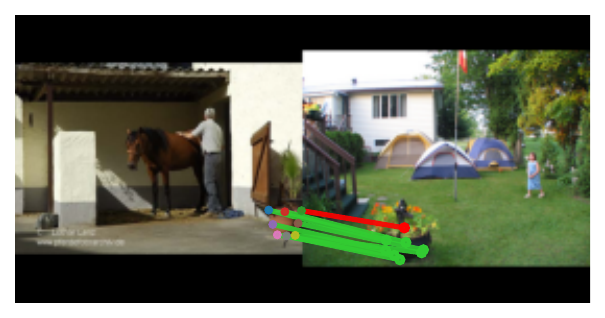} \\

         \includegraphics[width=0.24\linewidth]{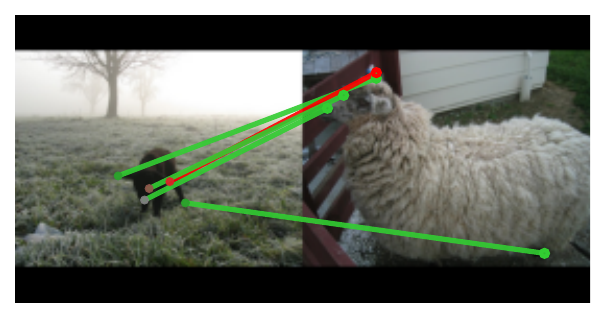}&
         \includegraphics[width=0.24\linewidth]{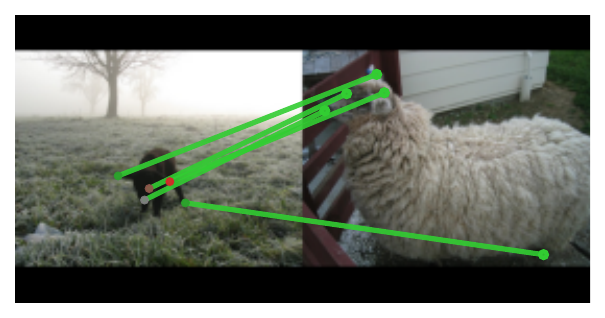} &
         \includegraphics[width=0.24\linewidth]{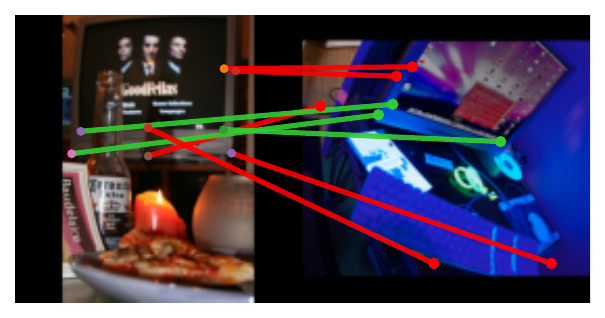}&
         \includegraphics[width=0.24\linewidth]{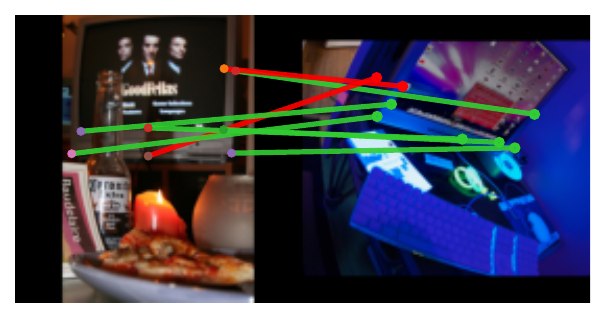} \\

         {\footnotesize TLR} & {\footnotesize TLR+Ours} & {\footnotesize TLR} & {\footnotesize TLR+Ours}\\
    \end{tabular}
    \caption{\textbf{Qualitative comparison on SPair-71k with TLR as the base method.} Correspondences are shown as colored lines: green for correct and red for incorrect matches (PCK@0.1). Our method (TLR+Ours) consistently reduces incorrect matches across all 18 categories.}
    \label{fig:tlr_spair_supp}
\end{figure*}

\begin{figure*}[tb]
    \centering
    \begin{tabular}{@{}c@{\hspace{0mm}}c@{\hspace{1mm}}|c@{\hspace{0mm}}c@{}}
         \includegraphics[width=0.24\linewidth]{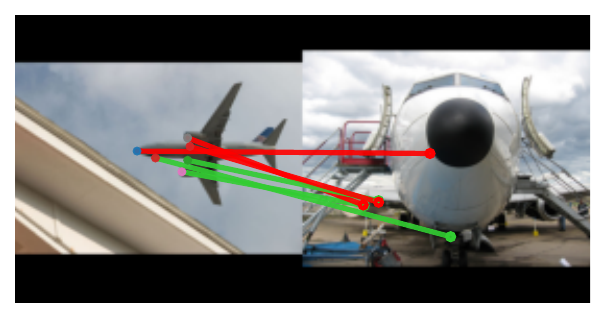} &
         \includegraphics[width=0.24\linewidth]{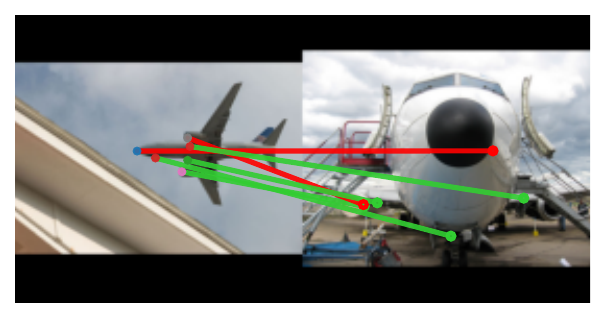} &
         \includegraphics[width=0.24\linewidth]{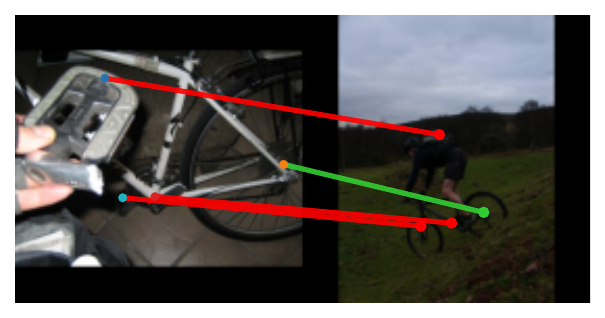} &
         \includegraphics[width=0.24\linewidth]{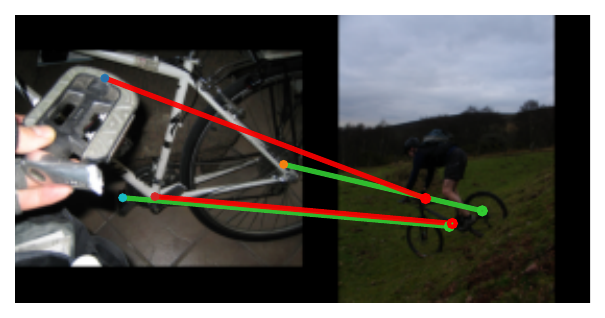} \\

         \includegraphics[width=0.24\linewidth]{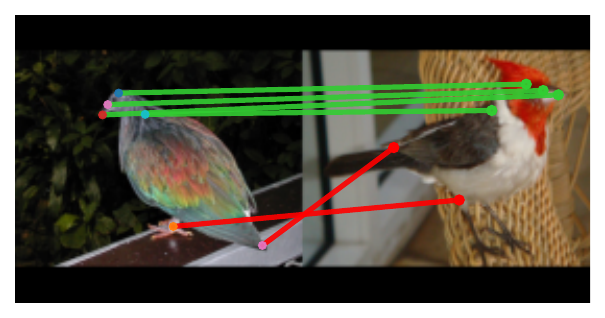} &
         \includegraphics[width=0.24\linewidth]{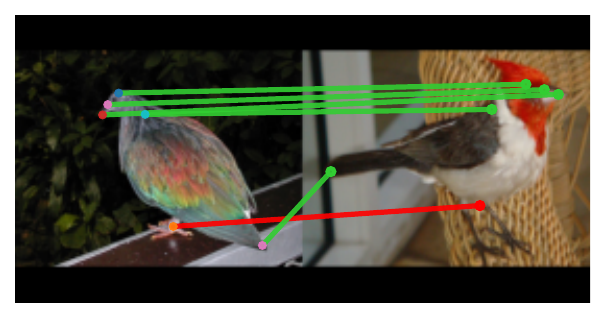} &
         \includegraphics[width=0.24\linewidth]{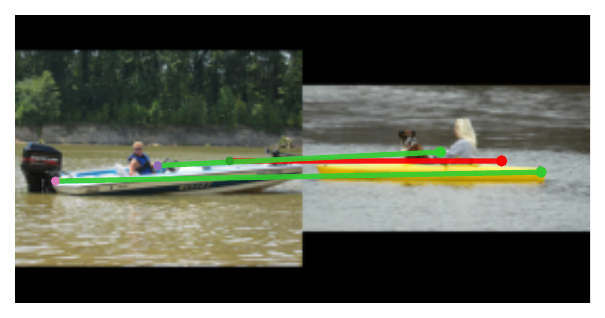} &
         \includegraphics[width=0.24\linewidth]{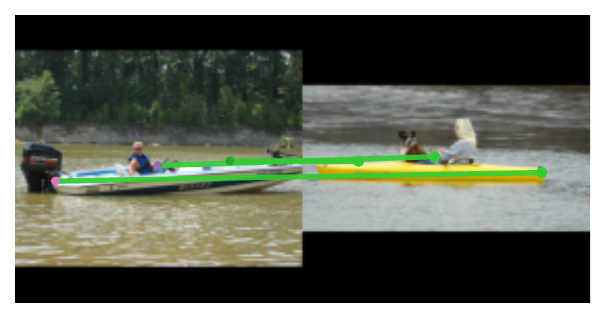} \\

         \includegraphics[width=0.24\linewidth]{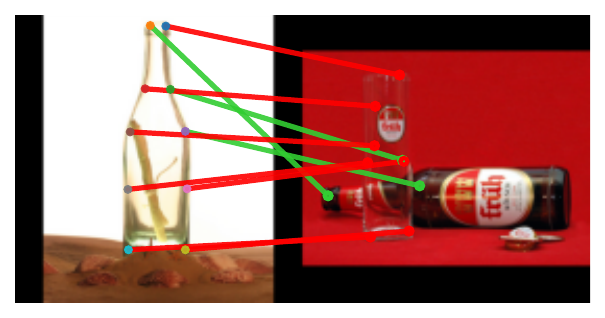} &
         \includegraphics[width=0.24\linewidth]{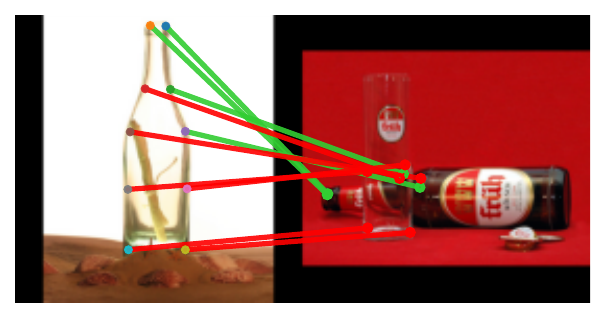} &
         \includegraphics[width=0.24\linewidth]{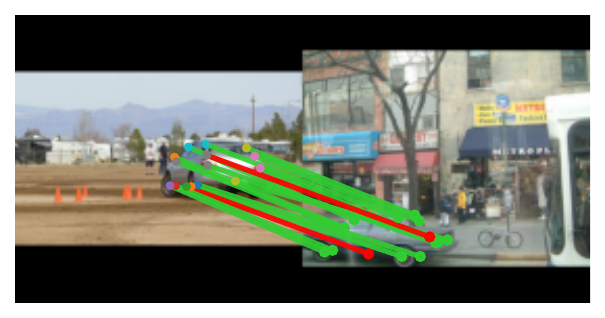} &
         \includegraphics[width=0.24\linewidth]{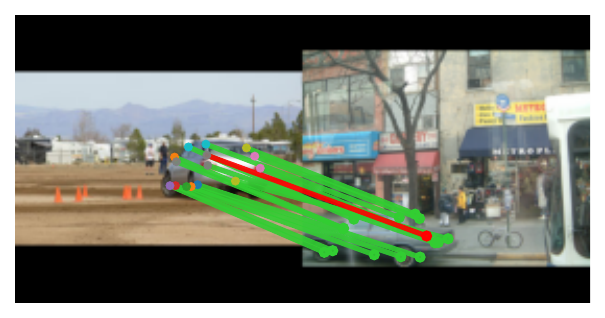} \\

         \includegraphics[width=0.24\linewidth]{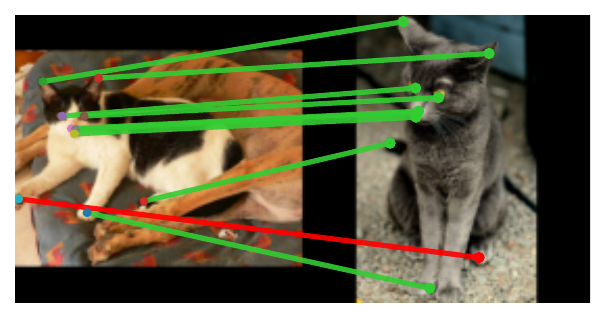} &
         \includegraphics[width=0.24\linewidth]{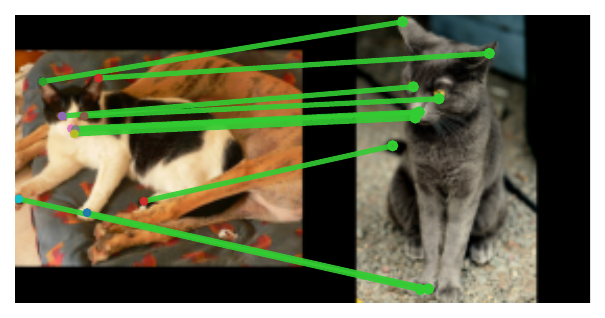} &
         \includegraphics[width=0.24\linewidth]{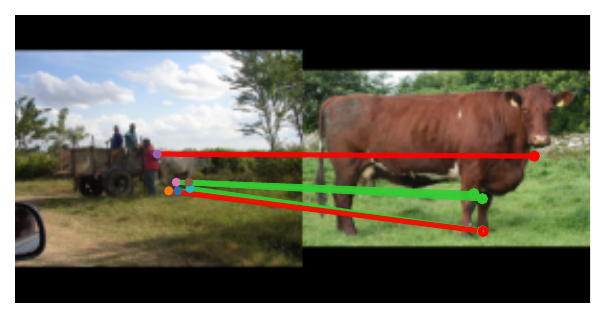} &
         \includegraphics[width=0.24\linewidth]{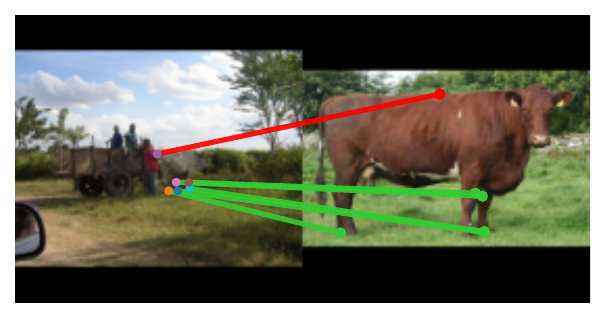} \\

         \includegraphics[width=0.24\linewidth]{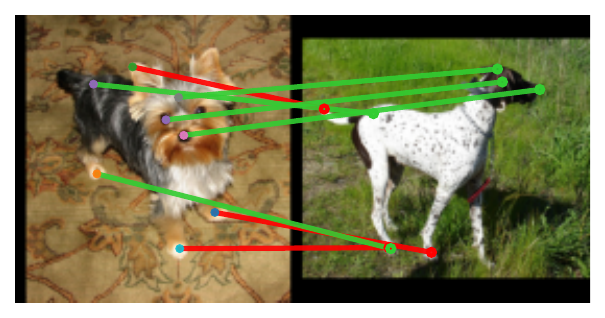} &
         \includegraphics[width=0.24\linewidth]{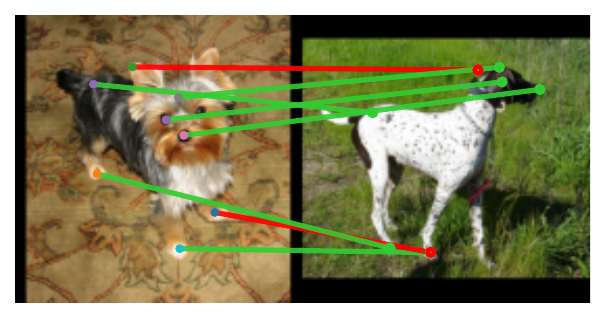} &
         \includegraphics[width=0.24\linewidth]{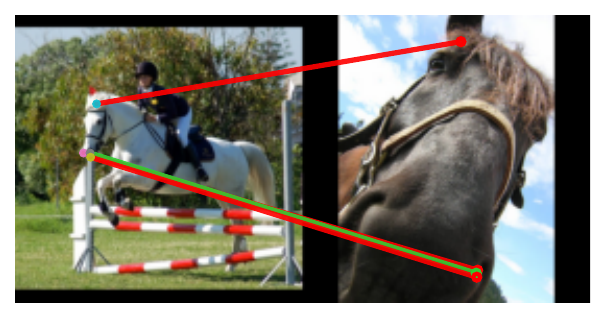} &
         \includegraphics[width=0.24\linewidth]{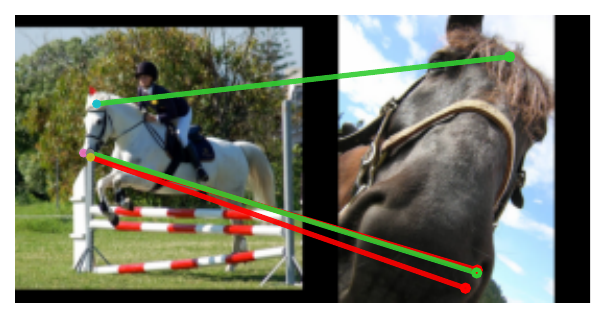} \\

         \includegraphics[width=0.24\linewidth]{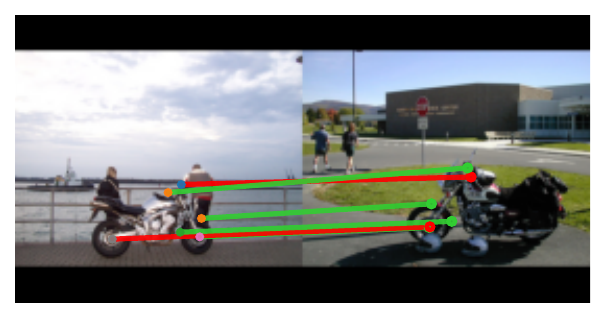} &
         \includegraphics[width=0.24\linewidth]{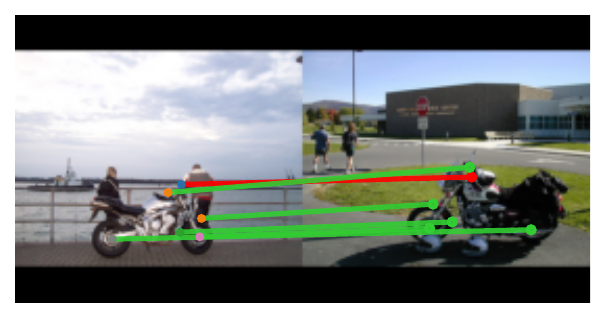} &
         \includegraphics[width=0.24\linewidth]{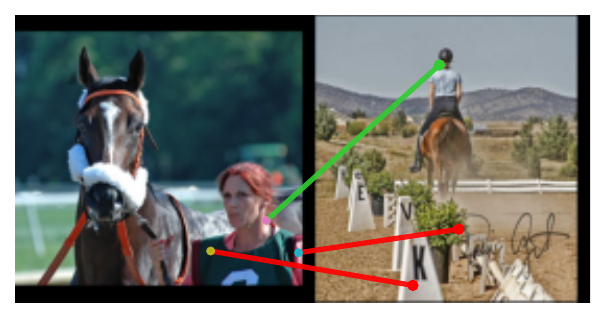} &
         \includegraphics[width=0.24\linewidth]{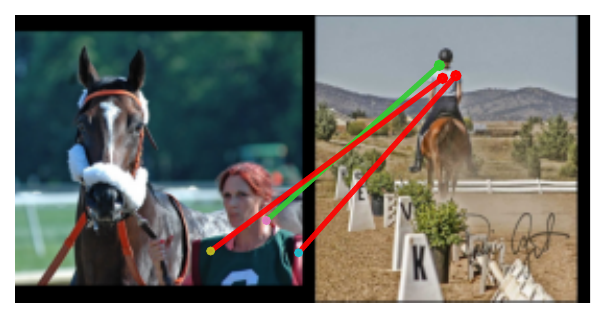} \\

         \includegraphics[width=0.24\linewidth]{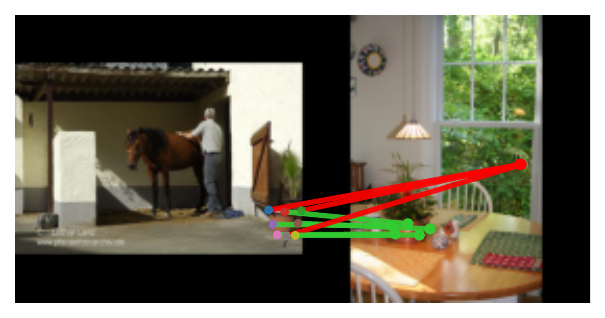} &
         \includegraphics[width=0.24\linewidth]{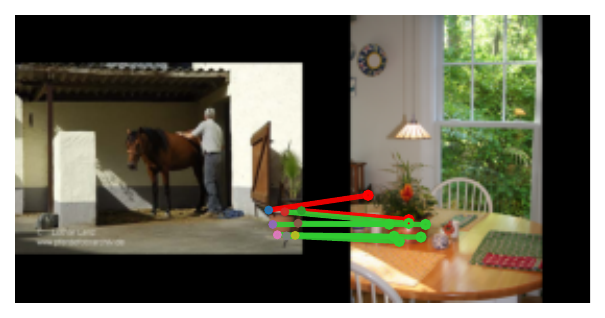} &
         \includegraphics[width=0.24\linewidth]{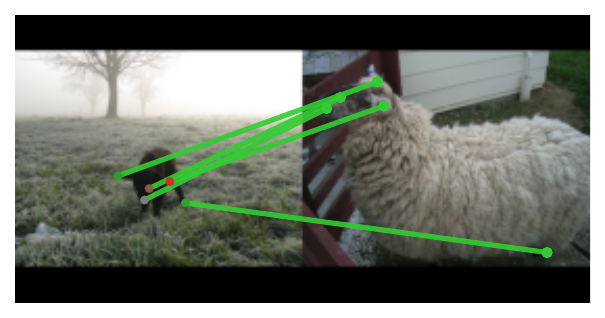} &
         \includegraphics[width=0.24\linewidth]{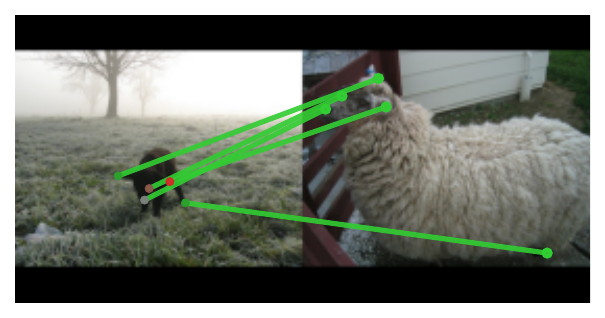} \\

         \includegraphics[width=0.24\linewidth]{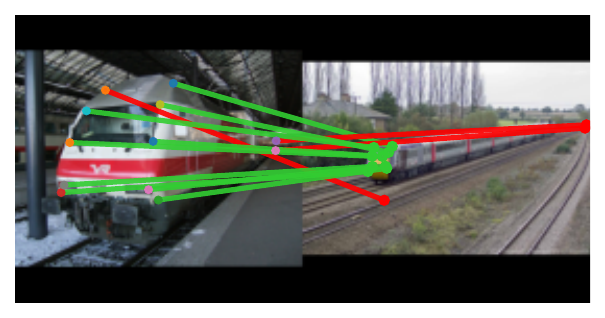} &
         \includegraphics[width=0.24\linewidth]{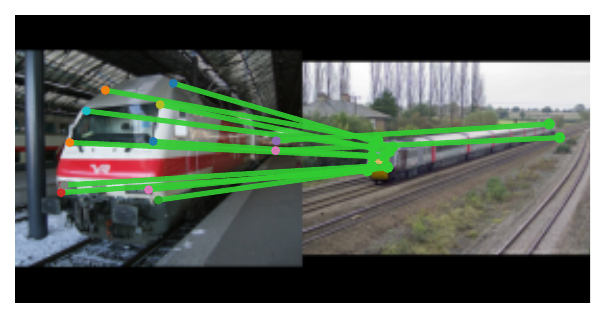} &
         \includegraphics[width=0.24\linewidth]{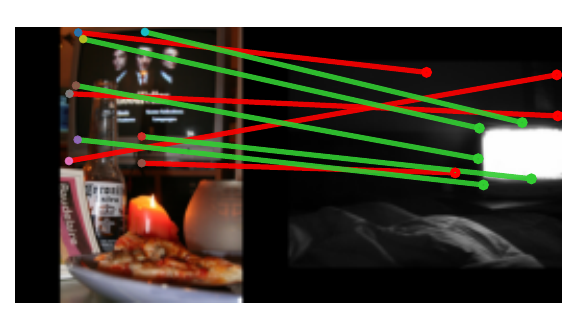} &
         \includegraphics[width=0.24\linewidth]{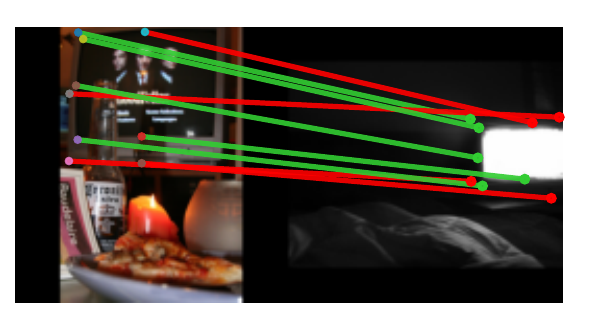} \\
         {\footnotesize CleanDIFT} & {\footnotesize CleanDIFT+Ours} & {\footnotesize CleanDIFT} & {\footnotesize CleanDIFT+Ours} \\
    \end{tabular}
    \caption{\textbf{Qualitative comparison on SPair-71k with CleanDIFT as the base method.} Correspondences are shown as colored lines: green for correct and red for incorrect matches (PCK@0.1). Our method (CleanDIFT+Ours) consistently produces fewer incorrect matches across all 18 categories.}
    \label{fig:cleandift_spair_supp}
\end{figure*}

\begin{figure*}[tb]
    \centering
    \begin{tabular}{@{}c@{\hspace{0mm}}c@{\hspace{1mm}}|c@{\hspace{0mm}}c@{}}
         \includegraphics[width=0.24\linewidth]{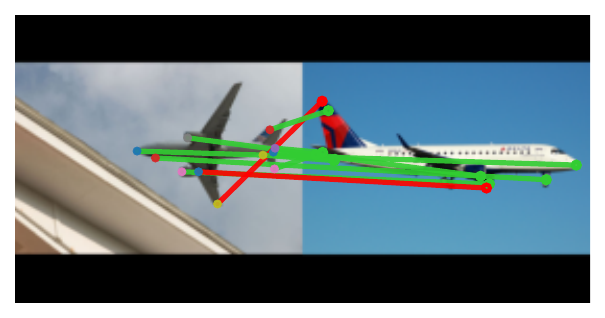}&  \includegraphics[width=0.24\linewidth]{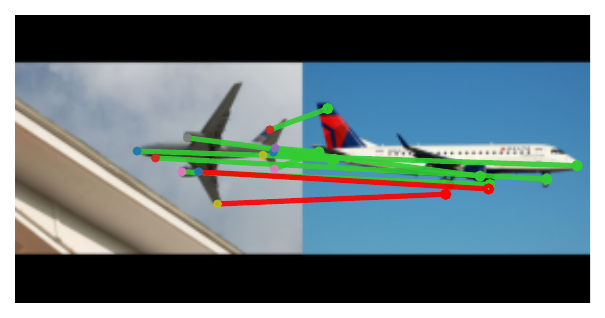} & \includegraphics[width=0.24\linewidth]{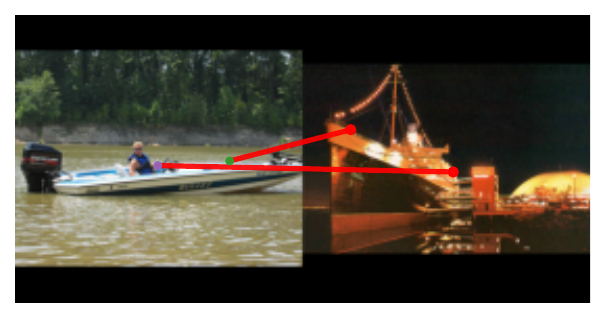}&  \includegraphics[width=0.24\linewidth]{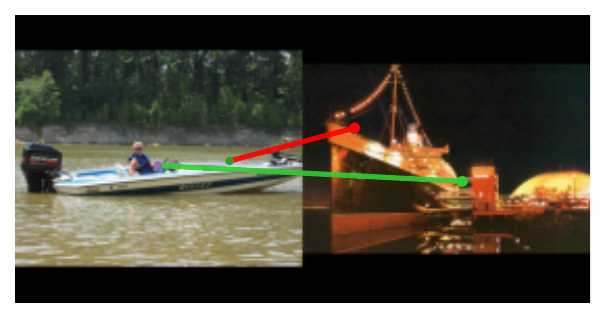} \\
         \includegraphics[width=0.24\linewidth]{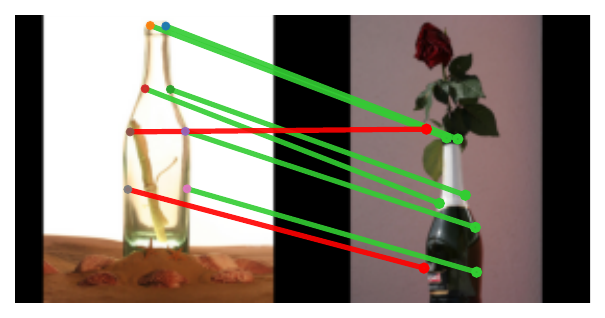}&  \includegraphics[width=0.24\linewidth]{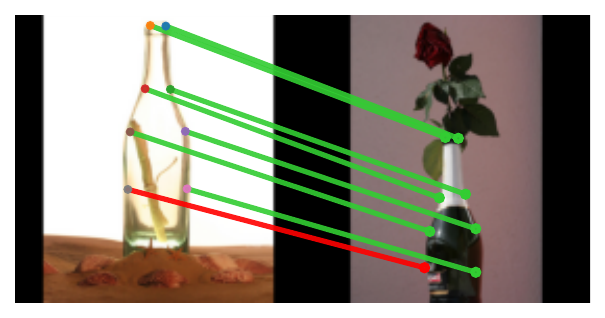} & \includegraphics[width=0.24\linewidth]{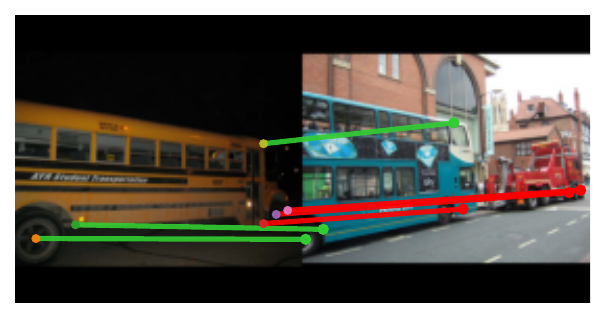}&  \includegraphics[width=0.24\linewidth]{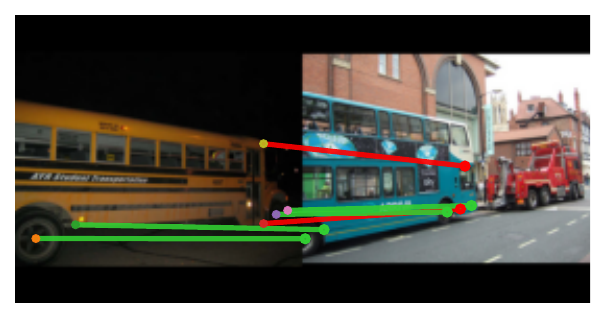} \\
         \includegraphics[width=0.24\linewidth]{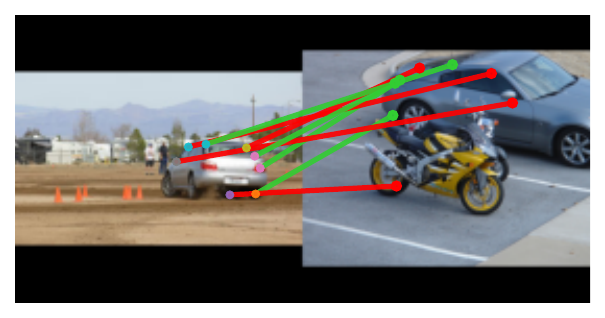}&  \includegraphics[width=0.24\linewidth]{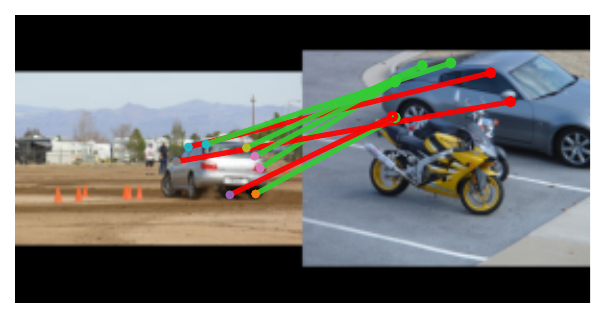} & \includegraphics[width=0.24\linewidth]{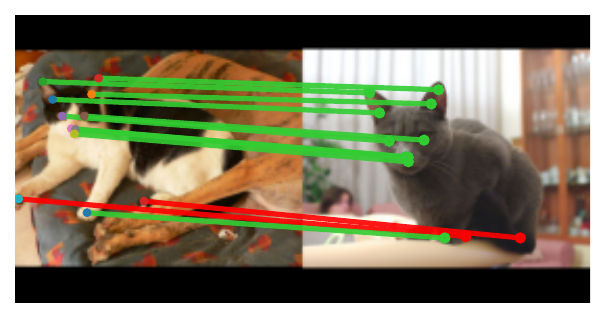}&  \includegraphics[width=0.24\linewidth]{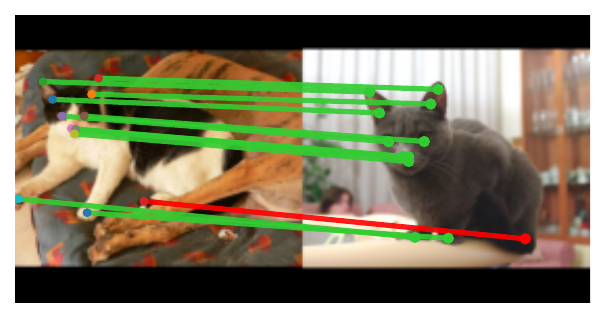} \\
         \includegraphics[width=0.24\linewidth]{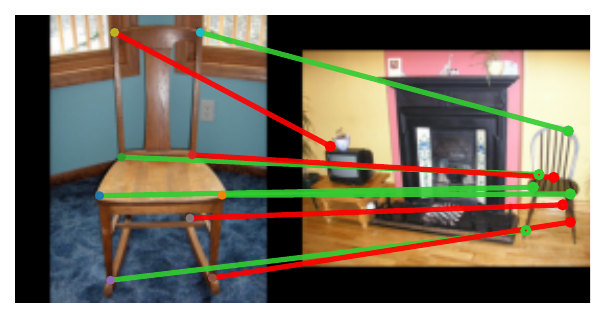}&  \includegraphics[width=0.24\linewidth]{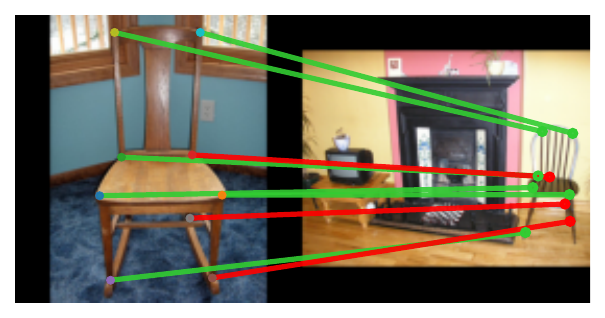} & \includegraphics[width=0.24\linewidth]{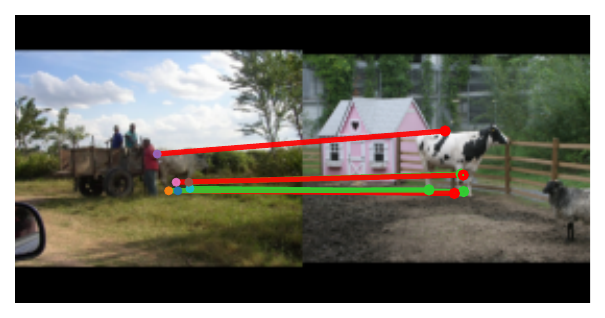}&  \includegraphics[width=0.24\linewidth]{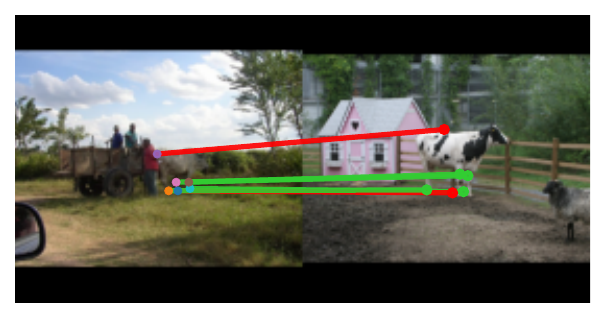} \\
         \includegraphics[width=0.24\linewidth]{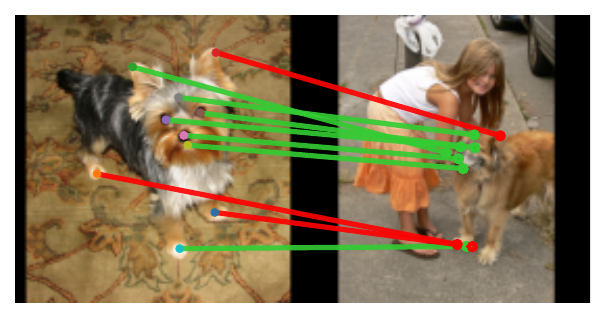}&  \includegraphics[width=0.24\linewidth]{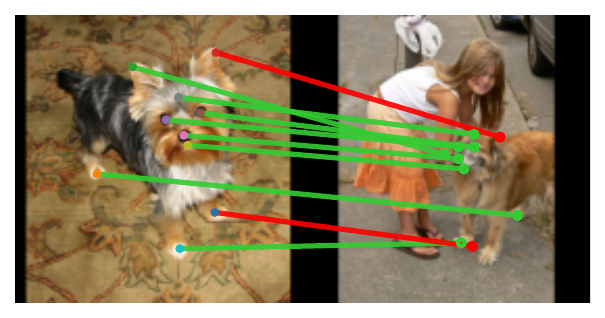} & \includegraphics[width=0.24\linewidth]{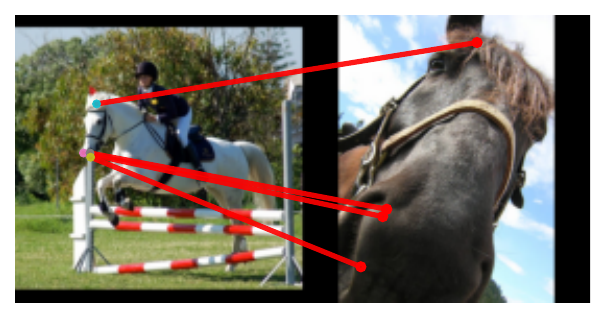}&  \includegraphics[width=0.24\linewidth]{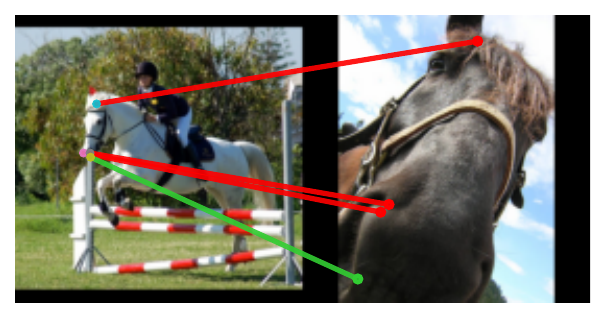} \\
         \includegraphics[width=0.24\linewidth]{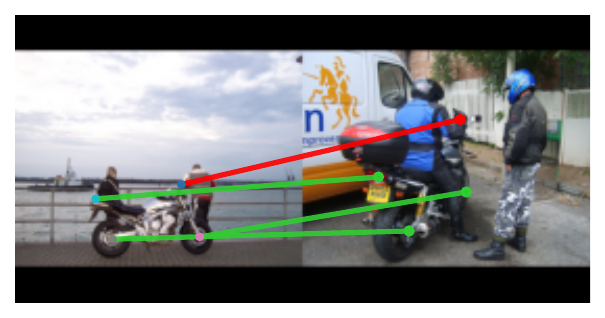}&  \includegraphics[width=0.24\linewidth]{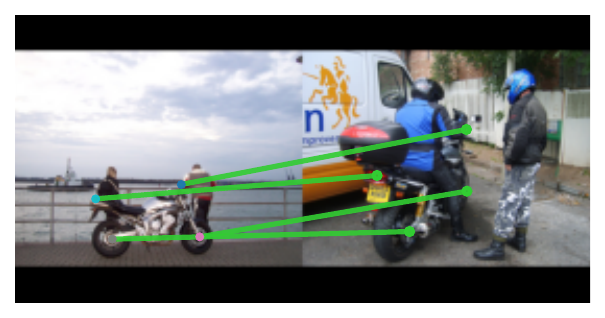} & \includegraphics[width=0.24\linewidth]{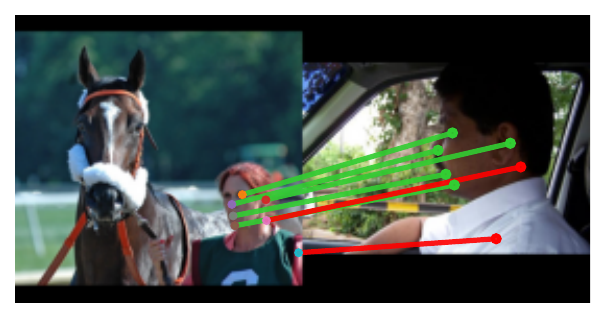}&  \includegraphics[width=0.24\linewidth]{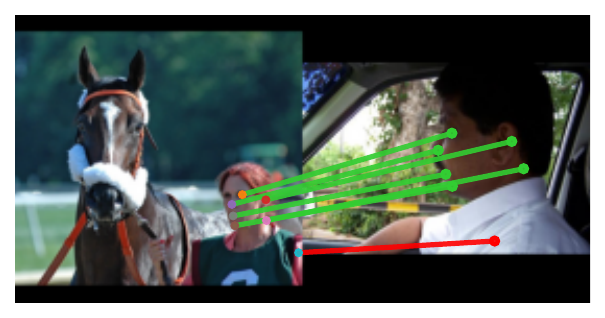} \\
         \includegraphics[width=0.24\linewidth]{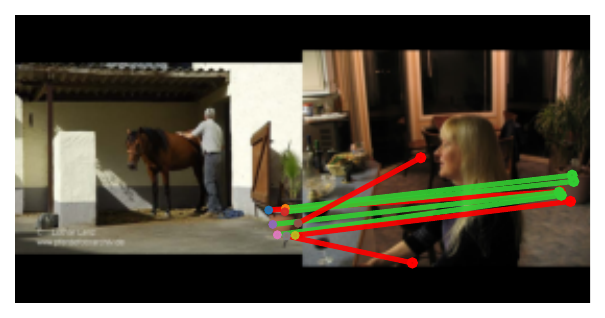}&  \includegraphics[width=0.24\linewidth]{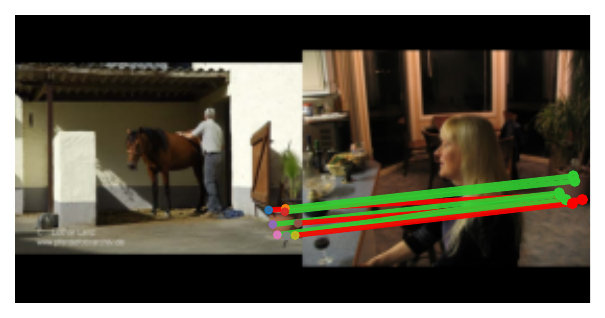} & \includegraphics[width=0.24\linewidth]{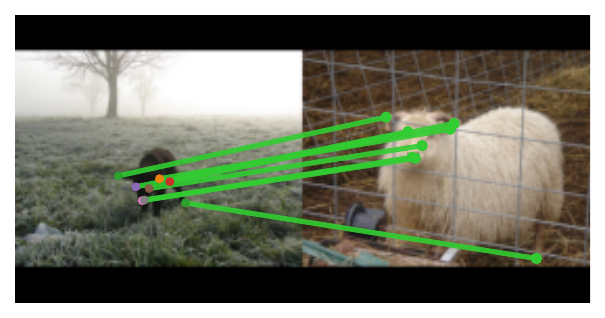}&  \includegraphics[width=0.24\linewidth]{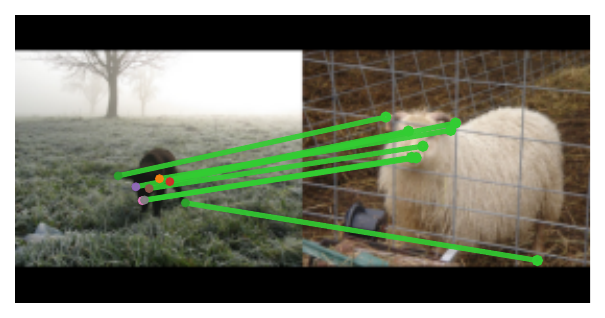} \\
         \includegraphics[width=0.24\linewidth]{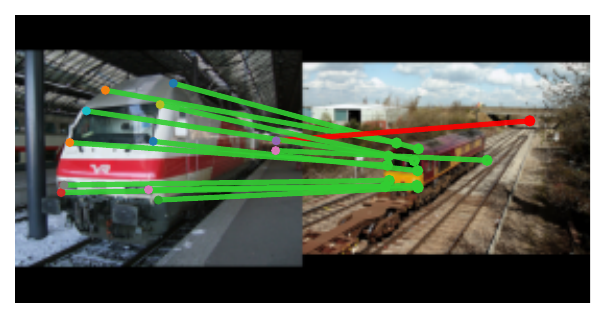}&  \includegraphics[width=0.24\linewidth]{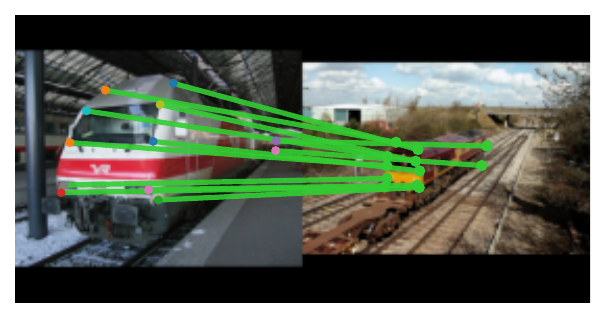} & \includegraphics[width=0.24\linewidth]{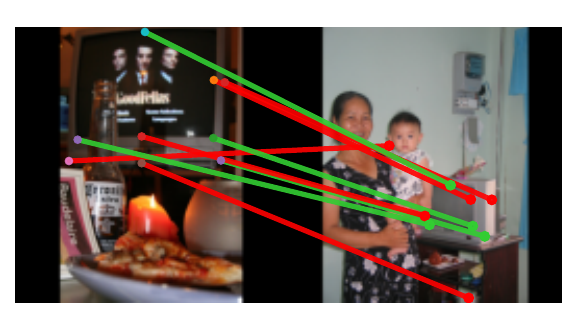}&  \includegraphics[width=0.24\linewidth]{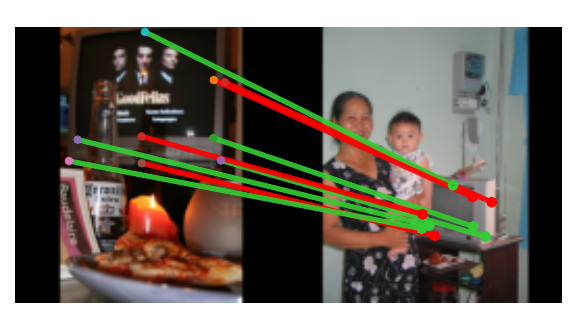} \\
         {\footnotesize DIY-SC} & {\footnotesize DIY-SC+Ours} & {\footnotesize DIY-SC} & {\footnotesize DIY-SC+Ours}\\
    \end{tabular}
    \caption{\textbf{Qualitative comparison on SPair-71k with DIY-SC as the base method.} Correspondences are shown as colored lines: green for correct and red for incorrect matches (PCK@0.1). Our method (DIY-SC+Ours) consistently improves correspondence quality across diverse object categories.}
    \label{fig:diy_spair_supp}
\end{figure*}

\Cref{fig:tlr_spair_supp,fig:cleandift_spair_supp,fig:diy_spair_supp} provide full per-category qualitative comparisons on SPair-71k for all three base methods. Each row shows one representative image pair per category; green lines indicate correct matches (PCK@0.1) and red lines indicate incorrect ones. Across all 18 categories and all three base methods, our re-ranking consistently reduces the number of incorrect matches, with particularly visible improvements in categories identified as high-gain in \Cref{sec:category_analysis} (tv, plant, chair).

\newpage

\newpage

\newpage

\subsection{Additional Qualitative Comparisons on PF-Pascal}

\begin{figure*}[tb]
    \centering
    \begin{tabular}{@{}c@{\hspace{0mm}}c@{\hspace{1mm}}|c@{\hspace{0mm}}c@{}}
         \includegraphics[width=0.24\linewidth]{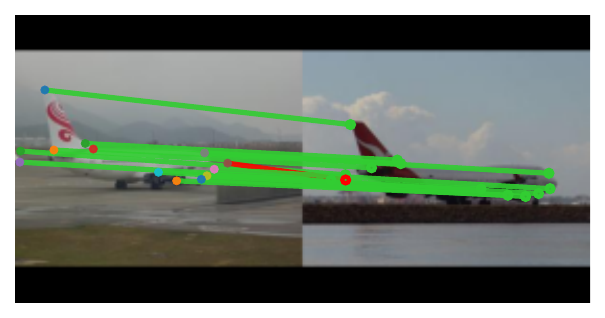}&  \includegraphics[width=0.24\linewidth]{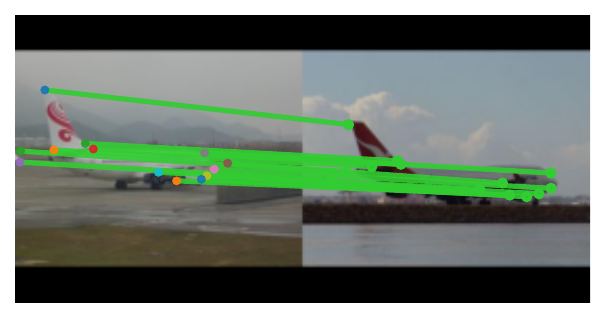} & \includegraphics[width=0.24\linewidth]{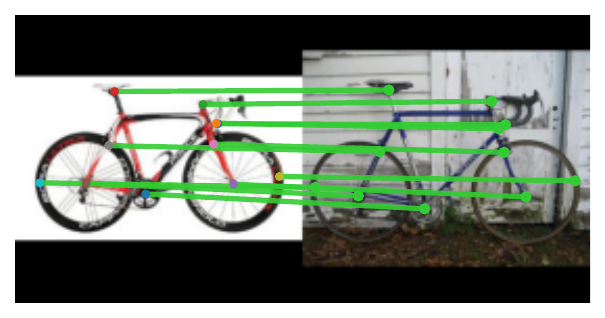}&  \includegraphics[width=0.24\linewidth]{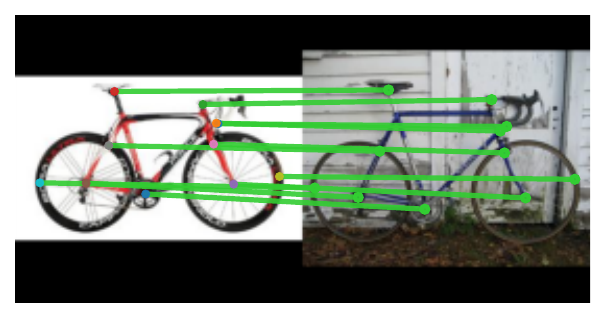} \\
         \includegraphics[width=0.24\linewidth]{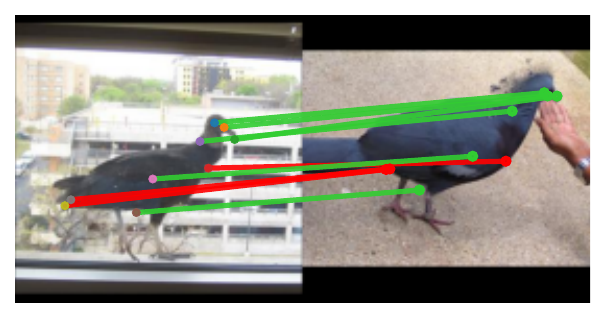}&  \includegraphics[width=0.24\linewidth]{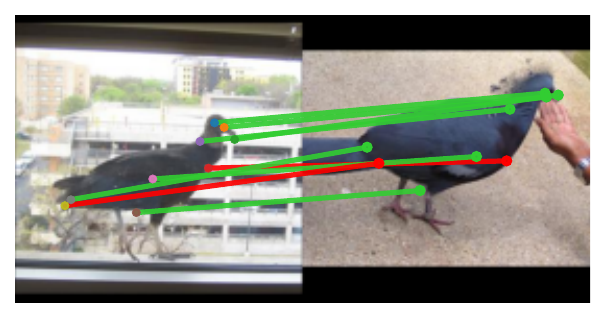} & \includegraphics[width=0.24\linewidth]{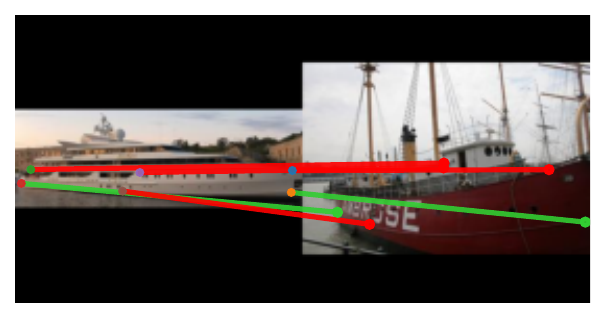}&  \includegraphics[width=0.24\linewidth]{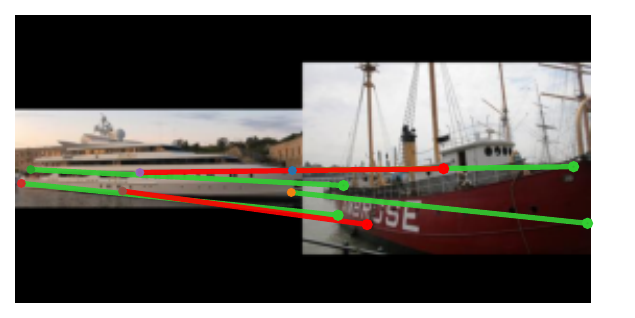} \\
         \includegraphics[width=0.24\linewidth]{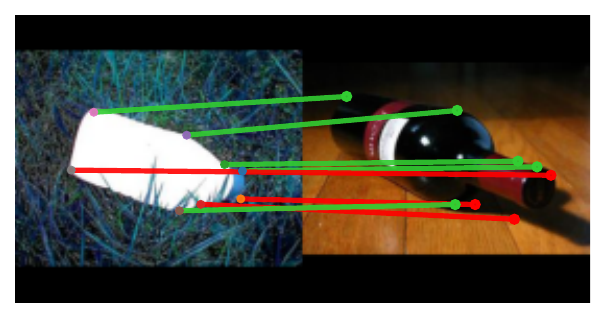}&  \includegraphics[width=0.24\linewidth]{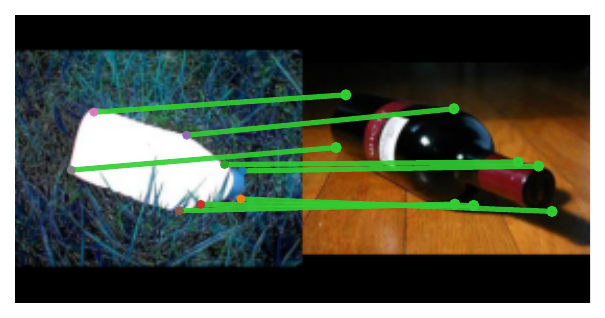} & \includegraphics[width=0.24\linewidth]{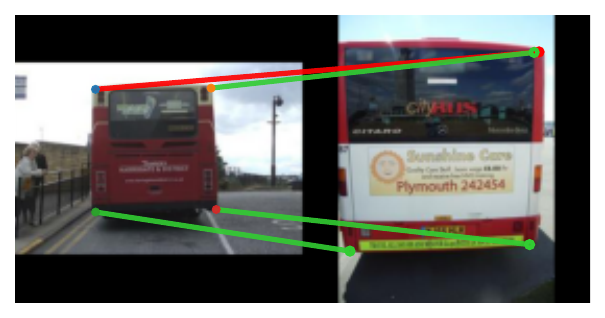}&  \includegraphics[width=0.24\linewidth]{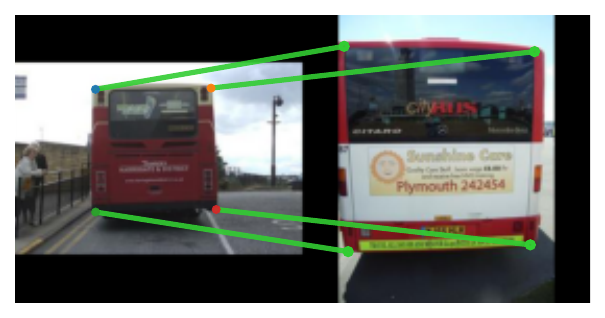} \\
         \includegraphics[width=0.24\linewidth]{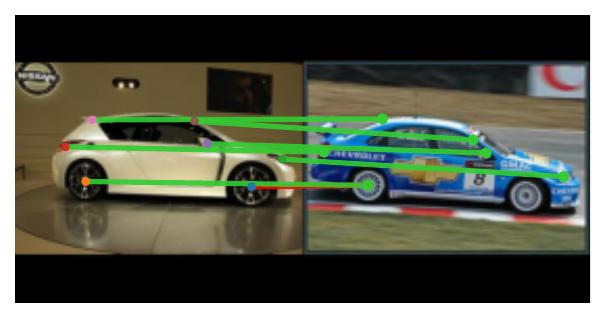}&  \includegraphics[width=0.24\linewidth]{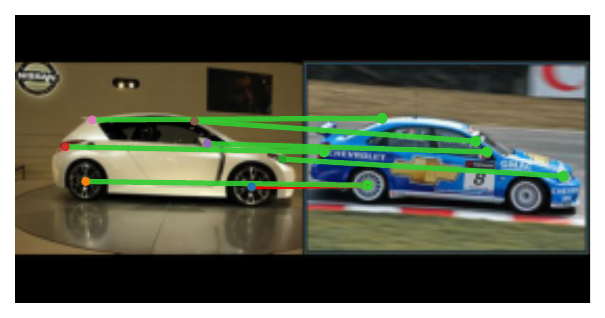} & \includegraphics[width=0.24\linewidth]{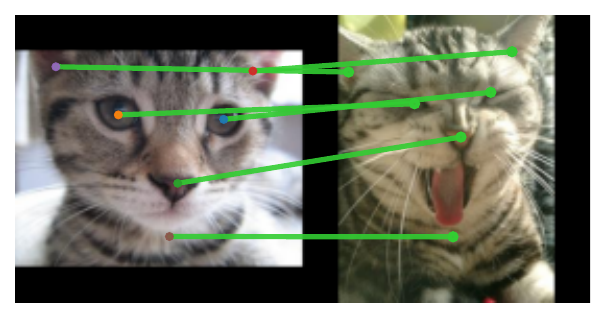}&  \includegraphics[width=0.24\linewidth]{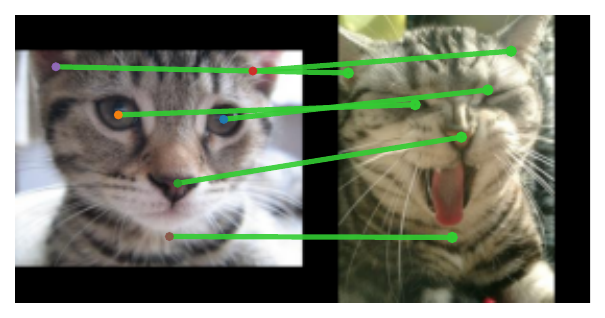} \\
         \includegraphics[width=0.24\linewidth]{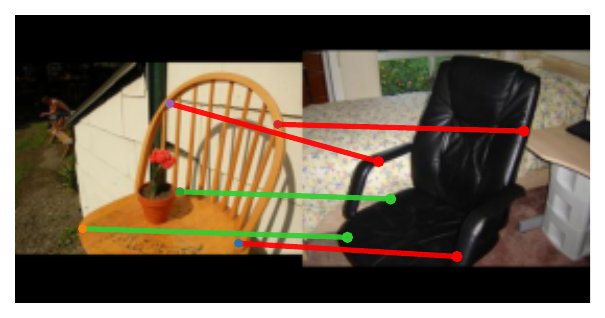}&  \includegraphics[width=0.24\linewidth]{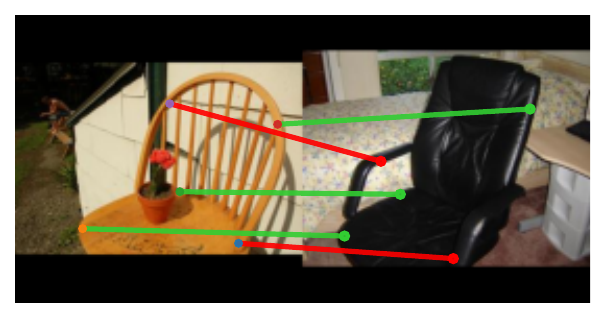} & \includegraphics[width=0.24\linewidth]{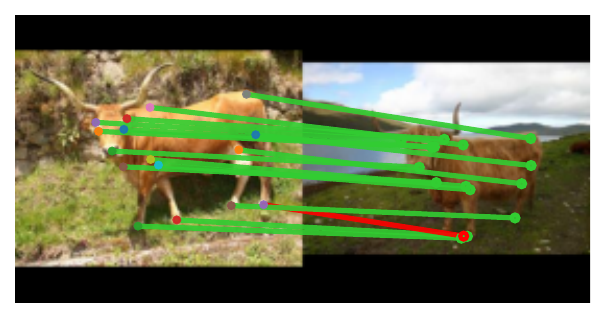}&  \includegraphics[width=0.24\linewidth]{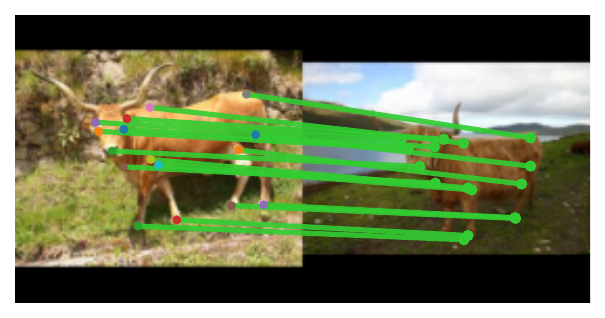} \\
         \includegraphics[width=0.24\linewidth]{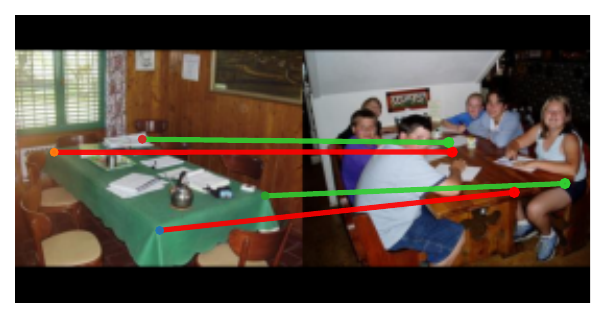}&  \includegraphics[width=0.24\linewidth]{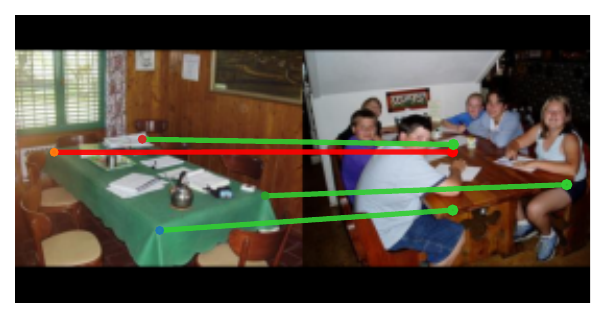} & \includegraphics[width=0.24\linewidth]{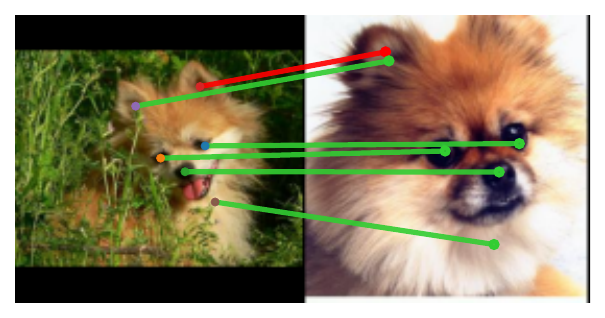}&  \includegraphics[width=0.24\linewidth]{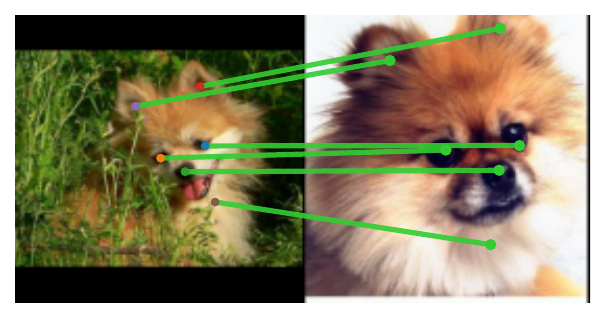} \\
         \includegraphics[width=0.24\linewidth]{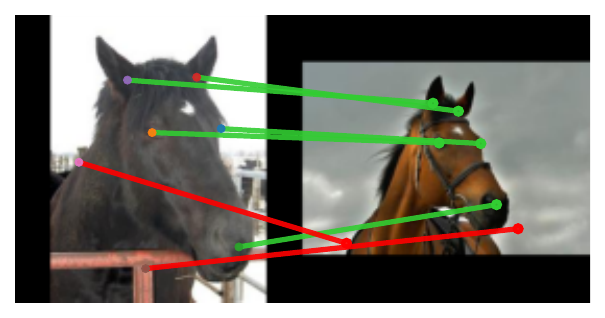}&  \includegraphics[width=0.24\linewidth]{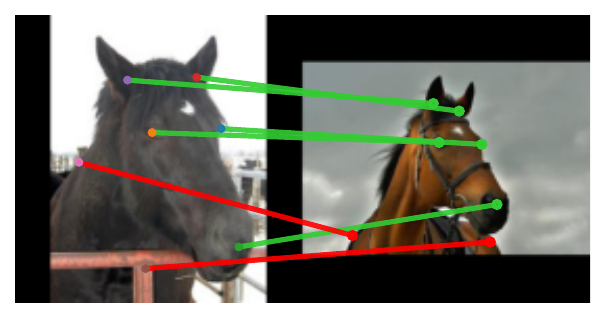} & \includegraphics[width=0.24\linewidth]{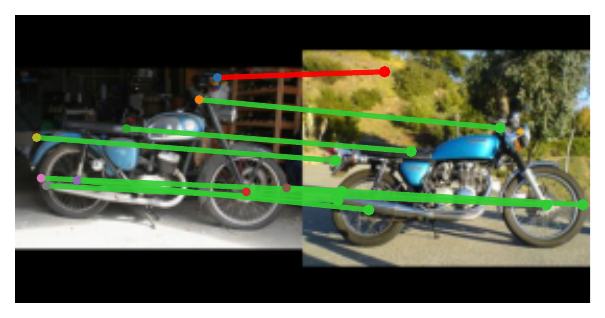}&  \includegraphics[width=0.24\linewidth]{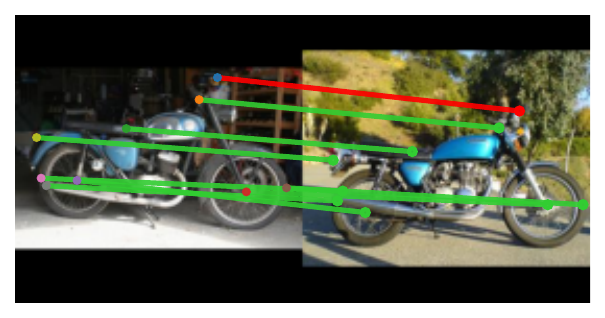} \\
         \includegraphics[width=0.24\linewidth]{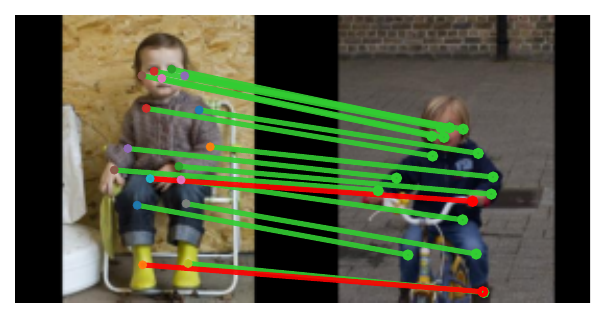}&  \includegraphics[width=0.24\linewidth]{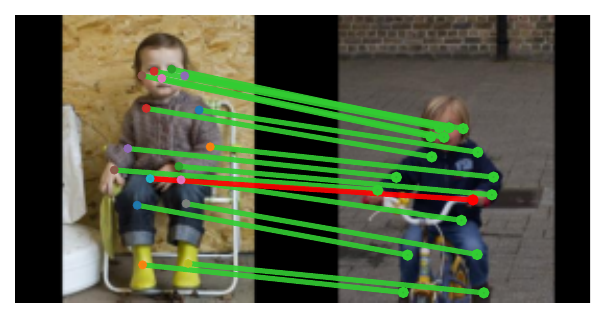} & \includegraphics[width=0.24\linewidth]{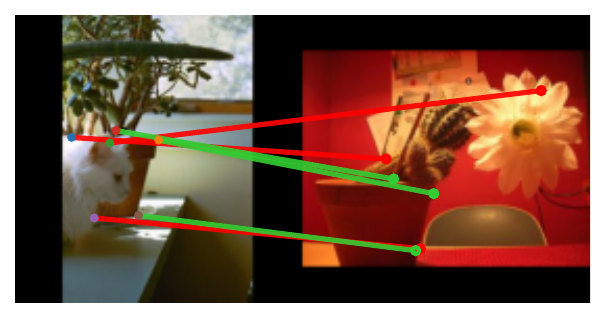}&  \includegraphics[width=0.24\linewidth]{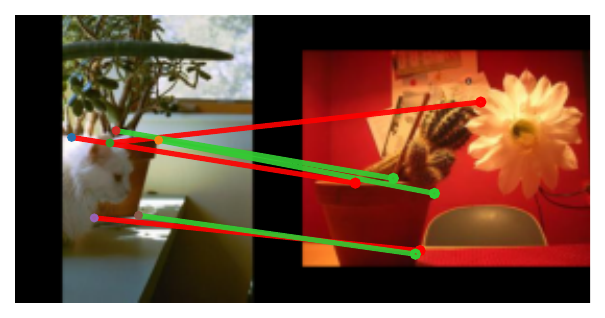} \\
         \includegraphics[width=0.24\linewidth]{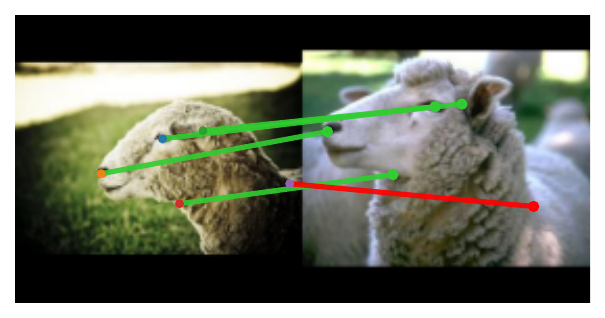}&  \includegraphics[width=0.24\linewidth]{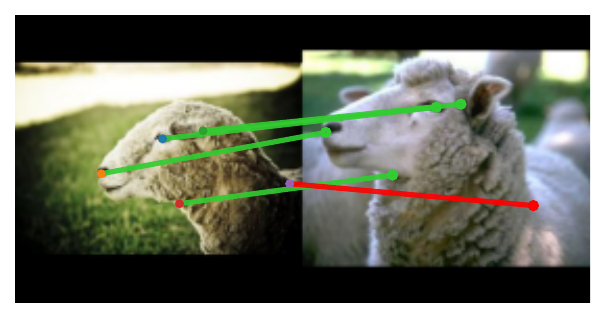} & \includegraphics[width=0.24\linewidth]{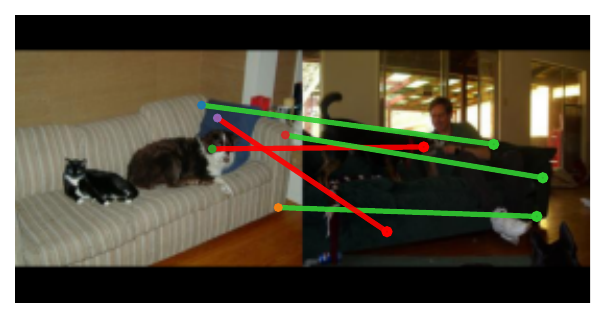}&  \includegraphics[width=0.24\linewidth]{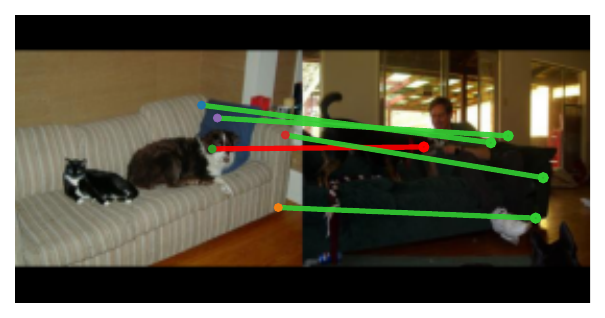} \\
         \includegraphics[width=0.24\linewidth]{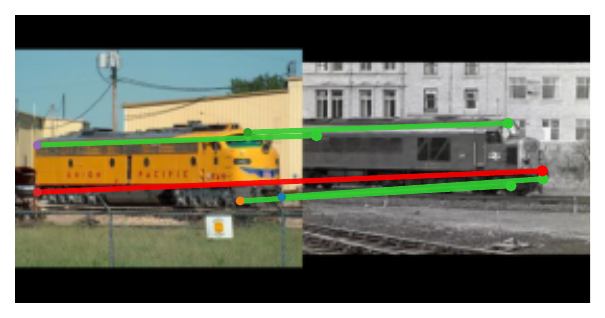}&  \includegraphics[width=0.24\linewidth]{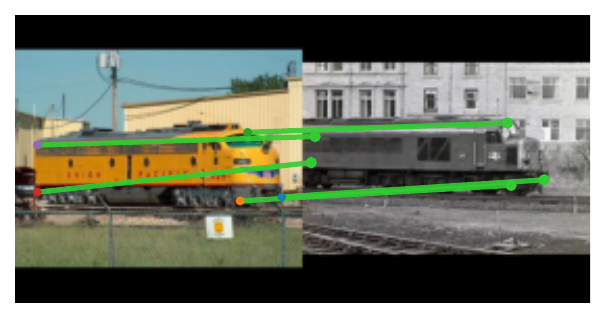} & \includegraphics[width=0.24\linewidth]{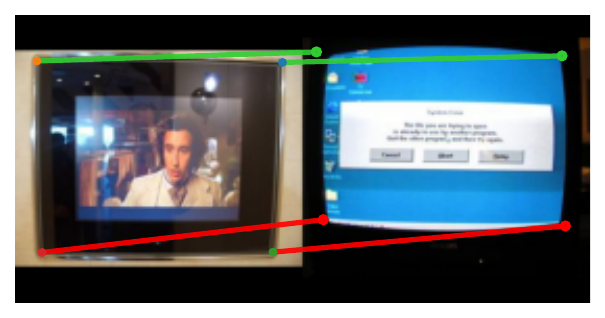}&  \includegraphics[width=0.24\linewidth]{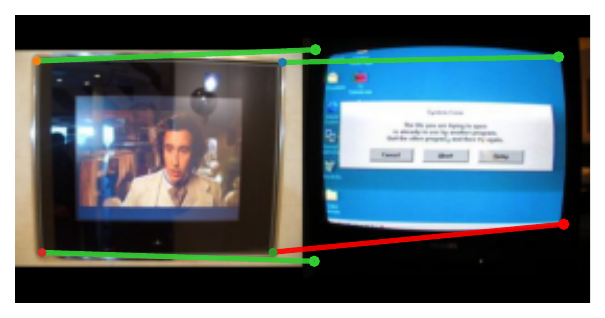} \\
         {\footnotesize TLR} & {\footnotesize TLR+Ours} & {\footnotesize TLR} & {\footnotesize TLR+Ours}\\
    \end{tabular}
    \caption{\textbf{Qualitative comparison on PF-Pascal with TLR as the base method.} Correspondences are shown as colored lines: green for correct and red for incorrect matches (PCK@0.1). Our method (TLR+Ours) yields more accurate correspondences across all 20 PF-Pascal categories.}
    \label{fig:tlr_pascal_supp}
\end{figure*}

\begin{figure*}[tb]
    \centering
    \begin{tabular}{@{}c@{\hspace{0mm}}c@{\hspace{1mm}}|c@{\hspace{0mm}}c@{}}
         \includegraphics[width=0.24\linewidth]{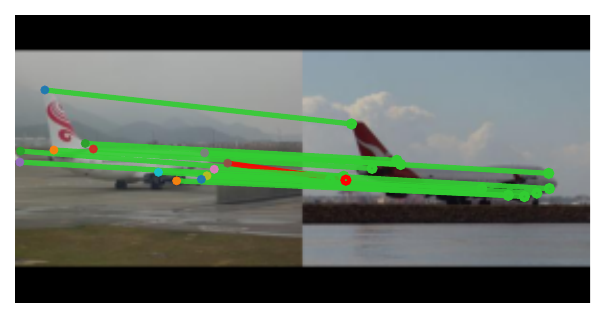}&  \includegraphics[width=0.24\linewidth]{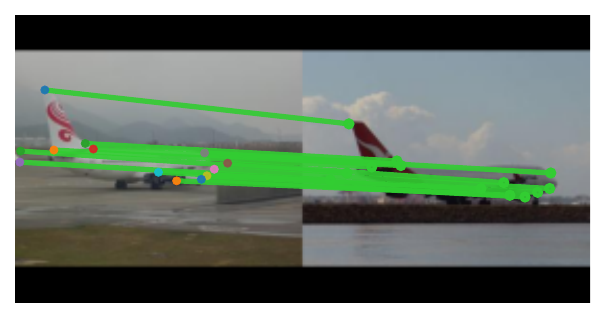} & \includegraphics[width=0.24\linewidth]{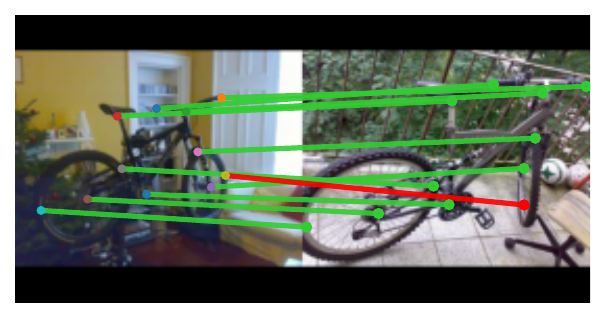}&  \includegraphics[width=0.24\linewidth]{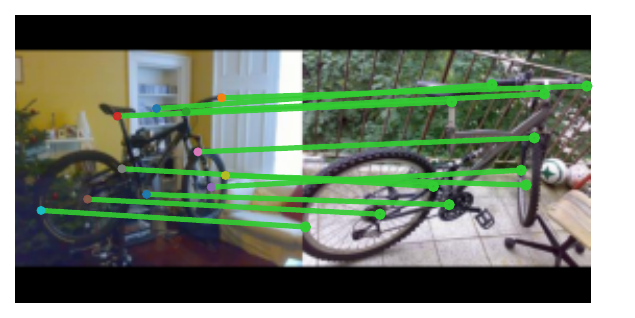} \\
         \includegraphics[width=0.24\linewidth]{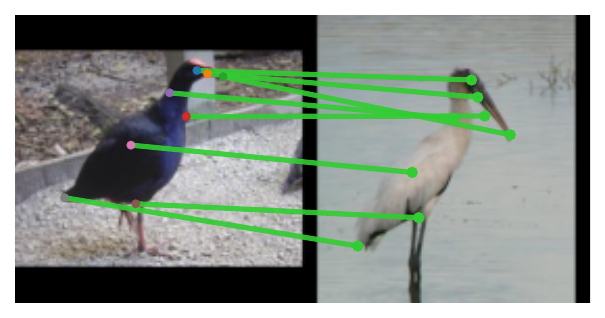}&  \includegraphics[width=0.24\linewidth]{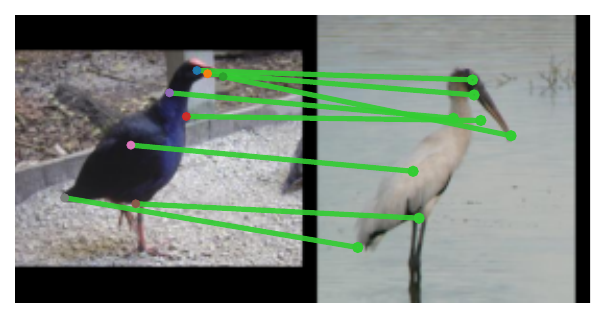} & \includegraphics[width=0.24\linewidth]{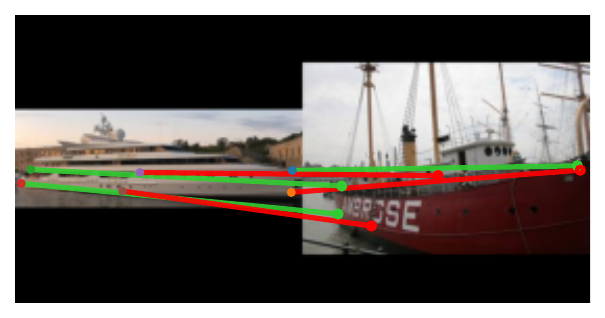}&  \includegraphics[width=0.24\linewidth]{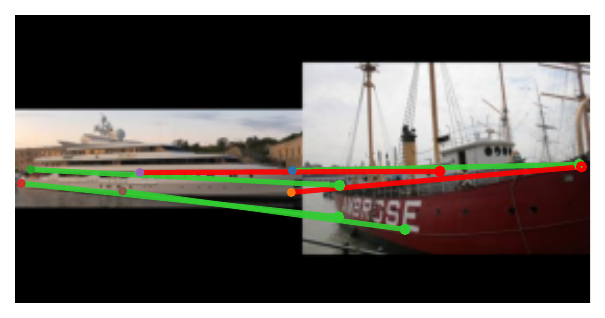} \\
         \includegraphics[width=0.24\linewidth]{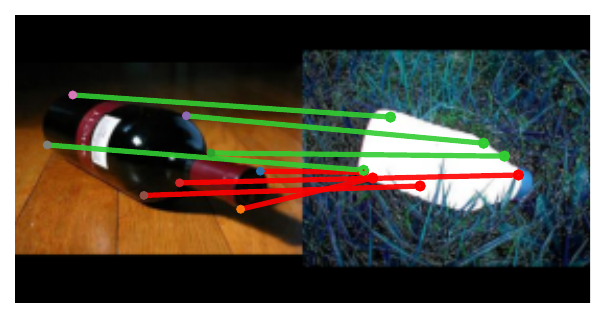}&  \includegraphics[width=0.24\linewidth]{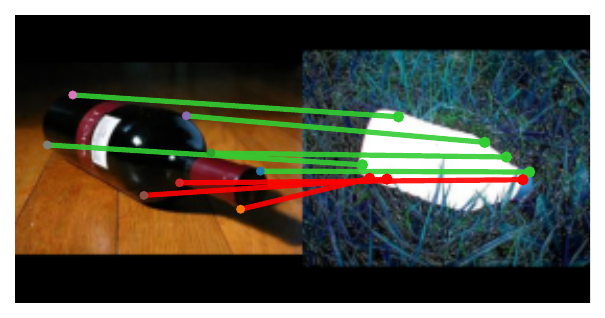} & \includegraphics[width=0.24\linewidth]{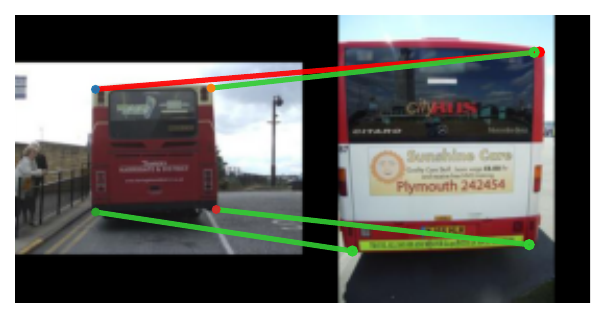}&  \includegraphics[width=0.24\linewidth]{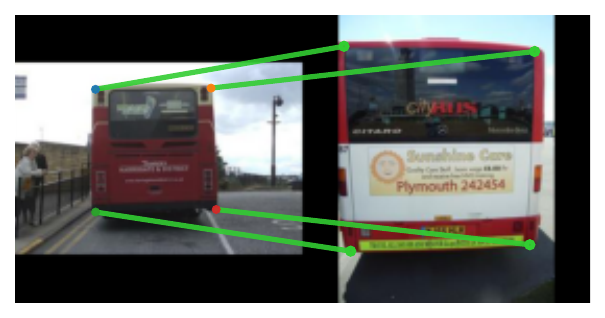} \\
         \includegraphics[width=0.24\linewidth]{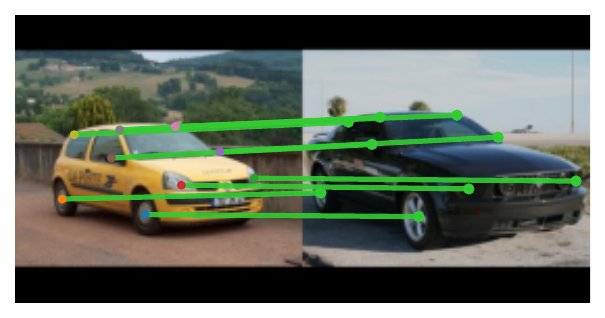}&  \includegraphics[width=0.24\linewidth]{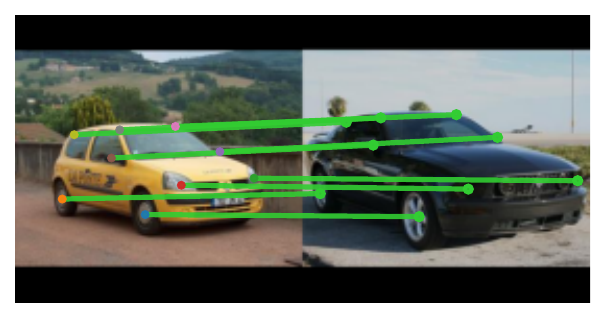} & \includegraphics[width=0.24\linewidth]{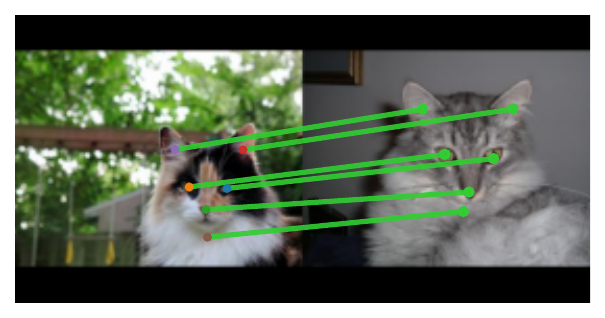}&  \includegraphics[width=0.24\linewidth]{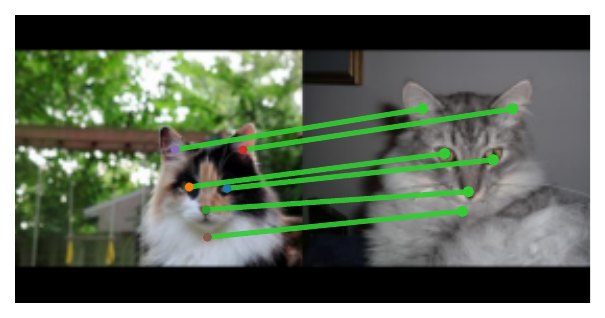} \\
         \includegraphics[width=0.24\linewidth]{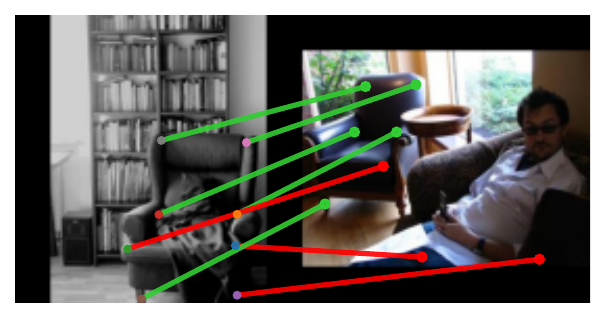}&  \includegraphics[width=0.24\linewidth]{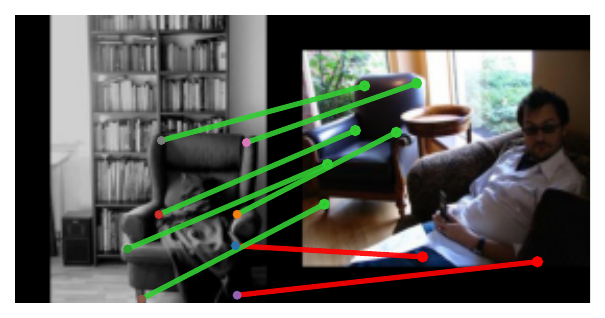} & \includegraphics[width=0.24\linewidth]{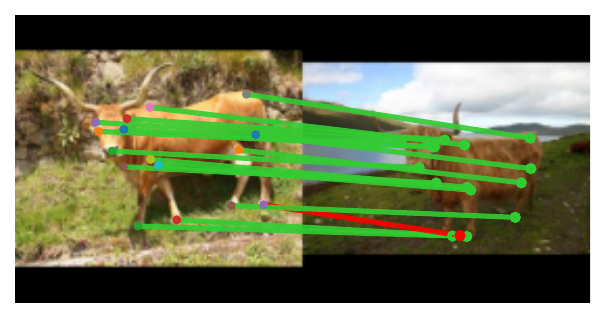}&  \includegraphics[width=0.24\linewidth]{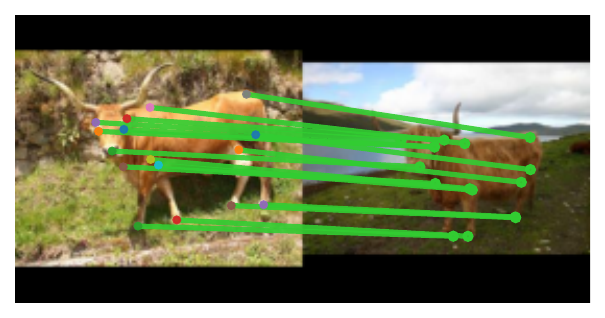} \\
         \includegraphics[width=0.24\linewidth]{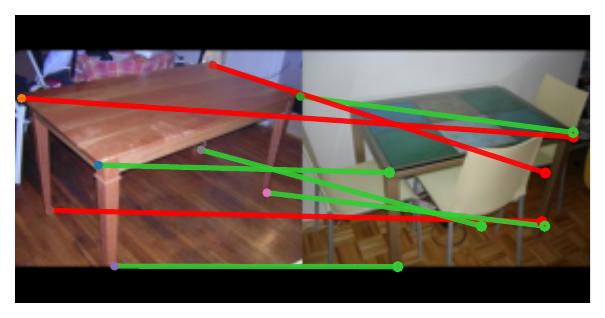}&  \includegraphics[width=0.24\linewidth]{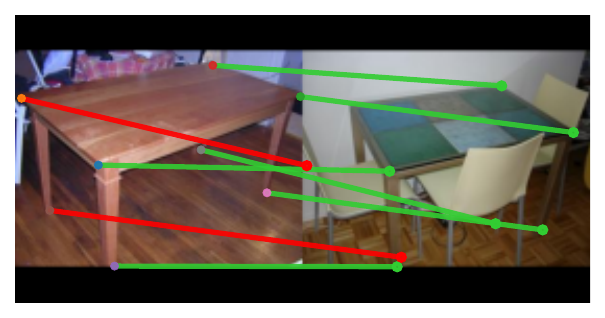} & \includegraphics[width=0.24\linewidth]{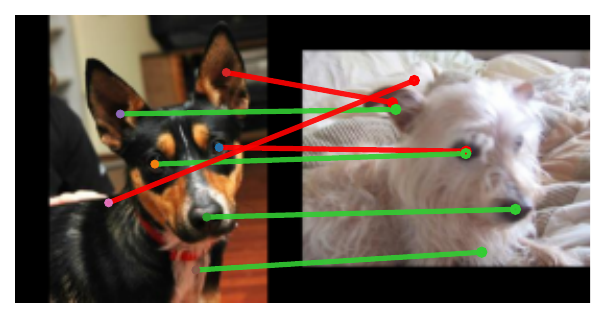}&  \includegraphics[width=0.24\linewidth]{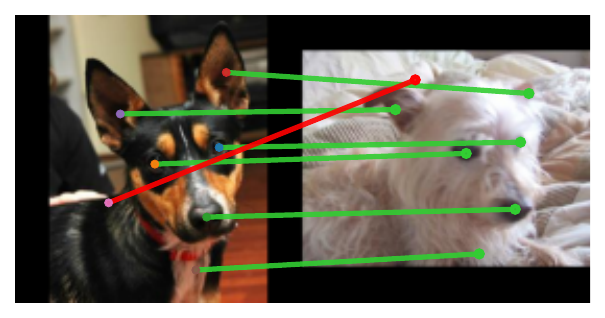} \\
         \includegraphics[width=0.24\linewidth]{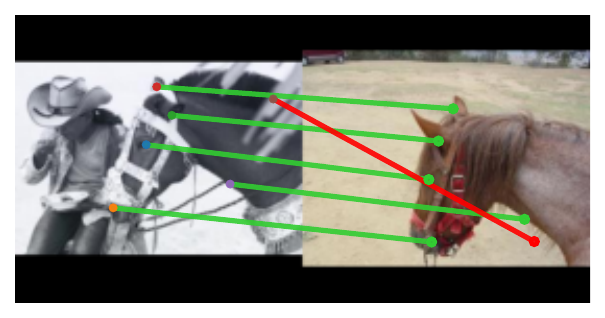}&  \includegraphics[width=0.24\linewidth]{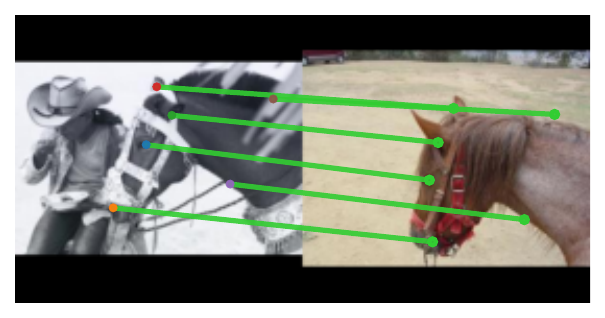} & \includegraphics[width=0.24\linewidth]{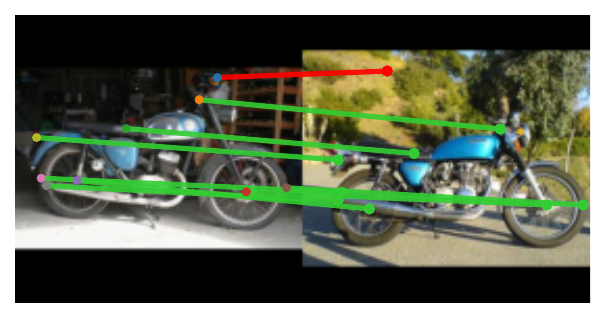}&  \includegraphics[width=0.24\linewidth]{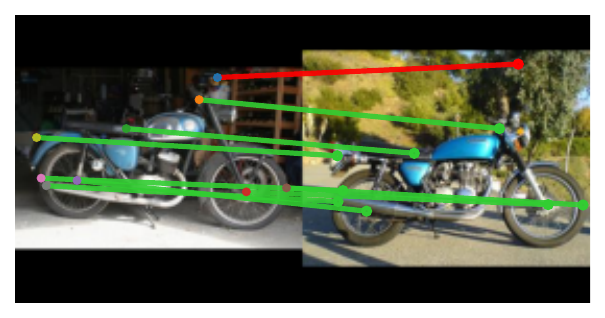} \\
         \includegraphics[width=0.24\linewidth]{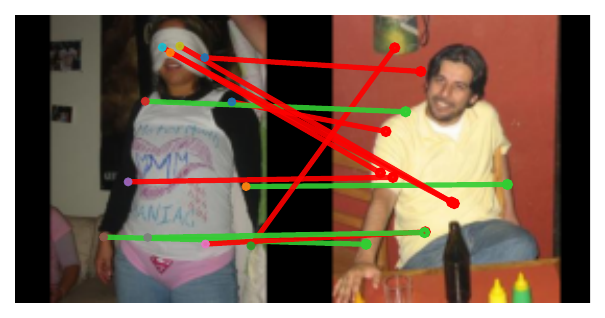}&  \includegraphics[width=0.24\linewidth]{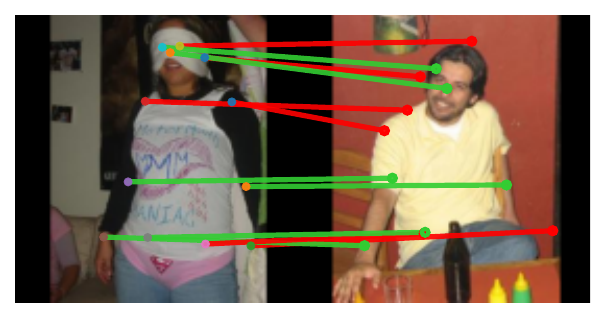} & \includegraphics[width=0.24\linewidth]{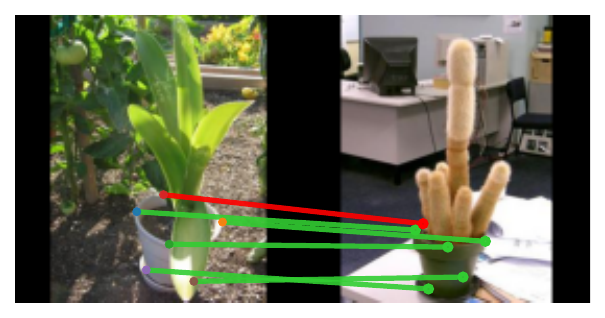}&  \includegraphics[width=0.24\linewidth]{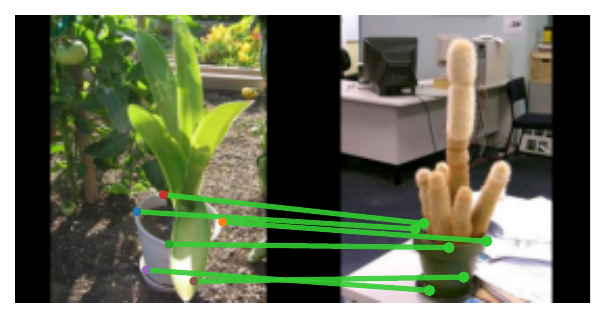} \\
         \includegraphics[width=0.24\linewidth]{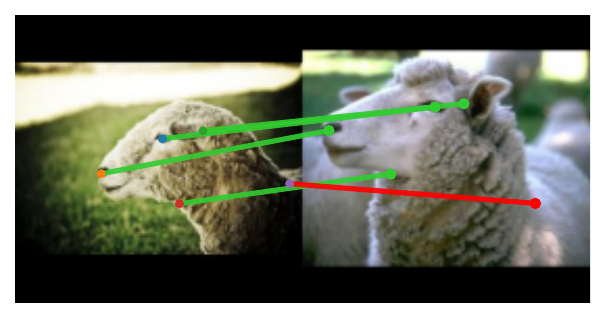}&  \includegraphics[width=0.24\linewidth]{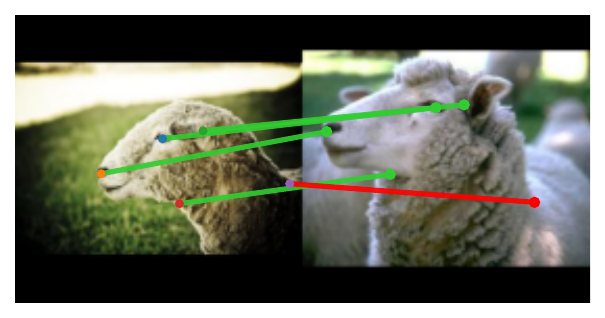} & \includegraphics[width=0.24\linewidth]{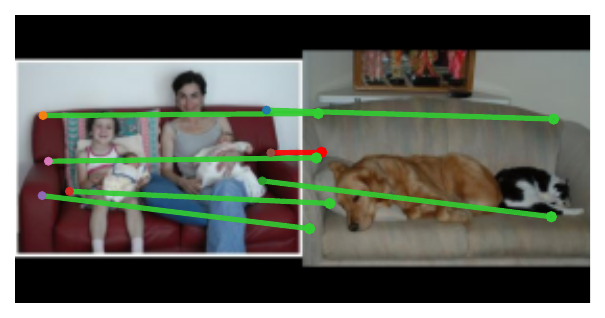}&  \includegraphics[width=0.24\linewidth]{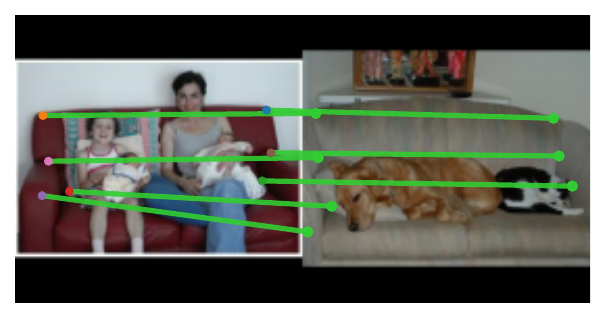} \\
         \includegraphics[width=0.24\linewidth]{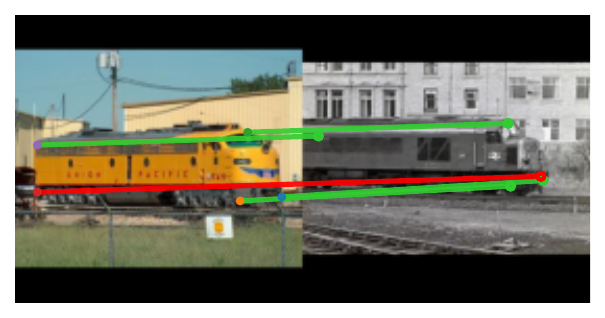}&  \includegraphics[width=0.24\linewidth]{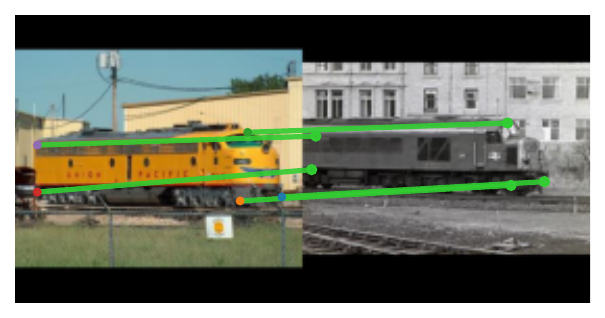} & \includegraphics[width=0.24\linewidth]{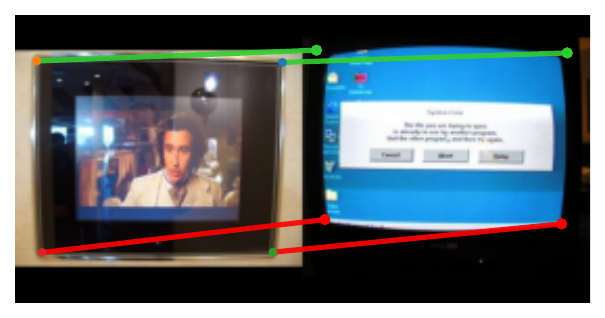}&  \includegraphics[width=0.24\linewidth]{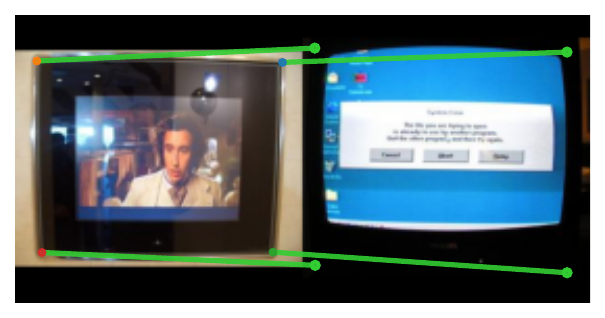} \\
         {\footnotesize CleanDIFT} & {\footnotesize CleanDIFT+Ours} & {\footnotesize CleanDIFT} & {\footnotesize CleanDIFT+Ours}\\
    \end{tabular}
    \caption{\textbf{Qualitative comparison on PF-Pascal with CleanDIFT as the base method.} Correspondences are shown as colored lines: green for correct and red for incorrect matches (PCK@0.1). Our method (CleanDIFT+Ours) consistently reduces incorrect matches across all 20 PF-Pascal categories.}
    \label{fig:cleandift_pascal_supp}
\end{figure*}

\Cref{fig:tlr_pascal_supp,fig:cleandift_pascal_supp} provide full per-category qualitative comparisons on PF-Pascal for TLR and CleanDIFT. PF-Pascal's 20 categories include several not present in SPair-71k (e.g., dining table, sofa), and the visual characteristics differ: images tend to have cleaner backgrounds and less extreme viewpoint variation. Nevertheless, cycle-consistent re-ranking yields consistent improvements across the majority of categories in both base methods, with particularly visible gains in categories with repeated parts (e.g., dining table legs).

\newpage

\newpage

%% file: tables/perkpts_allpck.tex
\begin{table*}[tb]
    \centering
    \begin{adjustbox}{width=\textwidth}
    \begin{tabular}{@{}cllll|lll|lll@{}}
    \toprule
    & \multirow{2}{*}{Method} & \multicolumn{3}{c}{SPair-71k} & \multicolumn{3}{c}{SPair-71k (Geo.)} & \multicolumn{3}{c}{PF-Pascal}\\
    \cmidrule(l){3-5} \cmidrule(l){6-8} \cmidrule(l){9-11} 
    & & 0.01 & 0.05 & 0.10 & 0.01 & 0.05 & 0.10 & 0.05 & 0.1 & 0.15\\\midrule
    \parbox[t]{0mm}{\multirow{1}{*}{\rotatebox[origin=c]{90}{\scriptsize \textbf{US}}}}     
    & SoY-FGW \cite{im_shape--you_2026} & 10.0 & 50.8 & 67.9 & - & - & - & - & - & - \\
    & TLR \cite{zhang_telling_2024} & 10.9 & 53.2 & 69.4 & \underline{7.5} & 44.9 & 62.3  & 74.1 & 88.7 & 91.2\\
    \myrowcolor
    & +Ours & \underline{11.1}\textsubscript{\color{ourgreen}$\blacktriangle$0.2} & \underline{54.9}\textsubscript{\color{ourgreen}$\blacktriangle$1.7} & \underline{71.2}\textsubscript{\color{ourgreen}$\blacktriangle$1.8} & \textbf{8.2\textsubscript{\color{ourgreen}$\blacktriangle$0.7}} & \underline{47.4}\textsubscript{\color{ourgreen}$\blacktriangle$2.5} & \underline{65.0}\textsubscript{\color{ourgreen}$\blacktriangle$2.7} & \underline{76.1}\textsubscript{\color{ourgreen}$\blacktriangle$2.0} & \textbf{89.5\textsubscript{\color{ourgreen}$\blacktriangle$0.8}} & \underline{93.8}\textsubscript{\color{ourgreen}$\blacktriangle$2.6}\\
    & CleanDIFT \cite{stracke_cleandift_2025} & 10.9 & 53.6 & 69.9 & 7.4 & 45.4 & 63.1 & 73.4 & 86.1 & 90.7\\
    \myrowcolor
    & +Ours & \textbf{11.2\textsubscript{\color{ourgreen}$\blacktriangle$0.3}} & \textbf{55.1\textsubscript{\color{ourgreen}$\blacktriangle$1.5}} & \textbf{71.5\textsubscript{\color{ourgreen}$\blacktriangle$1.6}} & \textbf{8.2\textsubscript{\color{ourgreen}$\blacktriangle$0.8}} & \textbf{47.9\textsubscript{\color{ourgreen}$\blacktriangle$2.5}} & \textbf{65.6\textsubscript{\color{ourgreen}$\blacktriangle$2.5}} & \textbf{76.4\textsubscript{\color{ourgreen}$\blacktriangle$3.0}} & \underline{89.4}\textsubscript{\color{ourgreen}$\blacktriangle$3.3} & \textbf{93.9\textsubscript{\color{ourgreen}$\blacktriangle$3.2}}\\
    \midrule\midrule
    \parbox[t]{0mm}{\multirow{1}{*}{\rotatebox[origin=c]{90}{\scriptsize \textbf{WS}}}}  & DistillDIFT$^*$ \cite{fundel_distillation_2024} & - & - & 70.6 & - & - & - & - & - & -\\   
    & DIY-SC$^\dagger$ \cite{dunkel_it_2025} & 10.1 & 54.7 & 74.3 & 6.6 & 46.2 & 68.0 & - & - & -\\
    \myrowcolor 
    & +Ours & 11.0\textsubscript{\color{ourgreen}$\blacktriangle$0.9} & \underline{56.9}\textsubscript{\color{ourgreen}$\blacktriangle$2.2} & \underline{74.6}\textsubscript{\color{ourgreen}$\blacktriangle$0.3} & \underline{7.8}\textsubscript{\color{ourgreen}$\blacktriangle$1.2} & \underline{49.1}\textsubscript{\color{ourgreen}$\blacktriangle$2.9} & \textbf{68.4\textsubscript{\color{ourgreen}$\blacktriangle$0.4}} & - & - & -\\
    & DIY-SC$^\ddagger$ \cite{dunkel_it_2025} & \underline{11.2} & 56.8 & 74.4 & 7.7 & 47.8 & 67.6 & - & - & -\\
    \myrowcolor
    & +Ours & \textbf{12.1\textsubscript{\color{ourgreen}$\blacktriangle$0.9}} & \textbf{58.8\textsubscript{\color{ourgreen}$\blacktriangle$2.0}} & \textbf{74.9\textsubscript{\color{ourgreen}$\blacktriangle$0.5}} & \textbf{9.1\textsubscript{\color{ourgreen}$\blacktriangle$1.4}} & \textbf{50.5\textsubscript{\color{ourgreen}$\blacktriangle$2.7}} & \underline{68.2}\textsubscript{\color{ourgreen}$\blacktriangle$0.6} & - & - & -\\
    \bottomrule
    \end{tabular}
    \end{adjustbox}
    \caption{\textbf{PCK scores (\textit{per-keypoint}) across multiple thresholds on SPair-71k, SPair-71k (Geo.), and PF-Pascal}. DIY-SC was evaluated in both DINO ($\dagger$) and SD+DINO ($\ddagger$) settings. The best results are bolded and the second best underlined.}
    \label{tab:perkpts_allpck}
\end{table*}

%% file: tables/spair_geo_percat_perkpts_pck010.tex
\begin{table*}[tb]
    \centering
    \begin{adjustbox}{width=\textwidth}
    \begin{tabular}{@{}clcccccccccccccccccc|ll@{}}
    \toprule
    & Method & \faIcon{plane} & \faIcon{bicycle} & \faIcon{crow} & \faIcon{ship} & \faIcon{wine-bottle} & \faIcon{bus} & \faIcon{car} & \faIcon{cat} & \faIcon{chair} & \Cow & \faIcon{dog} & \faIcon{horse} & \faIcon{motorcycle} & \faIcon{walking} & \Plant & \Sheep & \faIcon{train} & \faIcon{tv} & Avg \\
    \midrule
    \cmidrule{1-21}
    \parbox[t]{0mm}{\multirow{1}{*}{\rotatebox[origin=c]{90}{\scriptsize \textbf{US}}}}
    & TLR \cite{zhang_telling_2024} & \underline{60.6} & 47.4 & \underline{77.3} & \underline{30.0} & 55.1 & \underline{66.8} & 58.8 & 65.9 & 56.1 & 71.4 & 51.5 & \underline{65.3} & 44.7 & 44.9 & 59.4 & \underline{57.2} & 85.9 & 55.1 & 62.3\\
    \myrowcolor
    & +Ours & \textbf{62.0} & 47.1 & \textbf{77.5} & 29.2 & \underline{57.1} & \textbf{67.8} & \textbf{60.0} & \textbf{68.6} & \underline{61.5} & \underline{72.9} & \textbf{55.0} & 64.4 & \underline{45.8} & \underline{47.1} & \underline{64.9} & 57.0 & \underline{87.7} & \underline{62.2} & \underline{65.0}\textsubscript{\color{ourgreen}$\blacktriangle$2.7} \\
    & CleanDIFT \cite{stracke_cleandift_2025} & 60.0 & \textbf{49.0} & 75.9 & 29.6 & 55.3 & \textbf{67.8} & 59.0 & 66.6 & 58.0 & 72.4 & 49.6 & \textbf{65.4} & \textbf{48.1} & 43.8 & 60.2 & \textbf{57.7} & 86.2 & 58.2 & 63.1 \\
    \myrowcolor
    & +Ours & \textbf{62.0} & \underline{48.1} & 75.9 & \textbf{31.4} & \textbf{57.4} & \textbf{67.8} & \underline{59.9} & \underline{68.4} & \textbf{63.6} & \textbf{73.4} & \underline{53.1} & 64.6 & \textbf{48.1} & \textbf{47.2} & \textbf{66.0} & \underline{57.2} & \textbf{87.8} & \textbf{64.5} & \textbf{65.6}\textsubscript{\color{ourgreen}$\blacktriangle$2.5}\\
    \midrule\midrule
    \parbox[t]{0mm}{\multirow{1}{*}{\rotatebox[origin=c]{90}{\scriptsize \textbf{WS}}}}
    & DIY-SC$^\dagger$ \cite{dunkel_it_2025} & \textbf{53.7} & \underline{47.6} & \underline{73.7} & 32.7 & \underline{59.9} & 83.7 & \underline{78.3} & \textbf{76.1} & 54.0 & \underline{78.7} & \textbf{59.3} & \textbf{70.4} & 46.2 & 44.4 & 63.4 & \underline{63.3} & \underline{80.2} & 70.6 & 68.0 \\
    \myrowcolor
    & +Ours & \underline{52.7} & \textbf{49.6} & 72.5 & 33.3 & \textbf{60.3} & \underline{84.0} & \textbf{78.6} & \underline{76.0} & \underline{56.9} & 78.3 & \textbf{59.3} & \underline{69.5} & \textbf{46.6} & \textbf{46.0} & 64.7 & \textbf{64.7} & \textbf{80.3} & \underline{70.7} & \textbf{68.4}\textsubscript{\color{ourgreen}$\blacktriangle$0.4}\\
    & DIY-SC$^\ddagger$ \cite{dunkel_it_2025} & 50.4 & 46.2 & \textbf{75.3} & \underline{34.9} & 58.9 & 83.6 & 75.7 & 73.3 & 54.8 & 78.5 & \underline{56.7} & 69.1 & 45.6 & \underline{45.7} & \underline{72.4} & 61.3 & 77.9 & 70.0 & 67.6\\
    \myrowcolor
    & +Ours & 50.1 & 45.9 & 72.5 & \textbf{35.8} & \underline{59.9} & \textbf{84.8} & 75.6 & 71.6 & \textbf{58.1} & \textbf{78.9} & 56.0 & 68.7 & \underline{46.4} & 45.6 & \textbf{73.6} & 61.7 & 77.9 & \textbf{71.9} & \underline{68.2}\textsubscript{\color{ourgreen}$\blacktriangle$0.6}\\
    \bottomrule
    \end{tabular}
    \end{adjustbox}
    \caption{\textbf{Per-category PCK@0.10 scores (\textit{per-keypoint}) on SPair-71k (Geo.).} DIY-SC was evaluated in both DINO ($\dagger$) and SD+DINO ($\ddagger$) settings. The best results are bolded and the second best underlined.}
    \label{tab:spair_geo_percat_perkpts_pck010}
\end{table*}


%% file: tables/pascal_percat_pck015.tex
\begin{table*}[tb]
    \centering
    \begin{adjustbox}{width=\textwidth}
    \begin{tabular}{@{}clcccccccccccccccccccc|ll@{}}
    \toprule
    & Method & \faIcon{plane} & \faIcon{bicycle} & \faIcon{crow} & \faIcon{ship} & \faIcon{wine-bottle} & \faIcon{bus} & \faIcon{car} & \faIcon{cat} & \faIcon{chair} & \Cow & \DiningTable & \faIcon{dog} & \faIcon{horse} & \faIcon{motorcycle} & \faIcon{walking} & \faIcon{seedling} & \Sheep & \faIcon{couch} & \faIcon{train} & \faIcon{tv} & Avg \\
    \midrule
    \cmidrule{1-21}
    & TLR \cite{zhang_telling_2024} & 94.2 & 97.3 & \textbf{94.9} & \underline{84.4} & \underline{90.6} & 94.4 & 95.2 & 92.9 & 72.4 & 95.8 & 70.0 & 93.9 & 88.0 & \underline{97.3} & 84.2 & 87.8 & \textbf{100.0} & 85.7 & 87.9 & \underline{74.2} & 91.2\\
    \myrowcolor
    & +Ours & \underline{94.7} & \textbf{98.2} & \textbf{94.9} & \textbf{87.5} & \textbf{98.4} & \underline{97.2} & \underline{98.2} & \textbf{96.2} & \underline{78.1} & \textbf{100.0} & \underline{75.0} & \textbf{96.3} & \underline{89.2} & \textbf{98.1} & \underline{85.3} & \underline{90.2} & \textbf{100.0} & \textbf{92.0} & \textbf{92.9} & \textbf{82.3} & \underline{93.8}\textsubscript{\color{ourgreen}$\blacktriangle$2.6}\\
    & CleanDIFT \cite{stracke_cleandift_2025} & \underline{94.7} & 97.3 & \underline{93.6} & \textbf{87.5} & 87.5 & 94.4 & 96.4 & 90.8 & 71.4 & \underline{97.9} & 70.0 & 89.0 & 88.0 & 96.9 & 84.2 & \textbf{92.7} & \textbf{100.0} & 83.0 & 88.9 & \underline{74.2} & 90.7\\
    \myrowcolor
    & +Ours & \textbf{95.3} & \underline{97.9} & \underline{93.6} & \textbf{87.5} & 89.1 & \textbf{97.7} & \textbf{99.4} & \underline{95.1} & \textbf{84.8} & \textbf{100.0} & \textbf{80.0} & \underline{95.1} & \textbf{92.8} & \underline{97.3} & \textbf{89.8} & \textbf{92.7} & \textbf{100.0} & \underline{90.2} & \underline{91.9} & 72.6 & \textbf{93.9}\textsubscript{\color{ourgreen}$\blacktriangle$3.2}\\
    \midrule\midrule
    & TLR \cite{zhang_telling_2024} & 94.6 & 97.2 & \textbf{94.5} & 84.7 & \underline{90.6} & 93.7 & 94.8 & 92.5 & 68.7 & 95.8 & 71.4 & 94.8 & 85.4 & \underline{97.3} & 83.4 & 87.6 & \textbf{100.0} & 84.1 & 88.0 & \underline{74.4} & 90.0\\
    \myrowcolor
    & +Ours & \underline{95.1} & \textbf{98.1} & \textbf{94.5} & \textbf{88.9} & \textbf{98.4} & \underline{96.8} & \underline{98.2} & \textbf{96.7} & \underline{74.4} & \textbf{100.0} & \underline{75.5} & \textbf{96.7} & \underline{87.9} & \textbf{98.2} & \underline{85.3} & 90.5 & \textbf{100.0} & \textbf{90.1} & \textbf{93.0} & \textbf{82.2} & \textbf{93.2}\textsubscript{\color{ourgreen}$\blacktriangle$3.2}\\
    & CleanDIFT \cite{stracke_cleandift_2025} & \underline{95.1} & 97.2 & \underline{93.4} & \underline{87.5} & 87.5 & 93.4 & 96.3 & 89.8 & 67.8 & \underline{97.9} & 72.4 & 89.8 & 85.4 & 96.9 & 83.2 & \underline{92.4} & \textbf{100.0} & 79.6 & 89.0 & \underline{74.4} & 89.4\\
    \myrowcolor
    & +Ours & \textbf{95.6} & \underline{97.8} & \underline{93.4} & \underline{87.5} & 89.1 & \textbf{97.6} & \textbf{99.5} & \underline{95.4} & \textbf{81.5} & \textbf{100.0} & \textbf{80.2} & \underline{95.7} & \textbf{91.3} & \underline{97.3} & \textbf{89.2} & \textbf{92.9} & \textbf{100.0} & \underline{87.0} & \underline{92.0} & 72.8 & \underline{92.8}\textsubscript{\color{ourgreen}$\blacktriangle$3.4}\\
    \bottomrule
    \end{tabular}
    \end{adjustbox}
    \caption{\textbf{Per-category PCK@0.15 scores on PF-Pascal}, evaluated per-keypoint (top) and per-image (bottom) across 20 object categories. The best results are bolded and the second best underlined.}
    \label{tab:pascal_percat_pck015}
\end{table*}

%% file: tables/spair_percat_perimg_pck010.tex
\begin{table*}[tb]
    \centering
    \begin{adjustbox}{width=\textwidth}
    \begin{tabular}{@{}clcccccccccccccccccc|ll@{}}
    \toprule
    & Method & \faIcon{plane} & \faIcon{bicycle} & \faIcon{crow} & \faIcon{ship} & \faIcon{wine-bottle} & \faIcon{bus} & \faIcon{car} & \faIcon{cat} & \faIcon{chair} & \Cow & \faIcon{dog} & \faIcon{horse} & \faIcon{motorcycle} & \faIcon{walking} & \Plant & \Sheep & \faIcon{train} & \faIcon{tv} & Avg \\
    \midrule
    \cmidrule{1-21}
    \parbox[t]{0mm}{\multirow{1}{*}{\rotatebox[origin=c]{90}{\scriptsize \textbf{US}}}}
    & TLR \cite{zhang_telling_2024} & \underline{76.5} & 63.2 & \textbf{90.5} & 46.8 & 53.5 & \underline{57.1} & \textbf{50.4} & 82.9 & 49.4 & 79.0 & \underline{66.7} & \underline{72.6} & 63.6 & 68.4 & 54.5 & \textbf{64.1} & 82.3 & 53.5 & 65.0\\
    \myrowcolor
    & +Ours & \underline{76.5} & \underline{64.8} & \underline{90.0} & \underline{47.2} & \underline{56.0} & 56.3 & 49.6 & \textbf{83.3} & \underline{52.9} & \underline{79.6} & \textbf{67.6} & 71.1 & \underline{64.1} & \textbf{69.6} & \underline{58.7} & \underline{63.4} & \textbf{84.3} & \underline{60.1} & \underline{66.3}\textsubscript{\color{ourgreen}$\blacktriangle$1.3}\\
    & CleanDIFT \cite{stracke_cleandift_2025} & 76.4 & 63.7 & \underline{90.0} & 46.0 & 53.7 & \textbf{57.7} & \underline{49.9} & 82.9 & 50.7 & \textbf{79.7} & 65.8 & \textbf{73.4} & 63.8 & 67.9 & 55.9 & \textbf{64.1} & 82.3 & 56.6 & 65.4\\
    \myrowcolor
    & +Ours & \textbf{77.1} & \textbf{64.9} & 89.5 & \textbf{47.4} & \textbf{56.1} & 56.2 & 49.0 & \underline{83.0} & \textbf{54.3} & \underline{79.6} & 66.5 & 71.4 & \textbf{64.3} & \underline{69.4} & \textbf{60.9} & 63.2 & \underline{84.1} & \textbf{62.5} & \textbf{66.5}\textsubscript{\color{ourgreen}$\blacktriangle$1.1}\\
    \midrule\midrule
    \parbox[t]{0mm}{\multirow{1}{*}{\rotatebox[origin=c]{90}{\scriptsize \textbf{WS}}}}
    & DIY-SC \cite{dunkel_it_2025} & \underline{77.7} & 66.4 & \underline{90.9} & 54.0 & \underline{58.6} & \underline{80.8} & \underline{75.5} & \textbf{88.8} & 50.1 & \underline{85.2} & \underline{71.4} & \textbf{78.2} & 68.1 & 71.8 & 59.1 & \underline{70.6} & \underline{76.9} & \underline{68.3} & 71.2\\
    \myrowcolor
    & +Ours & 77.4 & \underline{66.5} & 90.0 & 54.8 & \textbf{59.2} & \textbf{81.2} & \textbf{75.6} & \underline{88.3} & \underline{52.0} & 85.1 & \textbf{71.6} & \underline{77.5} & 67.3 & \textbf{72.8} & 60.7 & \textbf{71.0} & \textbf{77.4} & \underline{68.3} & 71.5\textsubscript{\color{ourgreen}$\blacktriangle$0.3}\\
    & DIY-SC \cite{dunkel_it_2025} & \textbf{78.7} & \textbf{66.8} & \textbf{91.4} & \underline{55.9} & 57.2 & 79.6 & 71.2 & 86.7 & 50.6 & \textbf{85.4} & 70.5 & 76.8 & \textbf{70.9} & 72.0 & \underline{69.2} & 68.9 & 75.2 & 67.9 & \underline{71.6}\\
    \myrowcolor
    & +Ours & 77.6 & \textbf{66.8} & 90.4 & \textbf{57.0} & 58.5 & 80.0 & 71.1 & 85.7 & \textbf{52.8} & \textbf{85.4} & 69.9 & 76.0 & \underline{70.4} & \underline{72.3} & \textbf{69.4} & 68.2 & 75.8 & \textbf{69.6} & \textbf{71.7}\textsubscript{\color{ourgreen}$\blacktriangle$0.1}\\
    \bottomrule
    \end{tabular}
    \end{adjustbox}
    \caption{\textbf{Per-category PCK@0.10 scores (\textit{per-image}) on SPair-71k}. DIY-SC was evaluated in both DINO ($\dagger$) and SD+DINO ($\ddagger$) settings. The best results are bolded and the second best underlined.}
    \label{tab:spair_percat_perimg_pck010}
\end{table*}


%% file: tables/spair_geo_percat_perimg_pck010.tex
\begin{table*}[tb]
    \centering
    \begin{adjustbox}{width=\textwidth}
    \begin{tabular}{@{}clcccccccccccccccccc|ll@{}}
    \toprule
    & Method & \faIcon{plane} & \faIcon{bicycle} & \faIcon{crow} & \faIcon{ship} & \faIcon{wine-bottle} & \faIcon{bus} & \faIcon{car} & \faIcon{cat} & \faIcon{chair} & \Cow & \faIcon{dog} & \faIcon{horse} & \faIcon{motorcycle} & \faIcon{walking} & \Plant & \Sheep & \faIcon{train} & \faIcon{tv} & Avg \\
    \midrule
    \cmidrule{1-21}
    \parbox[t]{0mm}{\multirow{1}{*}{\rotatebox[origin=c]{90}{\scriptsize \textbf{US}}}}
    & TLR \cite{zhang_telling_2024} & \underline{60.6} & 47.4 & \underline{77.3} & \underline{30.0} & 55.1 & \underline{66.8} & 58.8 & 65.9 & 56.1 & 71.4 & 51.5 & \underline{65.3} & 44.7 & 44.9 & 59.4 & \underline{57.2} & 85.9 & 55.1 & 56.4\\
    \myrowcolor
    & +Ours & \textbf{62.0} & 47.1 & \textbf{77.5} & 29.2 & \underline{57.1} & \textbf{67.8} & \textbf{60.0} & \textbf{68.6} & \underline{61.5} & \underline{72.9} & \textbf{55.0} & 64.4 & \underline{45.8} & \underline{47.1} & \underline{64.9} & 57.0 & \underline{87.7} & \underline{62.2} & \underline{58.2}\textsubscript{\color{ourgreen}$\blacktriangle$1.8}\\
    & CleanDIFT \cite{stracke_cleandift_2025} & 60.0 & \textbf{49.0} & 75.9 & 29.6 & 55.3 & \textbf{67.8} & 59.0 & 66.6 & 58.0 & 72.4 & 49.6 & \textbf{65.4} & \textbf{48.1} & 43.8 & 60.2 & \textbf{57.7} & 86.2 & 58.2 & 56.8\\
    \myrowcolor
    & +Ours & \textbf{62.0} & \underline{48.1} & 75.9 & \textbf{31.4} & \textbf{57.4} & \textbf{67.8} & \underline{59.9} & \underline{68.4} & \textbf{63.6} & \textbf{73.4} & \underline{53.1} & 64.6 & \textbf{48.1} & \textbf{47.2} & \textbf{66.0} & \underline{57.2} & \textbf{87.8} & \textbf{64.5} & \textbf{58.7}\textsubscript{\color{ourgreen}$\blacktriangle$1.9}\\
    \midrule\midrule
    \parbox[t]{0mm}{\multirow{1}{*}{\rotatebox[origin=c]{90}{\scriptsize \textbf{WS}}}}
    & DIY-SC$^\dagger$ \cite{dunkel_it_2025} & \textbf{53.7} & \underline{47.6} & \underline{73.7} & 32.7 & \underline{59.9} & 83.7 & \underline{78.3} & \textbf{76.1} & 54.0 & \underline{78.7} & \textbf{59.3} & \textbf{70.4} & 46.2 & 44.4 & 63.4 & \underline{63.3} & \underline{80.2} & 70.6 & \underline{63.3}\\
    \myrowcolor
    & +Ours & \underline{52.7} & \textbf{49.6} & 72.5 & 33.3 & \textbf{60.3} & \underline{84.0} & \textbf{78.6} & \underline{76.0} & \underline{56.9} & 78.3 & \textbf{59.3} & \underline{69.5} & \textbf{46.6} & \textbf{46.0} & 64.7 & \textbf{64.7} & \textbf{80.3} & \underline{70.7} & \textbf{63.6}\textsubscript{\color{ourgreen}$\blacktriangle$0.3}\\
    & DIY-SC$^\ddagger$ \cite{dunkel_it_2025} & 50.4 & 46.2 & \textbf{75.3} & \underline{34.9} & 58.9 & 83.6 & 75.7 & 73.3 & 54.8 & 78.5 & \underline{56.7} & 69.1 & 45.6 & \underline{45.7} & \underline{72.4} & 61.3 & 77.9 & 70.0 & 63.0\\
    \myrowcolor
    & +Ours & 50.1 & 45.9 & 72.5 & \textbf{35.8} & \underline{59.9} & \textbf{84.8} & 75.6 & 71.6 & \textbf{58.1} & \textbf{78.9} & 56.0 & 68.7 & \underline{46.4} & 45.6 & \textbf{73.6} & 61.7 & 77.9 & \textbf{71.9} & 63.0\\
    \bottomrule
    \end{tabular}
    \end{adjustbox}
    \caption{\textbf{Per-category PCK@0.10 scores (\textit{per-image}) on SPair-71k (Geo.).} DIY-SC was evaluated in both DINO ($\dagger$) and SD+DINO ($\ddagger$) settings. The best results are bolded and the second best underlined.}
    \label{tab:spair_geo_percat_perimg_pck010}
\end{table*}


%% file: tables/composition.tex
\begin{table*}[tb]
    \centering
    \begin{adjustbox}{width=0.7\linewidth}
    \begin{tabular}{@{}cccccc@{}}
    \toprule
    Adaptive pose & Window soft-argmax & Ours & PCK@0.05 & PCK@0.10 & $\Delta$@0.10\\
    \midrule
    $\times$ & $\checkmark$ & $\times$ & 49.82 & 65.36 & --\\
    $\times$ & $\checkmark$ & $\checkmark$ & 50.18 & 66.24 & +0.88\\
    \midrule
    $\checkmark$ & $\times$ & $\times$ & 52.73 & 68.53 & --\\
    $\checkmark$ & $\times$ & $\checkmark$ & 54.46 & 70.36 & +1.83\\
    \midrule
    $\checkmark$ & $\checkmark$ & $\times$ & 54.10 & 69.79 & --\\
    \myrowcolor
    $\checkmark$ & $\checkmark$ & $\checkmark$ & \textbf{54.85} & \textbf{71.16} & +1.37\\
    \bottomrule
    \end{tabular}
    \end{adjustbox}
    \caption{\textbf{Composition with existing refinement components on SPair-71k (\textit{per-keypoint}, full test set)} with TLR as the base method. Our re-ranking adds to the pipeline in all three configurations, and adds \textit{more} when window soft-argmax is disabled (+1.83) than when it is enabled (+1.37).}
    \label{tab:composition}
\end{table*}

%% file: tables/winloss.tex
\begin{table*}[tb]
    \centering
    \begin{adjustbox}{width=\linewidth}
    \begin{tabular}{@{}lcccccc@{}}
    \toprule
    Method & Keypoints changed & right$\rightarrow$wrong ($b$) & wrong$\rightarrow$right ($c$) & Win rate & $\Delta$PCK@0.10 & $p$ (McNemar)\\
    \midrule
    MNN & 4.9\% & 1{,}030 & 1{,}279 & 55.4\% & +0.28 & $2.5\times10^{-7}$\\
    Binary penalty & 3.7\% & 665 & 1{,}038 & 61.0\% & +0.42 & $2.0\times10^{-19}$\\
    Relaxed MNN & 7.6\% & 1{,}331 & 2{,}085 & 61.0\% & +0.85 & $5.6\times10^{-38}$\\
    \myrowcolor
    Ours & \textbf{13.7\%} & 2{,}213 & \textbf{3{,}423} & 60.7\% & \textbf{+1.37} & $2.4\times10^{-58}$\\
    \bottomrule
    \end{tabular}
    \end{adjustbox}
    \caption{\textbf{How often each cycle rule fixes or breaks a prediction on SPair-71k (\textit{per-keypoint}, $N=88{,}328$)} with TLR as the base method, compared against the baseline without a cycle term from \Cref{tab:cycle_rules}. Win rate is $c/(b+c)$, the share of flipped keypoints that flip from wrong to right.}
    \label{tab:winloss}
\end{table*}

%% file: tables/hyperparam_sweep.tex
\begin{table*}[tb]
    \centering
    \begin{adjustbox}{width=0.6\linewidth}
    \begin{tabular}{@{}lcc@{}}
    \toprule
    Setting & PCK@0.10 & PCK $\Delta$ vs.\ default\\
    \midrule
    \myrowcolor
    Default ($\tau=0.1$, $K_{\min}=1$, $K_{\max}=30$) & 71.16 & --\\
    \midrule
    $\tau = 0.01$ & 70.99 & $-0.17$\\
    $\tau = 0.5$ & 71.16 & $0.00$\\
    \midrule
    $K_{\min} = 3$ & 71.16 & $0.00$\\
    $K_{\min} = 10$ & 71.16 & $0.00$\\
    \midrule
    $K_{\max} = 10$ & 71.10 & $-0.06$\\
    $K_{\max} = 3{,}600$ (no cutoff) & 71.03 & $-0.13$\\
    \bottomrule
    \end{tabular}
    \end{adjustbox}
    \caption{\textbf{Sensitivity to $\tau$, $K_{\min}$ and $K_{\max}$ on SPair-71k (\textit{per-keypoint}, full test set)} with TLR as the base method. The feature grid has $L=3{,}600$ patches, so $K_{\max}=3{,}600$ removes the cutoff entirely. The total range across every setting is $0.17$ PCK@0.10. Unlike $\lambda$, these three parameters do not appear in the re-ranking score of \Cref{eq:reranking_score}; they only control how many candidates are nominated.}
    \label{tab:hyperparam_sweep}
\end{table*}

%% file: tables/supervised.tex
\begin{table*}[tb]
    \centering
    \begin{adjustbox}{width=0.65\linewidth}
    \begin{tabular}{@{}lcccc@{}}
    \toprule
    Method & PCK@0.10 & $\Delta$ & Win rate & $p$ (McNemar)\\
    \midrule
    Baseline & 87.55 & -- & -- & --\\
    MNN & 86.51 & $-1.04$ & 26.6\% & $7.8\times10^{-95}$\\
    Relaxed MNN & 87.22 & $-0.33$ & 41.8\% & $4.1\times10^{-12}$\\
    Binary penalty & 87.41 & $-0.14$ & 42.0\% & $1.5\times10^{-5}$\\
    \myrowcolor
    Ours & \textbf{87.56} & $+0.01$ & 50.3\% & $0.81$ (n.s.)\\
    \bottomrule
    \end{tabular}
    \end{adjustbox}
    \caption{\textbf{Supervised setting on SPair-71k (\textit{per-keypoint}, full test set)}, using the released supervised TLR \cite{zhang_telling_2024} checkpoint and its released protocol.}
    \label{tab:supervised}
\end{table*}

%% file: tables/mask_ablation.tex
\begin{table*}[tb]
    \centering
    \begin{adjustbox}{width=0.6\linewidth}
    \begin{tabular}{@{}cccc@{}}
    \toprule
    Mask-constrained nomination & Cycle re-ranking & PCK@0.10 & $\Delta$\\
    \midrule
    $\times$ & $\times$ & 69.14 & --\\
    $\times$ & $\checkmark$ & 70.20 & $+1.06$\\
    \midrule
    $\checkmark$ & $\times$ & 69.79 & --\\
    \myrowcolor
    $\checkmark$ & $\checkmark$ & \textbf{71.16} & $+1.37$\\
    \bottomrule
    \end{tabular}
    \end{adjustbox}
    \caption{\textbf{Effect of mask-constrained candidate nomination on SPair-71k (\textit{per-keypoint}, full test set)} with TLR as the base method. With the mask stripped entirely, the cycle criterion is still worth $+1.06$. The mask adds no new dependency: the released TLR pipeline already computes these masks for adaptive pose alignment, and we simply reuse them.}
    \label{tab:mask_ablation}
\end{table*}

%% file: tables/kap.tex
\begin{table*}[tb]
    \centering
    \begin{adjustbox}{width=0.6\linewidth}
    \begin{tabular}{@{}lcccc@{}}
    \toprule
    Method & KAP@0.1 & $\Delta$ & KAP@0.05 & $\Delta$\\
    \midrule
    Baseline & 64.06 & -- & 54.97 & --\\
    Binary penalty & 66.11 & $+2.05$ & 56.75 & $+1.78$\\
    \myrowcolor
    Ours & \textbf{69.00} & $\mathbf{+4.94}$ & \textbf{57.60} & $\mathbf{+2.63}$\\
    \bottomrule
    \end{tabular}
    \end{adjustbox}
    \caption{\textbf{Keypoint Average Precision (KAP) \cite{mariotti_improving_2024} on SPair-71k (full test set)} with TLR as the base method, reported as the per-category mean AP. Unlike PCK, KAP compares the best score near the ground truth against the best score elsewhere, so it penalizes both missed matches and wrong matches to repeated or symmetric parts.}
    \label{tab:kap}
\end{table*}